\documentclass[journal]{IEEEtran}
\usepackage{graphicx}
\usepackage[fleqn]{amsmath}
\usepackage[linesnumbered,ruled]{algorithm2e}
\usepackage{color}

\ifCLASSINFOpdf
\else
\fi
\begin{document}
\title{An Automatic Design Framework of Swarm Pattern Formation based on Multi-objective Genetic Programming}


%\author{
%\IEEEauthorblockN{Zhun Fan\IEEEauthorrefmark{1},
%Zhaojun Wang\IEEEauthorrefmark{2},
%Wenji Li\IEEEauthorrefmark{2},
%Montgomery Scott\IEEEauthorrefmark{3}, and
%Eldon Tyrell\IEEEauthorrefmark{4},~\IEEEmembership{Fellow,~IEEE}}
%\IEEEauthorblockA{\IEEEauthorrefmark{1}School of Electrical and Computer Engineering,
%Georgia Institute of Technology, Atlanta, GA 30332 USA}
%\IEEEauthorblockA{\IEEEauthorrefmark{2}Twentieth Century Fox, Springfield, USA}
%\IEEEauthorblockA{\IEEEauthorrefmark{3}Starfleet Academy, San Francisco, CA 96678 USA}
%\IEEEauthorblockA{\IEEEauthorrefmark{4}Tyrell Inc., 123 Replicant Street, Los Angeles, CA 90210 USA}% <-this % stops an unwanted space
%\thanks{Manuscript received December 1, 2012; revised August 26, 2015.
%Corresponding author: M. Shell (email: http://www.michaelshell.org/contact.html).}
%}

\author{Zhun Fan,~\IEEEmembership{Senior Member,~IEEE,}
        Zhaojun Wang,
        Xiaomin Zhu,
        Bingliang Hu,
        Anmin Zou,
        and Dongwei Bao
        \\
        \textbf{}
\thanks{}% <-this % stops a space
}

\markboth{Journal of \LaTeX\ Class Files,~Vol.~14, No.~8, August~2015}%
{Shell \MakeLowercase{\textit{et al.}}: Bare Demo of IEEEtran.cls for IEEE Transactions on Magnetics Journals}



% for Transactions on Magnetics papers, we must declare the abstract and
% index terms PRIOR to the title within the \IEEEtitleabstractindextext
% IEEEtran command as these need to go into the title area created by
\IEEEtitleabstractindextext{%
\begin{abstract}
Most existing swarm pattern formation methods depend on a predefined gene regulatory network (GRN) structure that requires designers' priori knowledge, which is difficult to adapt to complex and changeable environments. To dynamically adapt to the complex and changeable environments, we propose an automatic design framework of swarm pattern formation based on multi-objective genetic programming. The proposed framework does not need to define the structure of the GRN-based model in advance, and it applies some basic network motifs to automatically structure the GRN-based model. In addition, a multi-objective genetic programming (MOGP) combines with NSGA-II, namely MOGP-NSGA-II, to balance the complexity and accuracy of the GRN-based model. In evolutionary process, an MOGP-NSGA-II and differential evolution (DE) are applied to optimize the structures and parameters of the GRN-based model in parallel. Simulation results demonstrate that the proposed framework can effectively evolve some novel GRN-based models, and these GRN-based models not only have a simpler structure and a better performance, but also are robust to the complex and changeable environments.
\end{abstract}

}

\maketitle

\IEEEdisplaynontitleabstractindextext

\IEEEpeerreviewmaketitle

\section{Introduction}

\IEEEPARstart{I}{n} general, multi-robot systems (MRSs) are composed of a large number of minimal, simple and low-cost robots, each of which has limited functions and performance \cite{oh2017bio}. In most cases, these simple robots can work together in a collaborative way to accomplish complex and changeable tasks that single robot can not accomplish. In addition, MRSs with parallelism, scalability, stability, low-cost, strong robustness and high adaptability, are widely applied into the collaborative target search and rescue \cite{li2019two}, \cite{bakhshipour2017swarm}, small satellites communication \cite{levchenko2018explore}, \cite{bezouska2019visual}, cooperation and navigation planning \cite{rao2019study}, \cite{xin2019application} and source localization \cite{feng2019source}, \cite{amjadi2019cooperative}, et al.

The swarm pattern formation is a typical task for MRSs and embodies pattern generation and pattern maintenance \cite{guo2012morphogenetic}. In different tasks and dynamic environments, swarm pattern formation represents the coordination and local interaction of multi-robots to generate and maintain a swarm pattern formation with a certain shape, in which the shape of pattern can be either predefined or adaptively formed in a self-organised, coordinate and cooperate way through local interaction with neighbouring robots and the environments \cite{oh2017bio}. In the former case of using a predefined pattern, swarm robots follow a predetermined trajectory and maintain a specific swarm pattern in the execution of tasks. For example, Jin \cite{jin2012hierarchical} proposed a hierarchical gene regulatory network for adaptive multi-robot pattern formation. In this work, the swarm pattern is designed as a band of circle, which encircles targets in a dynamic environments. In the latter case of using adaptive formation, swarm robots follow an adaptive pattern to encircle targets in the environment. For example, Oh et al. \cite{oh2014evolving} have introduced an evolving hierarchical gene regulatory network for morphogenetic pattern formation in order to generate adaptive patterns which are adaptable to dynamic environments.

In addition, swarm pattern formation has been widely explored in recent years, which can be divided into four categories \cite{oh2017bio}, namely morphogenesis, reaction-diffusion model, chemotaxis and gene regulatory network (GRN).

The basic idea of morphogenesis is that the morphogen-like signals can provide the information of relative locations for each robot in swarm robots. For example, Mamei \cite{mamei2004experiments} has studied swarm robots utilize morphogen diffusion to form a circular, ring or polygonal pattern formation. In addition, morphogenesis can be combined with other methods to form arbitrary swarm patterns. For example, Kondacs \cite{kondacs2003biologically} combines morphology and geometry to generate two-dimensional arbitrary swarm patterns.

The second category is reaction-diffusion model, which utilizes several morphogen in a cell to react with morphogen in other neighbouring cells to generate complex patterns \cite{jin2010morphogenetic}. For example, the reaction-diffusion turing pattern \cite{slavkov2018morphogenesis} is a typical example of reaction-diffusion model, which utilizes two hormones, an activator and an inhibitor, to form complex and arbitrary swarm patterns. Moreover, Kondo \cite{kondo2010reaction} has used mathematical models to explain the development patterns of Turing model in biological systems.

Without central control or coordination, chemotaxis can control the movement, aggregation and sorting of swarm robots in pattern formation. For example, Fate and Vlassopoulos \cite{fates2011robust} applied chemotaxis in swarm robots, which forms aggregation formation in a dynamic environment. Bai and Breen \cite{bai2008emergent} demonstrated that chemotaxis can form complex swarm patterns.

The GRN model \cite{cussat2019artificial} is inspired by the reaction-diffusion model, which can control the behavior of each robot in swarm robots. For example, Jin \cite{jin2012hierarchical} applied an evolutionary algorithm to evolve parameters of a GRN subnetwork, which gets a multi-robot pattern formation. Meng \cite{meng2013morphogenetic} combined GRN and B-spline to form complex patterns in swarm robotic systems. Oh \cite{oh2014evolving} utilized some network motifs to evolve GRN, which generates swarm pattern formation. Oh \cite{oh2014adaptive} utilized a two-layer hierarchical GRN to cover a desired region for target entrapment.

A substantial limitation of most existing GRN-based model for swarm pattern formation is that the structure of the network must be predefined, resulting in the inability to adapt to different tasks and dynamic environments. Furthermore, another limitation of swarm pattern formations is that existing evolutionary GRN methods are mainly to optimize the parameters of the network structure, and these methods are unable to optimize topology structures of a GRN-based model, resulting in inability to adapt to the complex dynamic environments.

In order to produce a GRN-based model adapted to the complex dynamic environments, we adopt an idea of automatic design \cite{lipson2000automatic}. The automatic design allows modules from elementary building blocks and operators to automatically generate models to satisfy predefined specifications. The automatic design process is that a modular modeling language is applied to describe elementary building blocks, and then a genetic programming (GP) is employed to automatic optimization to get an optimal model. When automatic design is applied to electromechanical systems, a modular modeling language is bond graph. For example, Erik Goodman \cite{fan2004novel} proposed an approach combining GP and bond graph to use in electromechanical system. Jean-Francois Dupuis \cite{dupuis2011evolutionary} proposed a hybrid system evolutionary design method by combining hybrid bond graph with GP. When automatic design is applied to swarm robots, a modular modeling language is finite-state machine. For example, Lorenzo Garattoni \cite{garattoni2018autonomous} used finite state machine to control swarm robot, which could collectively sequence tasks whose order of execution was a priori unknown.

To tackle limitations of most existing GRN-based model for swarm pattern formation, we propose an automatic design framework of swarm pattern formation based on multi-objective genetic programming (MOGP). The proposed framework does not need to define the structure of GRN-based model in advance. Some basic network motifs are created by the GRN-based of the differential equation model, and they are known as a modular modeling language to describe elementary building blocks. The GRN-based models are automatically structured by employing these basic network motifs. In evolutionary process, to balance the complexity and accuracy of GRN-based models, MOGP and NSGA-II are employed, namely MOGP-NSGA-II. An MOGP-NSGA-II and differential evolution (DE) are applied to optimize the structures and parameters of the GRN-based model in parallel, which can get some GRN-based models with simple structure and excellent performance. In addition, these models also have a good performance when they migrate directly to complex environments. Thus, the proposed framework can enhance human’s understanding of the structure of GRN-base model and realize an innovative design of GRN.

The remainder of this paper is structured as follows. In Section \ref{sec:Problem_Assumptions}, the problem statement of GRN-based model is introduced, and some assumptions are listed. Section \ref{sec:related_work} introduces a generic GRN basic framework and the proposed MOGP. An automatic design framework of GRN-based model for adaptive swarm pattern formations is proposed in Section \ref{sec:evo_design_framwwork}. Section \ref{sec:num_simu} presents numerical simulation results from moving targets and various obstacles, which verifies the adaptability, scalability and robustness of the proposed framework. In Section \ref{sec:conclusion}, conclusions and future works are discussed.

\section{Problem Statement and Assumptions}
\label{sec:Problem_Assumptions}

The studied problem mainly focus on how to generate adaptive swarm patterns in dynamic, complex and changeable environments. More specifically, a swarm pattern needs to form different shapes in various restricted environments.

In biological morphogenesis, morphogen gradients that guide cells migration are either directly obtained from the mother cells or generated by a few cells known as histiocytic cells \cite{jin2010morphogenetic}. Inspired by these biological studies, we assume that there are some mother robots (a mother robot is equivalent to a mother cell in biological morphogenesis) in swarm robots. The mother robots can detect obstacles and targets in the environment and they are responsible for swarm pattern formation. When the mother robots detecting targets (or obstacles and targets) in an environment, the mother robots can form a morphogen gradient space according to the location information of the targets (or obstacles and targets). In the morphogen gradient space, the concentration of morphogen decreases as it moves away form the targets. Points whose gradient value are higher than a certain percentage of the maximum concentration value are selected as candidate points, and swarm robots move to candidate points to form a swarm pattern.

In order to apply the proposed automatic design framework of GRN-based model to the swarm pattern formation. Some assumptions are listed as follows:
\begin{enumerate}
\item The swarm robots can use on-board sensors (such as encoders and sonar sensors) to locate themselves at any time, as reported in \cite{guo2010unified}.
\item The base station contains a sufficient number of robots to complete the pattern formation task. By the circumference of the swarm pattern, a sufficient number of robots can be called from the base station.
\item Each of swarm robot has a limited range of perception. Therefore, swarm robots can detect targets, obstacles and other robots within the range of perception.
\item Each of swarm robot also has a limited range of communication, and the communication between swarm robots and base stations is not limited. In the communication range, the robot can communicate position and speed information with other neighboring robots.
\item The maximum movement speed of each robot is faster than the maximum movement speed of the targets.
\item At least one mother robot can detect all targets and obstacles in the environment.
\end{enumerate}

%\IEEEpubidadjcol
\section{Related Work}
\label{sec:related_work}

In this section, a generic GRN framework using differential equation model is discussed and the basic idea of MOGP and DE is introduced.

\subsection{The Generic GRN Framework}

In order to develop an automatic design framework of GRN-based model, we need to understand the basic process of GRN. The GRN is a network consisted by the interaction of gens in cell-cell. That is, the expression of a gen is affected by other genes, which in turn affect the expression of other genes. The complex interaction between these genes constitutes a suitable GRN. One of the key issues is how to construct GRN. The conventional mathematical models of GRN include boolean network model \cite{kauffman1969metabolic}, bayesian network model \cite{chen2012dynamic} and differential equation model \cite{ji2010modelling}. In this paper, we use the differential equation model to study the swarm pattern formation.

In the differential equation model, a general form of differential equation model \cite{wang2007modeling} for describing GRNs is as follows:
\begin{eqnarray}
\label{equ:DEO_easy}
\frac{dx_{i}}{dt} = f_{i}(x)
\end{eqnarray}
Where $x_{i}$ is the concentration of the mRNA that reflect the expression level of the $i$th gene, and $\frac{dx_{i}}{dt}$ is the rate of change of the $i$th gene at time $t$, so the model is called the dynamic equation. Furthermore, $\frac{dx_{i}}{dt}$ illustrates the regulatory mechanism among genes. A simple form of a linear additive function is as follows:
\begin{eqnarray}
\label{equ:DEO_complex}
\frac{dx_{i}}{dt} = \Sigma^{n}_{j=1}w_{ij}x_{j} + b_{i}
\end{eqnarray}
Where $w_{ij}$ is a real number and a weight of $j$th gen. Furthermore, it describes the type and strength of the influence of the $j$th gene on the $i$th gene. $b_{i}$ is the $i$th gen's external stimulus. For added biological realism (all concentrations get saturated at some point in time $t$), a sigmoid function may be included into the equation.

\begin{figure}[!t]
\centering
\includegraphics[width=2.5in]{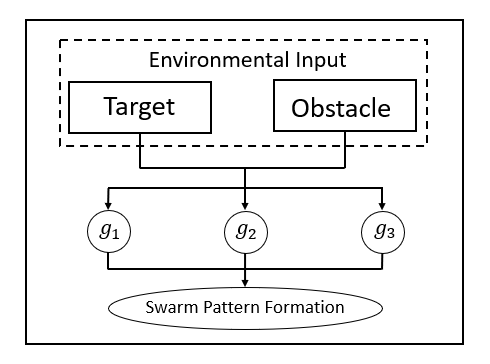}%
\hfil
\caption{A general framework of GRNs for swarm pattern formation.}
\label{fig:GRNs}
\end{figure}

We proposed a General Framework of GRN for Swarm Pattern Formation, as illustrated in Fig. \ref{fig:GRNs}. First, morphogen concentrations are produced based on the location of targets and obstacles in the environment, which serve to activate genes $g_{1}$, $g_{2}$ and $g_{3}$. Then, the activated genes $g_{1}$, $g_{2}$ and $g_{3}$ are combined through a variety of ordinary differential equations to generate swarm patterns.

\subsection{Genetic Programming}

\begin{figure}[!t]
\centering
\includegraphics[width=2.5in]{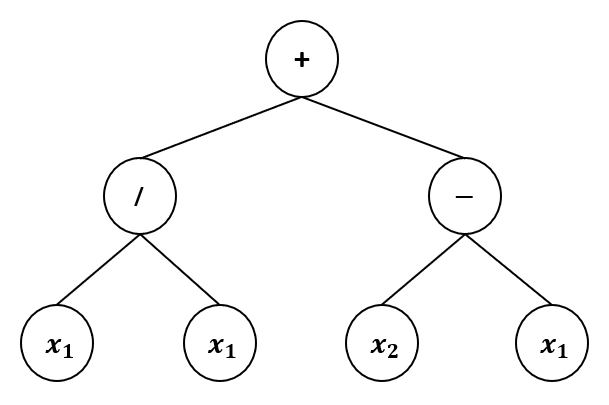}%
\hfil
\caption{A tree in genetic programming.}
\label{fig:GP_tree}
\end{figure}

Genetic programming (GP) is an extension of genetic algorithm (GA). The biggest difference between GP and GA is the encode method. In GP, a  hierarchical structured tree is used to encode the individuals in a population, that is, each individual in the population is a hierarchical structured tree consisting of functions and terminals. The structure of the tree is dynamically and adaptively adjusted. The structure shown in Fig. \ref{fig:GP_tree}, the individual is represented by functions and terminals. In this case, the model is represented as follows:
\begin{eqnarray}
\label{equ:GP_function}
y = \frac{x_{1}}{x_{2}} + x_{2} - x_{1}
\end{eqnarray}

When applying GP to solve some real-world optimization problems \cite{seo2003toward}, \cite{fan2004novel}, \cite{wang2005knowledge}, \cite{hu2005hierarchical}, \cite{hu2008gpbg}, \cite{wang2008cooperative}, \cite{fan2008structured}, \cite{dupuis2011evolutionary}, terminators and operators should be defined according to the characteristics of a problem. GP can optimize the topologies and parameters of the GRN at the same time. For swarm pattern formation. We need to build some predefined network motifs. Second, the positions of targets and obstacles are employed as terminals. In each iteration, the optimal individual is selected by a fitness function evaluation.

In real-world optimization problems, there are usually more than one conflicting objectives. For swarm pattern formation, the complexity and accuracy of the GRN-based model are two conflicting objectives. In this paper, the proposed automatic design framework applies a nesting algorithm, which combines GP and NSGA-II, namely MOGP-NSGA-II \cite{masood2016many}, to optimize the swarm pattern. In MOGP-NSGA-II, NSGA-II \cite{deb2002fast} is one of the state-of-the-art multi-objective evolutionary algorithm, which is applied to balance the complexity and accuracy of GRN-based models.

\subsection{Differential Evolution}

Differential Evolution (DE) \cite{Storn1997Differential} grew out of Ken Price's attempts to solve the Chebychev Polynomial fitting Problem that had been posed to him by Rainer Storn. DE is also a stochastic direct search method. As the number of iterations increases, individuals adapted to the environment are preserved. Like other kinds of evolutionary algorithms, DE contains three operations: mutation, crossover and selection. Among these three operations, mutation operation is very important for DE. Mutation operation generates new individuals by combining randomly selected or given individuals, and its main purpose is to improve the diversity of the population, so as to prevent the algorithm from falling into local optimum and premature convergence.

In this paper, DE is applied to optimize the parameters of GRN-based models, which aims to find optimal parameters in a GRN-based model.

\section{The Automatic Design Framework of Swarm Pattern Formation}
\label{sec:evo_design_framwwork}

In this section, the automatic design framework of swarm pattern formation based on MOGP is introduced. In addition, ten predefined basic network motifs and the fitness function are discussed.

\subsection{Basic Network Motifs}

When applying MOGP-NSGA-II to evolve GRNs, a typical task is to define some basic network motifs. Recent researches, such as biochemistry, neurobiology, ecology and engineering, find that patterns of inter-connections occurring in complex networks are significantly higher than those in randomized networks \cite{milo2002network}. Peter M. Bowers \cite{bowers2004use} proposed that the logic analysis of phylogenetic profiles was applied to identified triplets of proteins whose presence or absence obey certain logic relationships. In addition, the logic relationships in triplets of proteins are also frequently found in GRNs of a multi-cellular organism. Inspired by these researches, ten predefined network motifs, such as positive, negative, AND, OR, XOR and so on, are utilized as the basic network motifs, which are used to construct GRNs.

\emph{1) Positive correlation regulation}\\
A positive regulation is defined as gene $X$ activates gene $Y$. That is, Gene $X$ has a positive feedback effect on $Y$. The mathematical description of the positive correlation regulation from $X$ to $Y$ is defined as follows:
\begin{eqnarray}
\label{equ:positive}
\frac{dy}{dt} = - y + sig(x,\theta,k)
\end{eqnarray}
\begin{eqnarray}
\label{equ:sigmoid}
sig(x,\theta) = \frac{1}{1+e^{-k(x-\theta)}}
\end{eqnarray}
where $x$ represents the expression level of gene $X$, and $y$ represents the expression level of gene $Y$. $\theta$ represents a regulatory parameter for the gene expression, and $k$ is a scale factor for the gene expression.

\emph{2) Negative correlation regulation}\\
A negative regulation is defined as gene $X$ inhibits gene $Y$. That is, Gene $X$ has a negative feedback effect on $Y$. The mathematical description of the negative correlation regulation from $X$ to $Y$ is defined as follows:
\begin{eqnarray}
\label{equ:negative}
\frac{dy}{dt} = - y + (1 - sig(x,\theta,k))
\end{eqnarray}
where $x$ represents the expression level of gene $X$, and $y$ represents the expression level of gene $Y$.

\emph{3) Logical AND regulation}\\
A logical AND regulation is defined as if and only if both gene $X_{1}$ and gene $X_{2}$ express, gene $Y$ expresses. The mathematical description of the logical AND regulation is defined as follows:
\begin{eqnarray}
\label{equ:and}
\frac{dy}{dt} = - y + sig(g_{1} \ast g_{2} ,\theta,k))
\end{eqnarray}
where $g_{1}$ and $g_{2}$ are the expression levels of gene $X_{1}$ and
$X_{2}$, respectively.

\emph{4) Logical NAND regulation}\\
A logical NAND regulation is defined as if either gene $X_{1}$ or gene $X_{2}$ does not express, gene $Y$ expresses. The mathematical description of the logical NAND regulation is defined as follows:
\begin{eqnarray}
\label{equ:nand}
\frac{dy}{dt} = - y + 1 - sig(g_{1} \ast g_{2} ,\theta,k))
\end{eqnarray}
where $g_{1}$ and $g_{2}$ are the expression levels of gene $X_{1}$ and
$X_{2}$, respectively.

\emph{5) Logical OR regulation}\\
A logical OR regulation is defined as if either gene $X_{1}$ or gene $X_{2}$ expresses, gene $Y$ expresses. The mathematical description of the logical OR regulation is defined as follows:
\begin{eqnarray}
\label{equ:or}
\frac{dy}{dt} = - y + sig(g_{1} + g_{2} ,\theta,k))
\end{eqnarray}
where $g_{1}$ and $g_{2}$ are the expression levels of gene $X_{1}$ and
$X_{2}$, respectively.

\emph{6) Logical NOR regulation}\\
A logical OR regulation is defined as if both gene $X_{1}$ and gene $X_{2}$ do not express, gene $Y$ expresses. The mathematical description of the logical NOR regulation is defined as follows:
\begin{eqnarray}
\label{equ:nor}
\frac{dy}{dt} = - y + 1 - sig(g_{1} + g_{2} ,\theta,k))
\end{eqnarray}
where $g_{1}$ and $g_{2}$ are the expression levels of gene $X_{1}$ and
$X_{2}$, respectively.

\emph{7) Logical ANDN regulation}\\
A logical ANDN regulation is defined as if gene $X_{1}$ expresses and gene $X_{2}$ does not express, gene $Y$ expresses,as defined by Eq. \eqref{equ:ANDN_1}. Or if gene $X_{1}$ does not express and gene $X_{2}$ expresses, gene $Y$ expresses, as defined by Eq. \eqref{equ:ANDN_2}.
\begin{eqnarray}
\label{equ:ANDN_1}
\frac{dy}{dt} = - y + sig(g_{1} \ast (1 - g_{2}) ,\theta,k))
\end{eqnarray}
\begin{eqnarray}
\label{equ:ANDN_2}
\frac{dy}{dt} = - y + sig((1 - g_{1}) \ast g_{2} ,\theta,k))
\end{eqnarray}
where $g_{1}$ and $g_{2}$ are the expression levels of gene $X_{1}$ and
$X_{2}$ in Eq. \eqref{equ:ANDN_1} and Eq. \eqref{equ:ANDN_2}, respectively.

\emph{8) Logical ORN regulation}\\
A logical ORN regulation is defined as if gene $X_{1}$ expresses or gene $X_{2}$ does not express, gene $Y$ expresses, as defined by Eq. \eqref{equ:ORN_1}. Or if gene $X_{1}$ does not express or gene $X_{2}$ expresses, gene $Y$ expresses, as defined by Eq. \eqref{equ:ORN_2}.
\begin{eqnarray}
\label{equ:ORN_1}
\frac{dy}{dt} = - y + sig(g_{1} + (1 - g_{2}) ,\theta,k))
\end{eqnarray}
\begin{eqnarray}
\label{equ:ORN_2}
\frac{dy}{dt} = - y + sig((1 - g_{1}) + g_{2} ,\theta,k))
\end{eqnarray}
where $g_{1}$ and $g_{2}$ are the expression levels of gene $X_{1}$ and
$X_{2}$ in Eq. \eqref{equ:ORN_1} and Eq. \eqref{equ:ORN_2}, respectively.

\emph{9) Logical XOR regulation}\\
A logical XOR regulation is defined as iff gene $X_{1}$ and gene $X_{2}$ both express or gene $X_{1}$ and gene $X_{2}$ both do not express, gene $Y$ expresses. The mathematical description of the logical XOR regulation is defined as follows:
\begin{equation}
\label{equ:xor}
\begin{split}
\frac{dy}{dt}=&- y + sig(g_{1} \ast (1 - g_{2}) ,\theta,k))\\
&+ sig((1 - g_{1}) \ast g_{2} ,\theta,k))
\end{split}
\end{equation}
where $g_{1}$ and $g_{2}$ are the expression levels of gene $X_{1}$ and
$X_{2}$, respectively.

\emph{10) Logical XNOR regulation}\\
A logical XNOR regulation is defined as iff one of either gene $X_{1}$ or gene $X_{2}$ express, gene $Y$ expresses. The mathematical description of the logical XNOR regulation is defined as follows:
\begin{equation}
\label{equ:xnor}
\begin{split}
\frac{dy}{dt}=&- y + 1 - sig(g_{1} \ast (1 - g_{2}) ,\theta,k))\\
&- sig((1 - g_{1}) \ast g_{2} ,\theta,k))
\end{split}
\end{equation}
where $g_{1}$ and $g_{2}$ are the expression levels of gene $X_{1}$ and
$X_{2}$, respectively.

When applying MOGP-NSGA-II to evolve GRNs, these ten predefined basic network motifs are employed to constitute the GRNs. Furthermore, a regulatory parameter $\theta$ will be optimised by DE. Here, the scale factor $k$ in each basic network motif is set to 1.

\subsection{Fitness Function}

When applying MOGP-NSGA-II to evolve GRN-based models, another typical task is to define fitness functions. Since MOGP-NSGA-II may create some complex models in evolutionary process. The fitness functions are set to balance the complexity and accuracy of GRN-based models. In addition, literatures \cite{jin2012hierarchical}, \cite{oh2014evolving} suggest environmental restrictions into a fitness function in order to make the swarm pattern can get cross the restricted environment without colliding any obstacles.

In order to explain the GRN-based model, i.e, the fewer nodes these models have, the better the models are. To define the complexity of the model, an important indicator is the number of nodes in the GRN-based model, and this indicator can be regarded as a fitness function, as follow:
\begin{eqnarray}
\label{equ:fitness_one}
f_{1} = node(m_{i})
\end{eqnarray}
where $m_{i}$ is the $i$th GRN-based model and $node(m_{i})$ is the number of nodes of the $i$th model.

For swarm pattern formation, an important task is to encircle targets without colliding them. To satisfy this task, the nearest and furthest distances of swarm robots from targets are set. That is, $d_{min}$ and $d_{max}$ are the allowed minimum and maximum distances between the pattern and the targets, respectively. In this paper, $d_{min}$ and $d_{max}$ are set to 1 and 2, respectively. In addition, another important task is that swarm robots can not collide obstacles in the restricted environment. To satisfy this task, the nearest distance of swarm robots from obstacles is set. That is, $d^{obs}_{min}$ is the allowed minimum distance between the pattern and the obstacles. In this paper, $d^{obs}_{min}$ is set to 2. Hence, the fitness function is set up as follows:
\begin{equation}
\label{equ:fitness_two}
\begin{split}
f_{2} = &\Sigma^{N_{p}}_{i=1}\Sigma^{N_{t}}_{j=1}\frac{sig(d^{ij}_{pt},d_{max},k_{1}) + sig(d_{min},d^{ij}_{pt},k_{2})}{N_{p}N_{t}}\\
&+\Sigma^{N_{p}}_{i=1}\Sigma^{N_{o}}_{k=1}\frac{sig(d^{obs}_{min},d^{ik}_{po},k_{3})}{N_{p}N_{o}}\\
\end{split}
\end{equation}
where $N_{p}$, $N_{t}$ and $N_{o}$ are the number of swarm robots, the number of targets and the number of obstacles, respectively. In addition, $d^{ij}_{pt}$ is the distance from the $i$th swarm robot to the $j$th target. Similarly, $d^{ik}_{po}$ is the distance from the $i$th swarm robot to the $k$th obstacle.

\subsection{Automatic Design Framework Embedded in GRN-based Model}

The proposed automatic design framework of GRN-based model creates ten basic network motifs according to the interaction of gens in cell-cell. Then this framework employs the automatic design method to design the GRN-based model automatically according to the requirements of the scene. Finally, an MOGP-NSGA-II and DE are applied to optimize the structures and parameters of the GRN-based model in parallel, as so to get an optimal GRN-based model. The main difference between the proposed framework and the manual design framework is that the proposed framework does not need to predefine different GRN-based model structures under various environments, which solves the limitation of GRNs for swarm pattern formations. In other words, the proposed framework applies MOGP-NSGA-II and DE to optimize the automatic design GRN-based model in order that the optimized model can cross the restricted environment without colliding obstacles.

\begin{figure*}[!htp]
\centering
\includegraphics[width=4.5in]{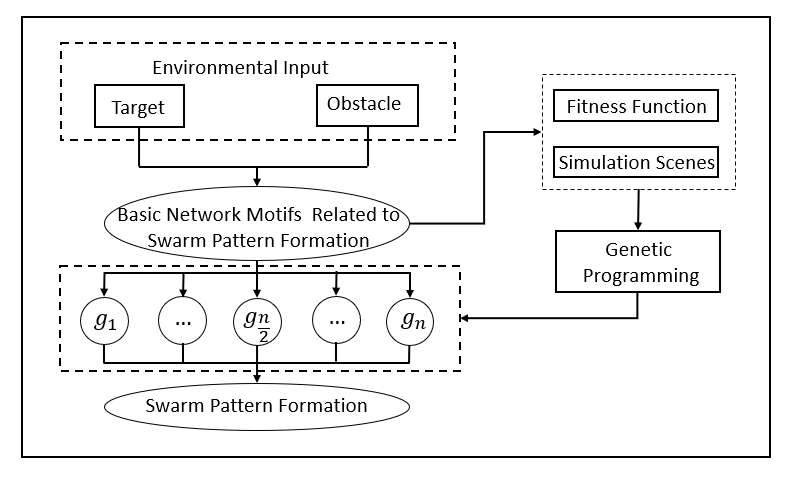}%
\hfil
\caption{An automatic design framework of swarm pattern formation on MOGP.}
\label{fig:Evo_design}
\end{figure*}

The General structure of the proposed automatic design framework is illustrated in Fig. \ref{fig:Evo_design}, which is divided into three stages. The three stages are the input stage, automatic generation of initialization model stage and the optimization stage. In the input stage, the location information of targets and obstacles is translated into an integrated morphogen gradient space. It is noticeable that this transformation behavior is activated when mother robots detect targets or obstacles. This integrated morphogen gradient space is served as the input to the proposed framework. In the automatic generation of initialization model stage, GRN-based models based on swarm pattern information are automatically generated by an integrated morphogen gradient space and ten basic network motifs. The automatically generated GRN-based models are constituted by activate genes $g_{1}, g_{2},...,g_{n}$. Once the GRN-based models are generated, they will function as the input of the optimization stage to trigger their optimization. MOGP-NSGA-II and DE are employed in the optimization stage, MOGP-NSGA-II is responsible for structural optimization of GRN-based models and DE is responsible for parameter optimization of each GRN-based model. Through the optimization process, the automatically generated GRN-based models are optimized to the optimal models satisfying the restricted environment.

The framework of the proposed algorithm is introduced in Algorithm \ref{alg:GP-NSGA-II}.In Algorithm \ref{alg:GP-NSGA-II}, the algorithm is initialized at lines 1-3. At line 1, a initial population $P_0$ is created by using the ramped half-and-half method. At line 2, the regulatory parameters of each individual in the initial population are optimized by DE. At line 3, the fitness functions are applied to evaluate each individual in the initial population. At line 6, the offspring population $Q_{g}$ is created by using the crossover, mutation and reproduction of MOGP-NSGA-II. At lines 7-10, the regulatory parameters of each individual $q$ in the offspring population are optimized by using DE, and the fitness values of $q$ is obtained by using the fitness functions. At line 12, the new population $P_{g+1}$ is selected by the NSGA-II selection. At line 13, the generation counter is updated. At line 15, a set of non-dominated and feasible solutions is selected.

\begin{algorithm}
    \KwIn{
    	\begin{enumerate}
    	\item[] $G$: the integrated morphogen gradient space;
        \item[] $gen_{max}$: the maximum generation.
    	\end{enumerate}
    }
    \KwOut{a set of non-dominated and feasible solutions.}

    Initialize: Randomly create an initial population $P_{0}$ of GRN-base models form ramped half-and-half method;\\
    Optimal Parameters: the parameters of these models are optimized by DE;\\
    Evaluate: evaluate the population $P_{0}$ by fitness functions;\\
    Set $gen = 0$;\\
    \While{$gen \le gen_{max}$}{
    	Generate the offspring population $Q_{g}$ by applying genetic operations;\\
        \ForEach{$q \in Q_{g}$}{
               the parameters of $q$ are optimized by DE; // $q$ is an individual in $Q_{g}$;\\
               evaluate each individual $q$ in $Q_{g}$ using fitness functions;\\
        }
        $R_{g} \leftarrow P_{g} \cup Q_{g}$; // $P_{g}$ is $g$th parent population;\\
        Form the new population $P_{g+1}$ from $R_{g}$ by the NSGA-II selection;\\
		$gen=gen+1$;\\
	}
    Output the non-dominated and feasible solutions.
\caption{The framework of MOGP-NSGA-II for swarm pattern formation}
\label{alg:GP-NSGA-II}
\end{algorithm}

\section{Numerical Simulations}
\label{sec:num_simu}

In this section, we will evaluate the proposed automatic design framework of swarm pattern formation based on multi-objective genetic programming by numerical simulations. First, the proposed framework forms a swarm pattern to encircle one or more targets in a channel, which verifies the feasibility of the framework. Second, swarm patterns are formed to verify the adaptability of the proposed framework in a compound channel. Finally, swarm patterns that are formed in the compound channel migrate directly to an environment where obstacles are randomly distributed, so as to verify the transferability of the proposed framework.

\subsection{Swarm Pattern Formation in a channel}

\begin{figure}[!t]
\centering
\includegraphics[width=2.5in]{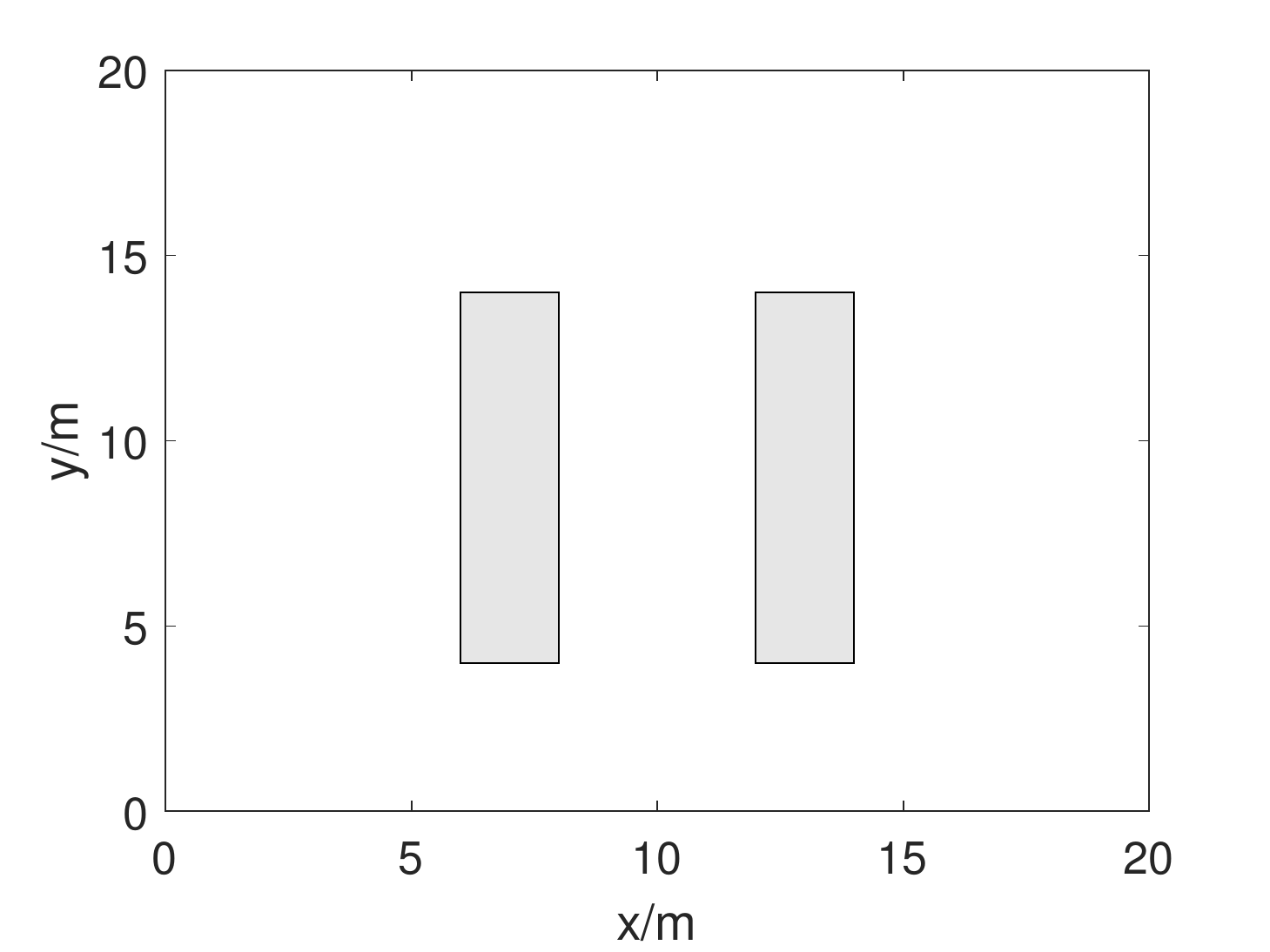}%
\hfil
\caption{A channel is in a region of 20 by 20 meters, and green shadows represent the channel.}
\label{fig:task_one}
\end{figure}

To demonstrate the feasibility of the proposed automatic framework, the proposed framework and an evolving hierarchical gene regulatory network (EH-GRN) \cite{oh2014evolving} are compared in a simply simulated area. The simply simulated area is that a swarm pattern encircles one or more targets without colliding a channel, and the channel is in a region of 20 by 20 meters, as shown in Fig. \ref{fig:task_one}. An evaluation criterion is that swarm pattern can not collide with the channel while encircling all targets.

In evolutionary process, the detailed parameters are listed as follows:
\begin{enumerate}
\item[1.] For the proposed automatic framework, the regulatory parameter $\theta$ of each basic network motif is initialized randomly between 0 to 2. During optimization, $\theta$ of each basic network motif is optimized by DE.
\item[2.] For EH-GRN, the regulatory parameter $\theta$ of each basic network motif is optimized by the covariance matrix adaptation evolution strategy \cite{hansen2001completely}, and the regulatory parameter $\theta$ ranges from 0 to 2.
\item[3.] For the proposed automatic framework, the population size for structure optimization is 40, and the population size for parameters optimization is 10. For EH-GRN, the population size is set to 40.
\item[4.] For the proposed automatic framework, the maximum depth of tree is 4, and the minimum of depth of tree is 1.
\item[5.] The crossover rate of MOGP-NSGA-II and DE are 1.0 and 0.9, respectively. And the mutation rate of MOGP-NSGA-II and DE are 0.1 and 0.5, respectively.
\item[6.] The evaluation number of the proposed automatic framework and EH-GRN are both 4000.
\end{enumerate}
%As described above, the number of function evaluation is 4000.

\begin{figure}[!t]
\centering
\includegraphics[width=3in]{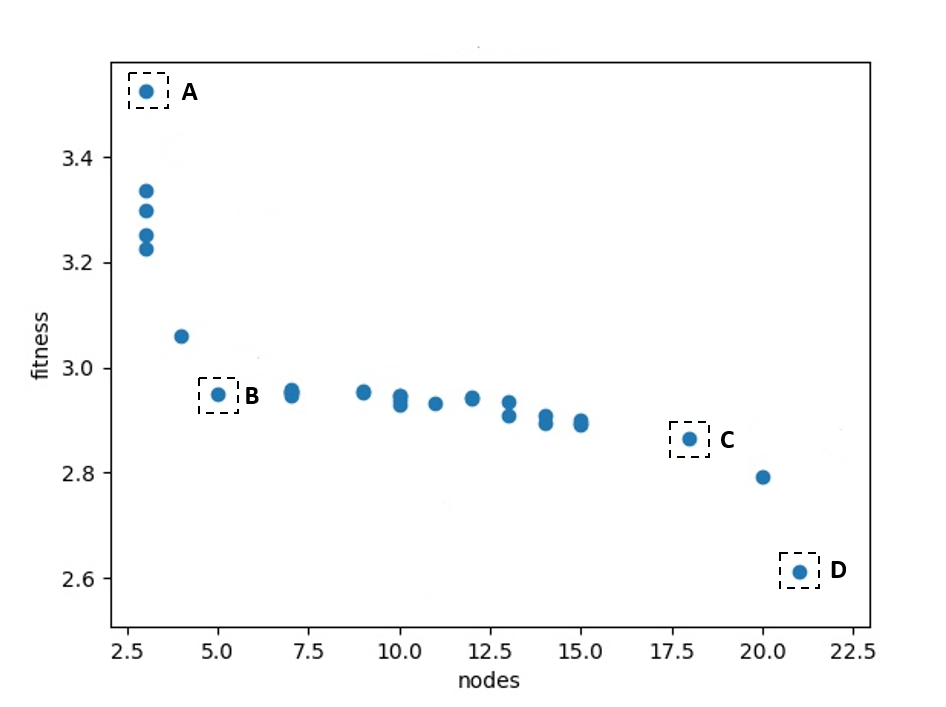}%
\hfil
\caption{The non-dominated solutions are achieved by MOGP-NSGA-II. The A, B, C and D are selected four points. The point A and D have two extreme cases. The point B and C are called knee points. }
\label{fig:task1_PF_solution}
\end{figure}

\begin{figure}[!t]
\centering
\includegraphics[width=2in]{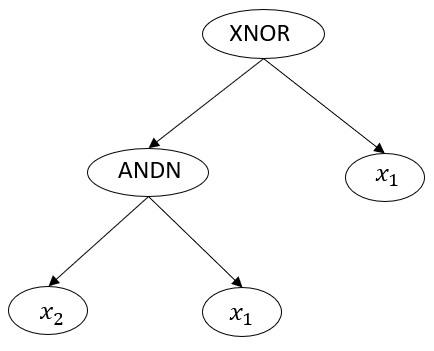}%
\hfil
\caption{The syntax tree of point B is achieved by using MOGP-NSGA-II. $x_{1}$ is that the location information of target forms a morphogen gradient space. $x_{2}$ is that the location information of obstacle forms a morphogen gradient space. $NAND$ and $XNOR$ are two basic network motifs.}
\label{fig:task1_tree}
\end{figure}

\begin{figure}[!t]
\centering
\includegraphics[width=2in]{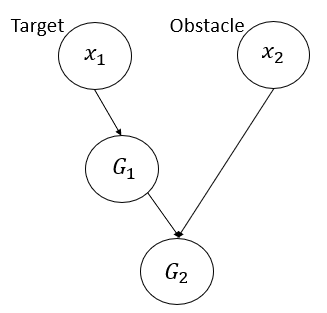}%
\hfil
\caption{In the considered scenario, the automatic design framework is optimized to get GRN-based model. $x_{1}$ is that the location information of target forms a morphogen gradient space. $x_{2}$ is that the location information of obstacle forms a morphogen gradient space. $G_{1}$ and $G_{2}$ are activate genes, and the concentration of $G_{2}$ represent a morphogen gradient space to form the desired swarm pattern.}
\label{fig:task1_GRN}
\end{figure}

In Fig. \ref{fig:task1_PF_solution}, the non-dominated solutions are achieved by MOGP-NSGA-II when the evaluation number reaches 4000. Fig. \ref{fig:task1_PF_solution} shows that the lower the number of nodes, the higher the fitness value. In other words, the number of nodes at point A is 3, but it has a high fitness value. Although the fitness value of point D is the smallest among all non-dominated solutions, the structure of point D is the most complex. To balance the complexity and accuracy of GRN-based models without a priori knowledge, point B and C are selected by using \cite{branke2004finding}, they are called knee points. Furthermore, the fitness values of point B and point C are very close, and the structure of point B is simpler than that of point C. Thus, point B is selected in the task. Fig. \ref{fig:task1_tree} shows the syntax tree of the point B, and Fig. \ref{fig:task1_GRN} shows the GRN structure of the point B.

In addition, $x_{1}$ and $x_{2}$ are environmental inputs, that is, $x_{1}$ is that the location information of target forms a morphogen gradient space, and $x_{2}$ represents a morphogen concentration that is produced from obstacles. $NAND$ and $XNOR$ are two basic network motifs. For point B, the mathematical description of the syntax tree is as follows:
\begin{equation}
\label{equ:tree_1}
\begin{split}
\frac{dy_{1}}{dt} = &- y_{1} + 1 - sig(x_{1} \ast x_{1} ,\theta_{1},k)) \\
\frac{dy_{2}}{dt} = & - y_{2} + 1 - sig(y_{1} \ast (1 - x_{2}) ,\theta_{2},k))\\
& - sig((1 - y_{1}) \ast x_{2} ,\theta_{2},k))
\end{split}
\end{equation}
%\begin{eqnarray}
%\label{equ:tree_1}
%\frac{dy_{1}}{dt} = - y_{1} + 1 - sig(x_{1} \ast x_{1} ,\theta_{1},k))
%\end{eqnarray}
%\begin{equation}
%\label{equ:tree_2}
%\begin{split}
%\frac{dy_{2}}{dt} = & - y_{2} + 1 - sig(y_{1} \ast (1 - x_{2}) ,\theta_{2},k))\\
%& - sig((1 - y_{1}) \ast x_{2} ,\theta_{2},k))
%\end{split}
%\end{equation}
where $\theta_{1}$ and $\theta_{2}$ are 0.8393 and 0.9256, respectively. Eq.\eqref{equ:tree_1} is a mathematical description of $NAND$ and $XNOR$. $y_{1}$ and $y_{2}$ are morphogen concentrations, where $y_{2}$ is a morphogen gradient that defines the swarm pattern.

\begin{figure}[!t]
\centering
\includegraphics[width=2.5in]{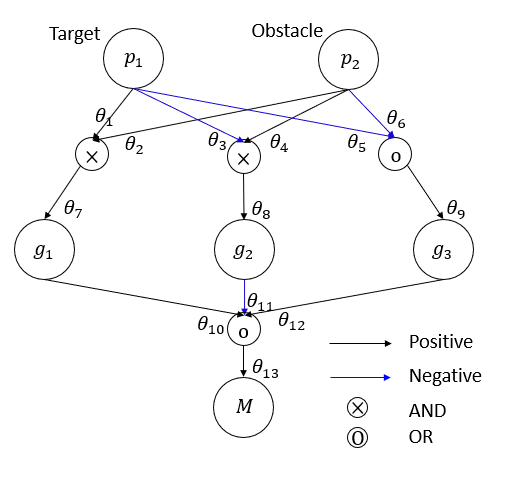}%
\hfil
\caption{Illustration of an EH-GRN structure for swarm pattern formation. The structure of model is predefined, and CMA-ES is applied to optimise the regulatory parameters. $g_{1}$, $g_{2}$ and $g_{3}$ are activate genes. The concentration of $M$ represent a morphogen gradient space to form the desired swarm pattern.}
\label{fig:task1_EH}
\end{figure}

To measure the performance of the proposed automatic design framework, a state-of-the-art method, namely EH-GRN, is tested in the scenario. Fig. \ref{fig:task1_EH} shows an EH-GRN structure for swarm pattern formation, and the structure is optimized to achieve by CMA-ES. In addition, $p_{1}$ and $p_{2}$ represent a morphogen concentration produced by the environmental inputs in Fig. \ref{fig:task1_EH}. In other words, $p_{1}$ and $p_{2}$ represent the morphogen concentrations produced by the environmental inputs. $g_{1}$, $g_{2}$ and $g_{3}$ are activate genes. The concentration of $M$ represent a morphogen gradient space to form the desired swarm pattern. $\theta_{i}$ ($i=1,...,13$) is a regulatory parameter.

For an EH-GRN structure for swarm pattern formation in the considered scenario, each swarm robots follows the following dynamic equations to generate a morphogen gradient space that define the swarm pattern.
\begin{align}
& \frac{dy_{1}}{dt} = - y_{1} + 1 - sig(p_{1},\theta_{1},k))  \label{eq:HGRN_1} \\
& \frac{dy_{2}}{dt} = - y_{2} + 1 - sig(p_{2},\theta_{2},k))  \label{eq:HGRN_2} \\
& \frac{dg_{1}}{dt} = - g_{1} + sig(y_{1} \ast y_{2},\theta_{7},k)) \label{eq:HGRN_3} \\
& \frac{dy_{3}}{dt} = - y_{3} + sig(p_{1},\theta_{3},k)) \label{eq:HGRN_4} \\
& \frac{dy_{4}}{dt} = - y_{4} + sig(p_{2},\theta_{4},k)) \label{eq:HGRN_5} \\
& \frac{dg_{2}}{dt} = - g_{2} + sig(y_{3} + y_{4},\theta_{8},k)) \label{eq:HGRN_6} \\
& \frac{dy_{5}}{dt} = - y_{5} + sig(p_{1},\theta_{5},k))  \label{eq:HGRN_7} \\
& \frac{dy_{6}}{dt} = - y_{6} + sig(p_{2},\theta_{6},k)) \label{eq:HGRN_8} \\
& \frac{dg_{3}}{dt} = - g_{3} + sig(y_{5} \ast y_{6},\theta_{9},k)) \label{eq:HGRN_9} \\
& \frac{dy_{7}}{dt} = - y_{7} + 1 - sig(g_{1},\theta_{10},k)) \label{eq:HGRN_10}
\end{align}
\begin{align}
& \frac{dy_{8}}{dt} = - y_{8} + sig(g_{2},\theta_{11},k)) \label{eq:HGRN_11} \\
& \frac{dy_{9}}{dt} = - y_{9} + sig(g_{3},\theta_{12},k)) \label{eq:HGRN_12} \\
& \frac{dM}{dt} = - M + sig(y_{7} + y_{8} + y_{9},\theta_{13},k)) \label{eq:HGRN_13}
\end{align}

%\begin{eqnarray}
%\label{equ:HGRN_1}
%\frac{dy_{1}}{dt} = - y_{1} + 1 - sig(p_{1},\theta_{1},k))
%\end{eqnarray}
%\begin{eqnarray}
%\label{equ:HGRN_2}
%\frac{dy_{2}}{dt} = - y_{2} + 1 - sig(p_{2},\theta_{2},k))
%\end{eqnarray}
%\begin{eqnarray}
%\label{equ:HGRN_3}
%\frac{dg_{1}}{dt} = - g_{1} + sig(y_{1} \ast y_{2},\theta_{7},k))
%\end{eqnarray}
%\begin{eqnarray}
%\label{equ:HGRN_4}
%\frac{dy_{3}}{dt} = - y_{3} + sig(p_{1},\theta_{3},k))
%\end{eqnarray}
%\begin{eqnarray}
%\label{equ:HGRN_5}
%\frac{dy_{4}}{dt} = - y_{4} + sig(p_{2},\theta_{4},k))
%\end{eqnarray}
%\begin{eqnarray}
%\label{equ:HGRN_6}
%\frac{dg_{2}}{dt} = - g_{2} + sig(y_{3} + y_{4},\theta_{8},k))
%\end{eqnarray}
%\begin{eqnarray}
%\label{equ:HGRN_7}
%\frac{dy_{5}}{dt} = - y_{5} + sig(p_{1},\theta_{5},k))
%\end{eqnarray}
%\begin{eqnarray}
%\label{equ:HGRN_8}
%\frac{dy_{6}}{dt} = - y_{6} + sig(p_{2},\theta_{6},k))
%\end{eqnarray}
%\begin{eqnarray}
%\label{equ:HGRN_9}
%\frac{dg_{3}}{dt} = - g_{3} + sig(y_{5} \ast y_{6},\theta_{9},k))
%\end{eqnarray}
%\begin{eqnarray}
%\label{equ:HGRN_10}
%\frac{dy_{7}}{dt} = - y_{7} + 1 - sig(g_{1},\theta_{10},k))
%\end{eqnarray}
%\begin{eqnarray}
%\label{equ:HGRN_11}
%\frac{dy_{8}}{dt} = - y_{8} + sig(g_{2},\theta_{11},k))
%\end{eqnarray}
%\begin{eqnarray}
%\label{equ:HGRN_12}
%\frac{dy_{9}}{dt} = - y_{9} + sig(g_{3},\theta_{12},k))
%\end{eqnarray}
%\begin{eqnarray}
%\label{equ:HGRN_13}
%\frac{dM}{dt} = - M + sig(y_{7} + y_{8} + y_{9},\theta_{13},k))
%\end{eqnarray}
where the following parameter values: $\theta_{1} = 1$, $\theta_{2} = 0.5328$, $\theta_{3} = 1$, $\theta_{4} = 0.4448$, $\theta_{5} = 0$, $\theta_{6} = 0.934$, $\theta_{7} = 2$, $\theta_{8} = 1.2095$, $\theta_{9} = 1.6798$, $\theta_{10} = 1$, $\theta_{11} = 0.5385$, $\theta_{12} = 0.2763$, $\theta_{13} = 1.3445$.

\begin{figure*}[ht]
	\begin{tabular}{cc}
		\begin{minipage}[t]{0.32\linewidth}  %  width = 4.5cm,height = 3.6cm
			\includegraphics[width = 6.8cm, height = 5.8cm]{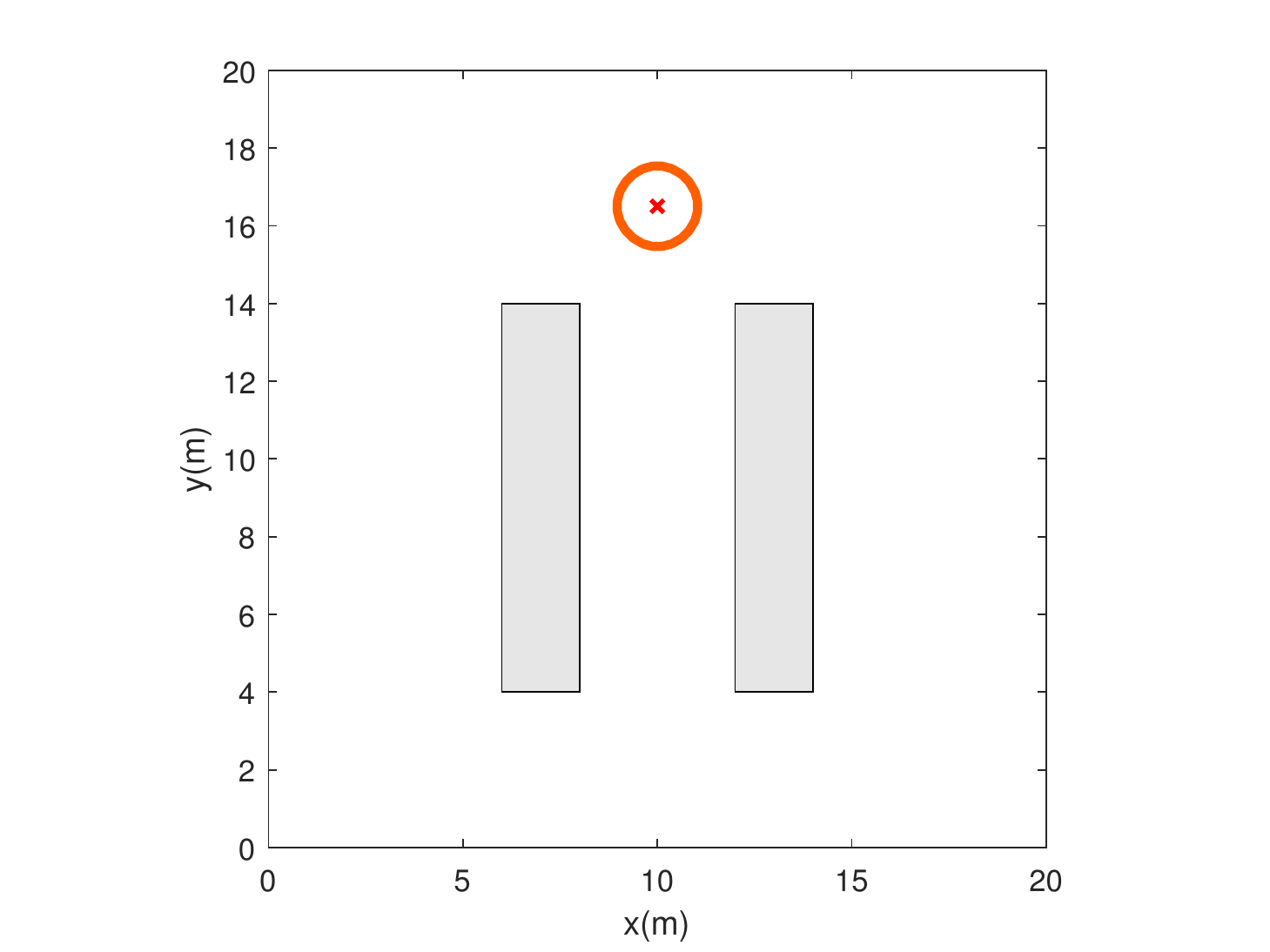}\\
			\centering{\scriptsize{(a)}}
		\end{minipage}
        \begin{minipage}[t]{0.32\linewidth}
			\includegraphics[width = 6.8cm, height = 5.8cm]{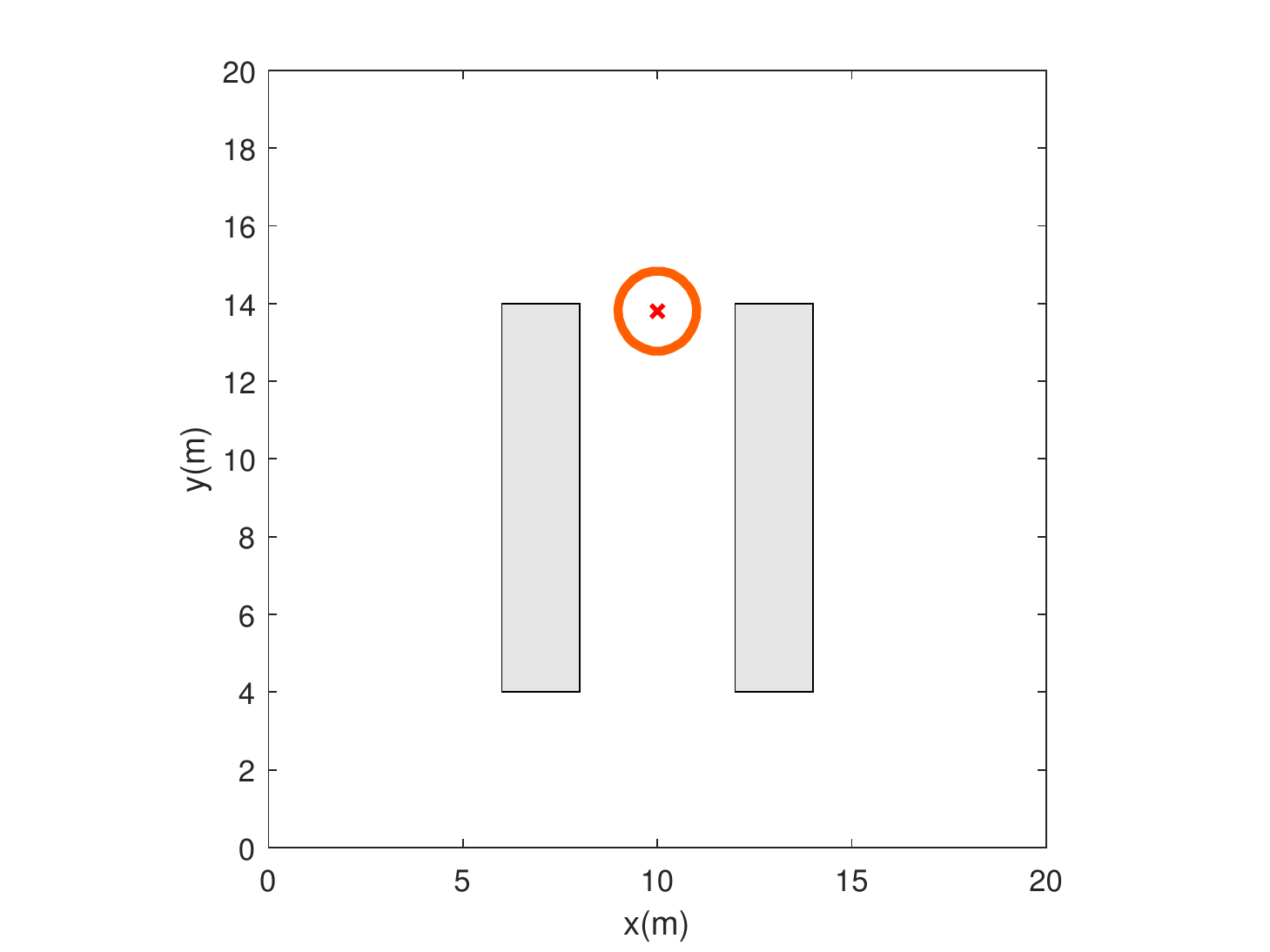}\\
			\centering{\scriptsize{(b)}}
		\end{minipage}
        \begin{minipage}[t]{0.32\linewidth}
			\includegraphics[width = 6.8cm, height = 5.8cm]{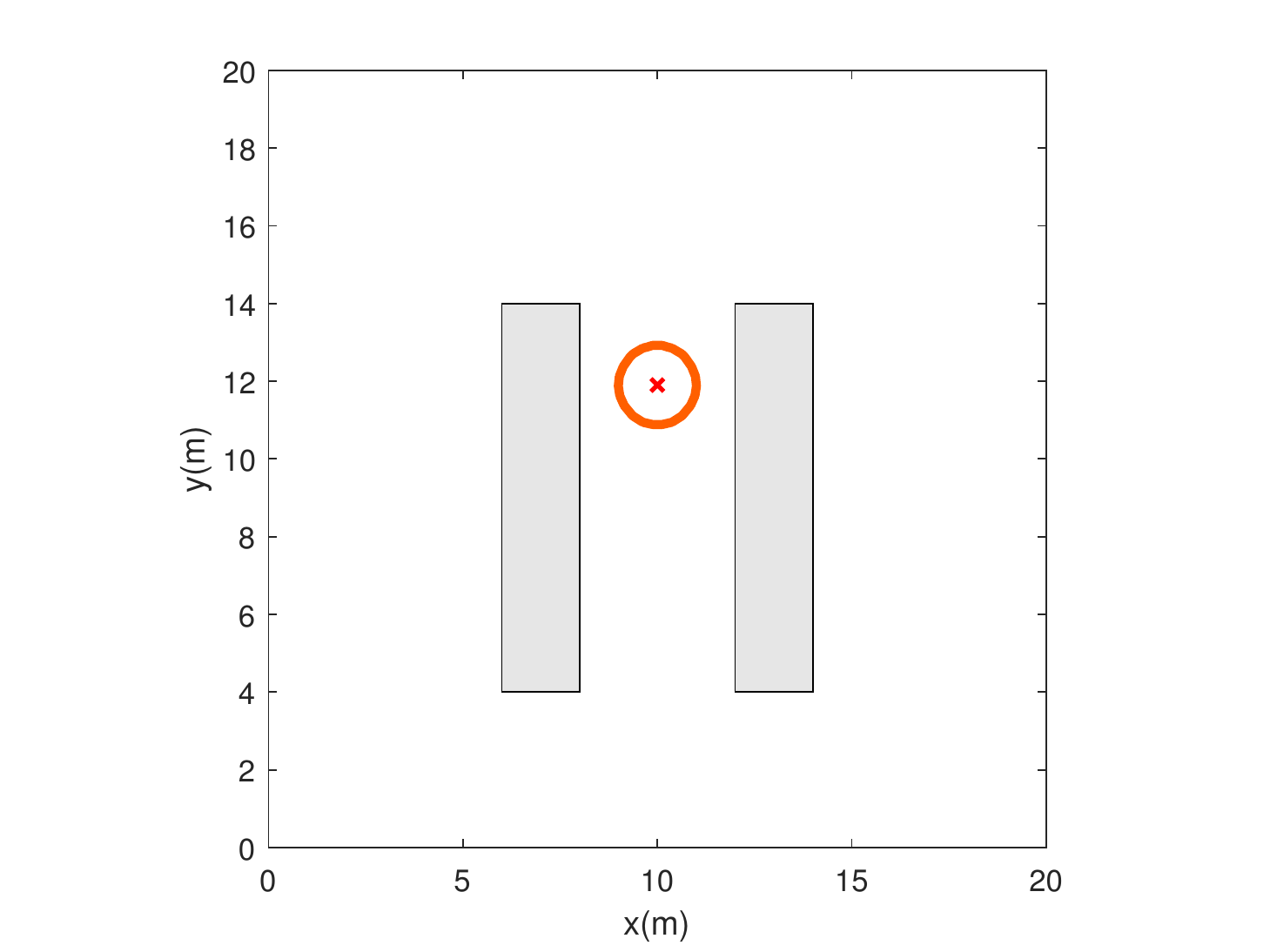}\\
			\centering{\scriptsize{(c)}}
		\end{minipage}
	\end{tabular}
	
    \begin{tabular}{cc}	
        \begin{minipage}[t]{0.32\linewidth}
			\includegraphics[width = 6.8cm, height = 5.8cm]{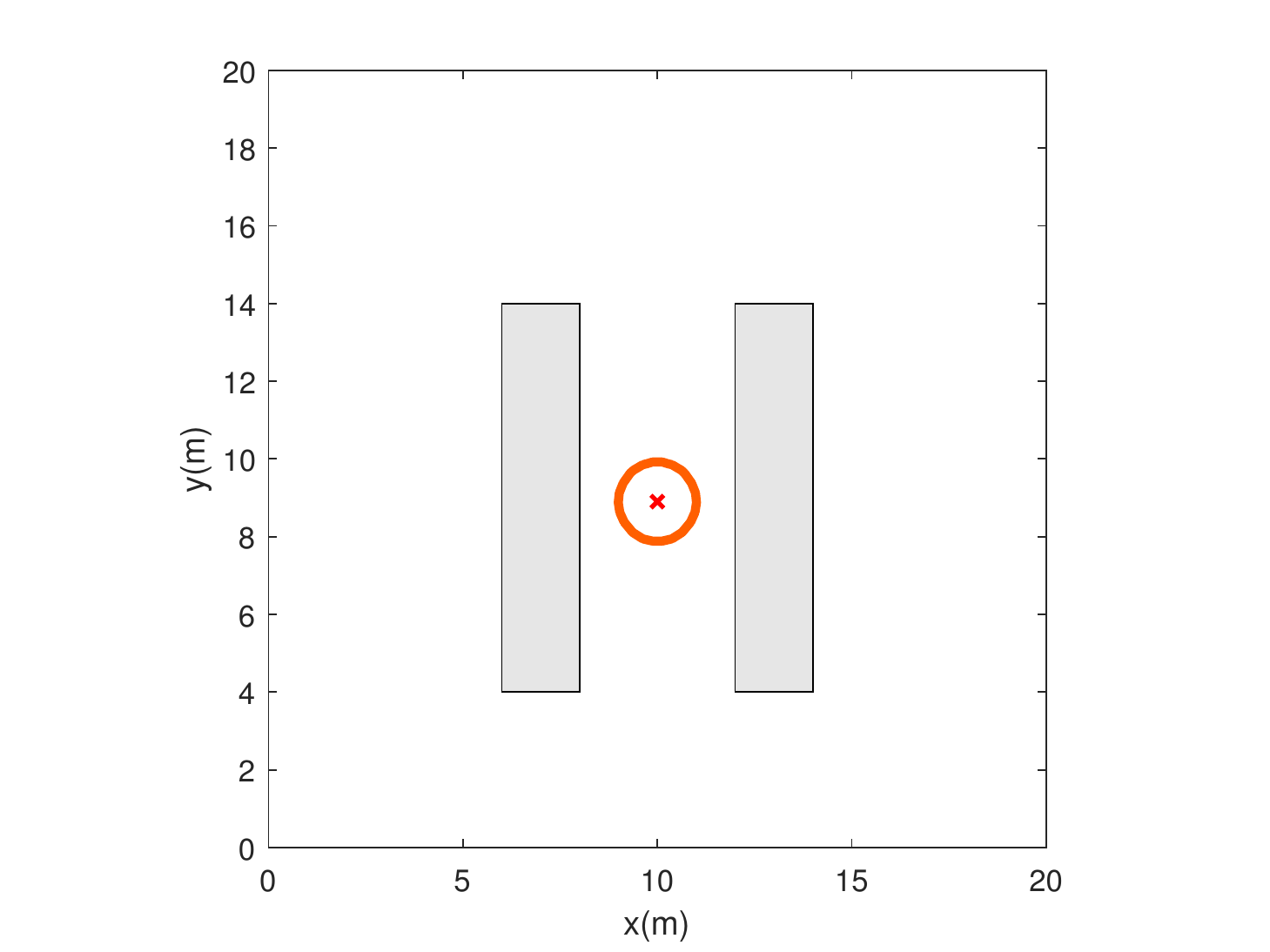}\\
			\centering{\scriptsize{(d)}}
		\end{minipage}
		\begin{minipage}[t]{0.32\linewidth}
			\includegraphics[width = 6.8cm, height = 5.8cm]{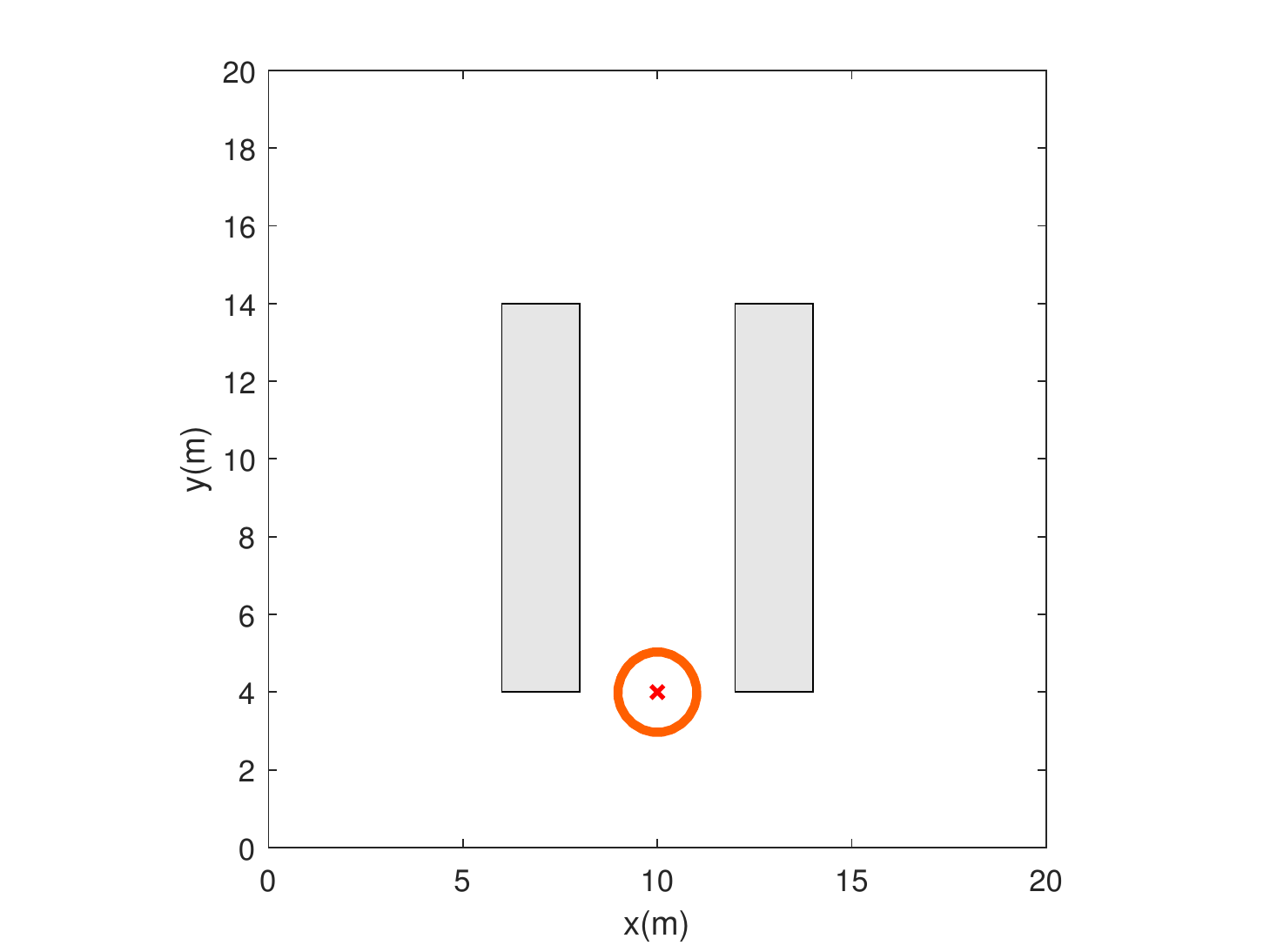}\\
			\centering{\scriptsize{(e)}}
		\end{minipage}
        \begin{minipage}[t]{0.32\linewidth}
			\includegraphics[width = 6.8cm, height = 5.8cm]{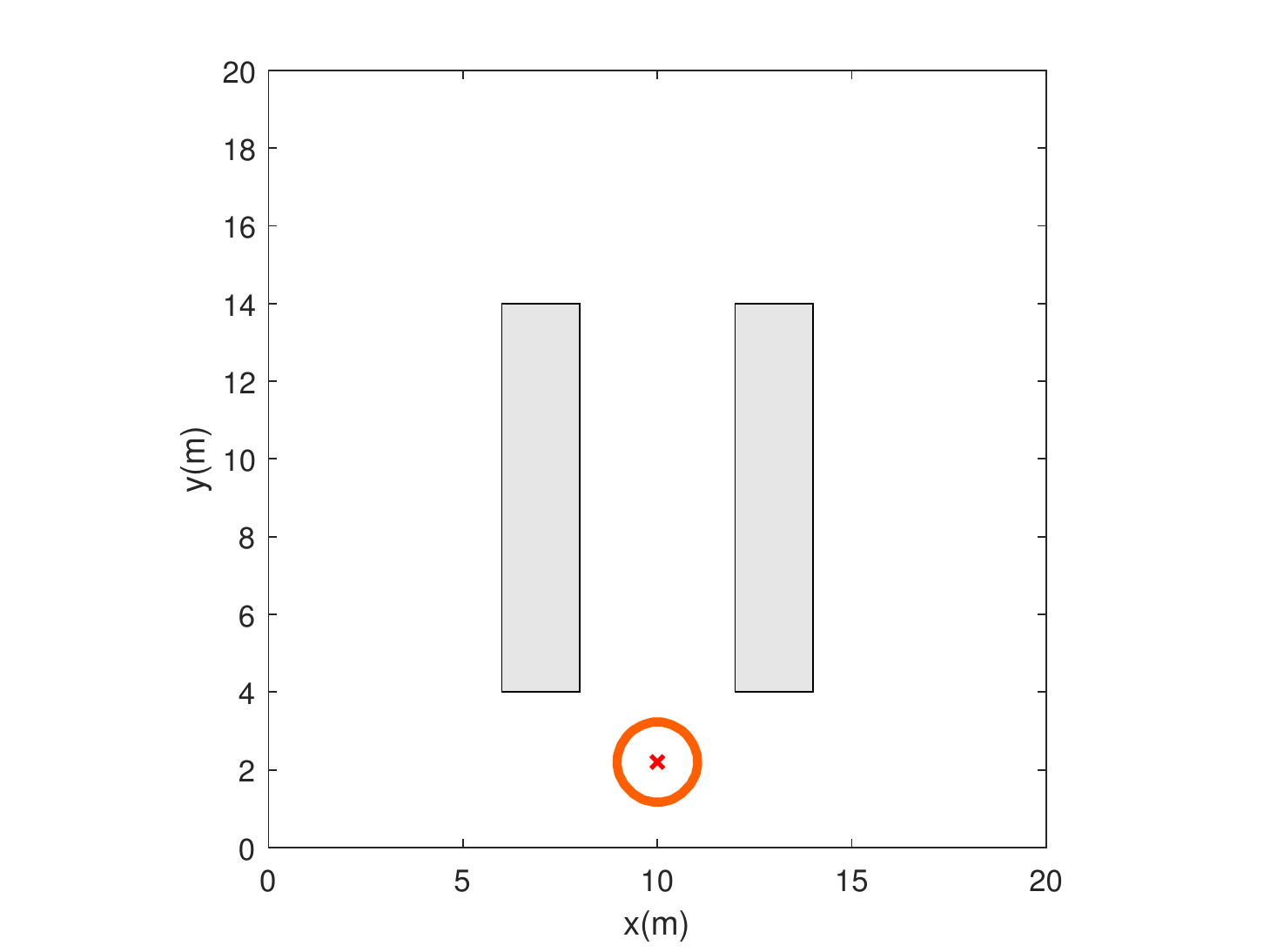}\\
			\centering{\scriptsize{(f)}}
		\end{minipage}
	\end{tabular}
	\caption{\label{fig:case1_GP_one} The swarm pattern that optimized to achieve by the proposed automatic design framework encircles a target across a channel without colliding the channel. (a)-(f) show that the swarm pattern encircle the target in a circle shape way, when the target passes through the channel.}
\end{figure*}

Fig. \ref{fig:case1_GP_one} shows that the swarm pattern that optimized to achieve by the proposed automatic design framework encircles a target across a channel without colliding the channel. In particular, when the target does not enter the channel, the swarm pattern is a circular shape way to encircle the target, as illustrated by Fig. \ref{fig:case1_GP_one} (a) and (f). This is because the mother robots do not detect obstacles. The concentration of swarm pattern is only affected by the target input. Fig. \ref{fig:case1_GP_one} (b)-(d) show how the target passes through the channel. The swarm pattern still encircle the target in a circle shape way. This is because the width of the channel is larger than or equal to the furthest distances of swarm robots from the target. That is, the swarm pattern is not affected by the channel.

\begin{figure*}[ht]
	\begin{tabular}{cc}
		\begin{minipage}[t]{0.32\linewidth}  %  width = 4.5cm,height = 3.6cm
			\includegraphics[width = 6.8cm, height = 5.8cm]{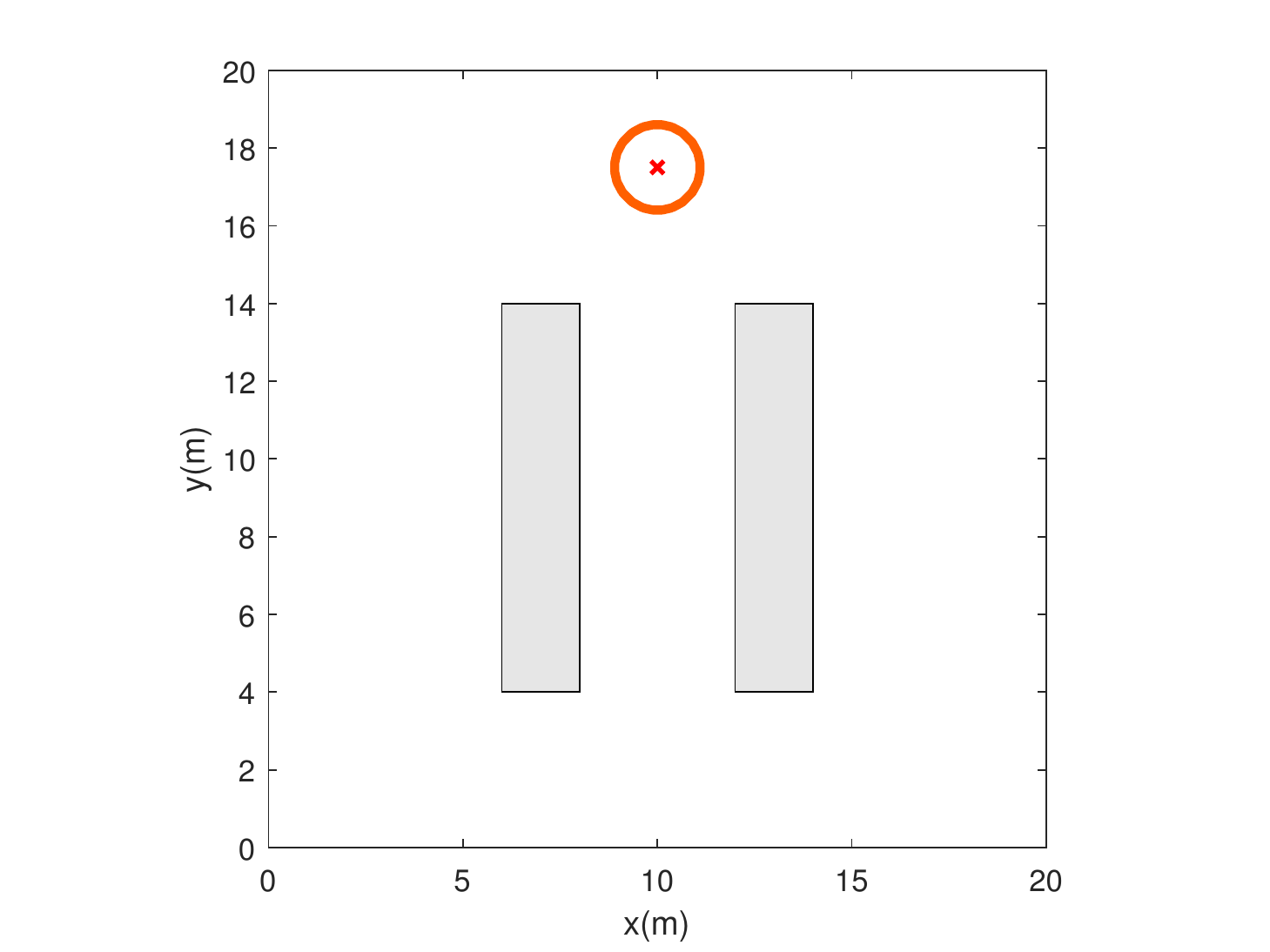}\\
			\centering{\scriptsize{(a)}}
		\end{minipage}
        \begin{minipage}[t]{0.32\linewidth}
			\includegraphics[width = 6.8cm, height = 5.8cm]{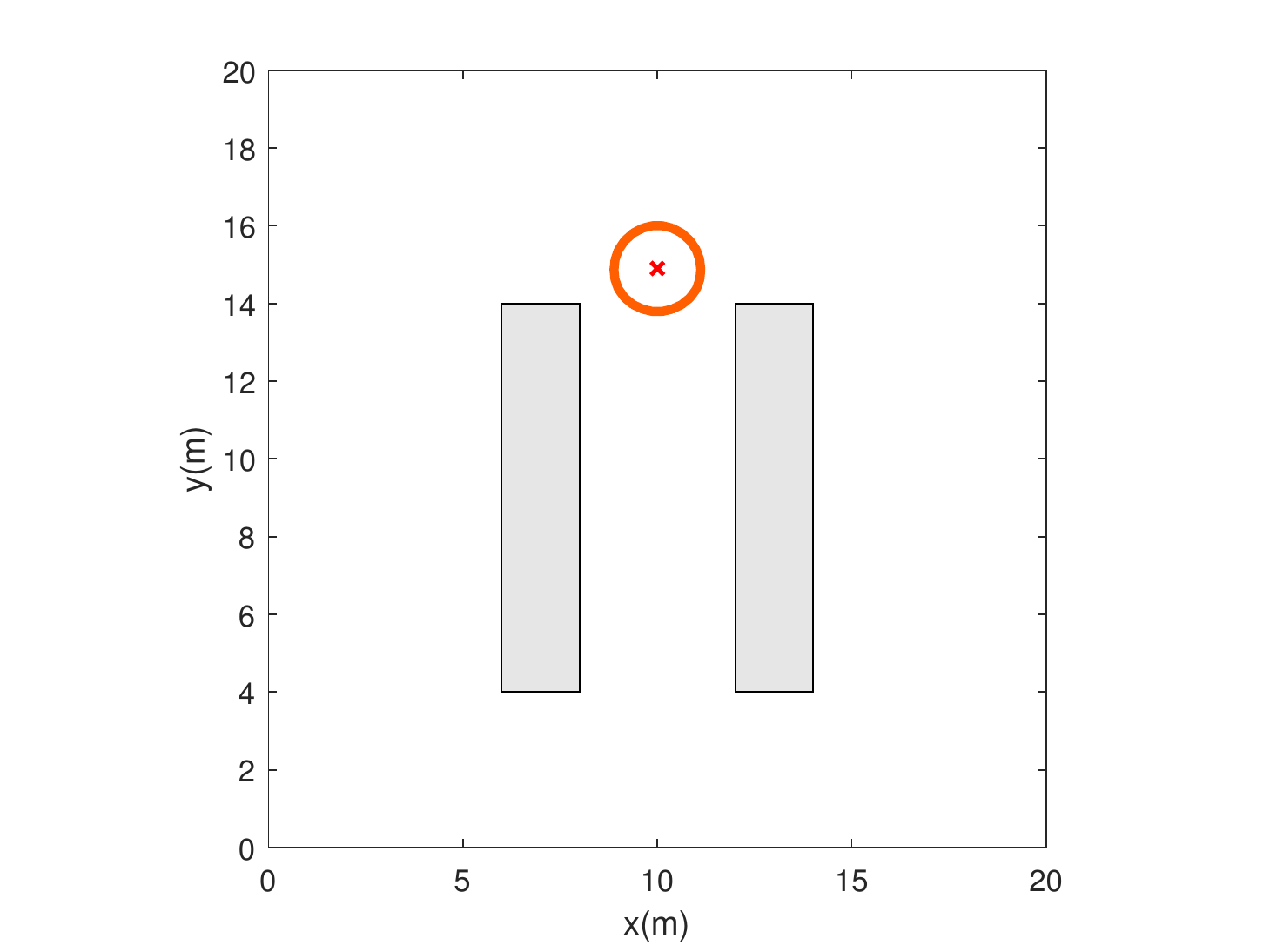}\\
			\centering{\scriptsize{(b)}}
		\end{minipage}
        \begin{minipage}[t]{0.32\linewidth}
			\includegraphics[width = 6.8cm, height = 5.8cm]{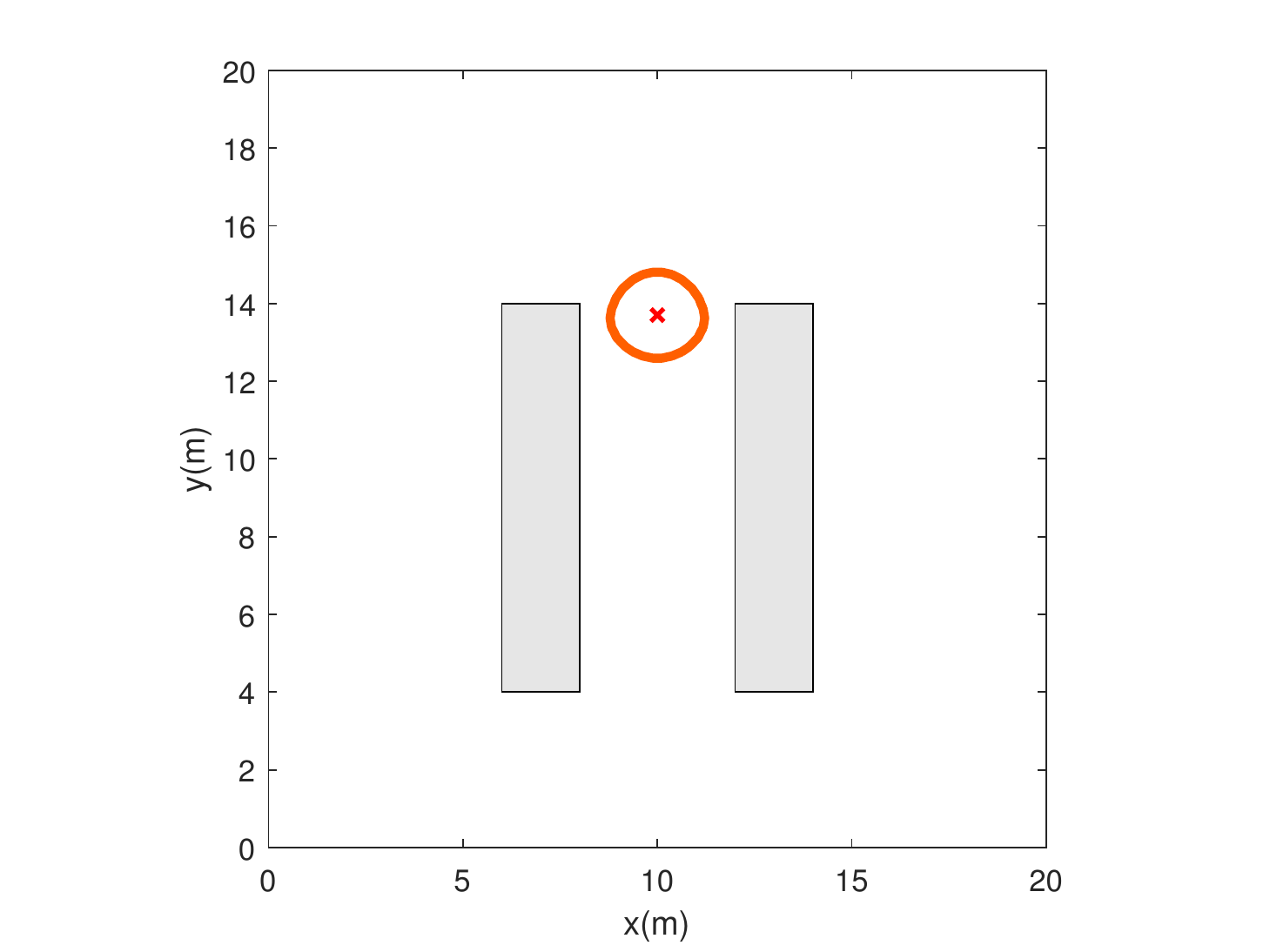}\\
			\centering{\scriptsize{(c)}}
		\end{minipage}
	\end{tabular}
	
    \begin{tabular}{cc}	
        \begin{minipage}[t]{0.32\linewidth}
			\includegraphics[width = 6.8cm, height = 5.8cm]{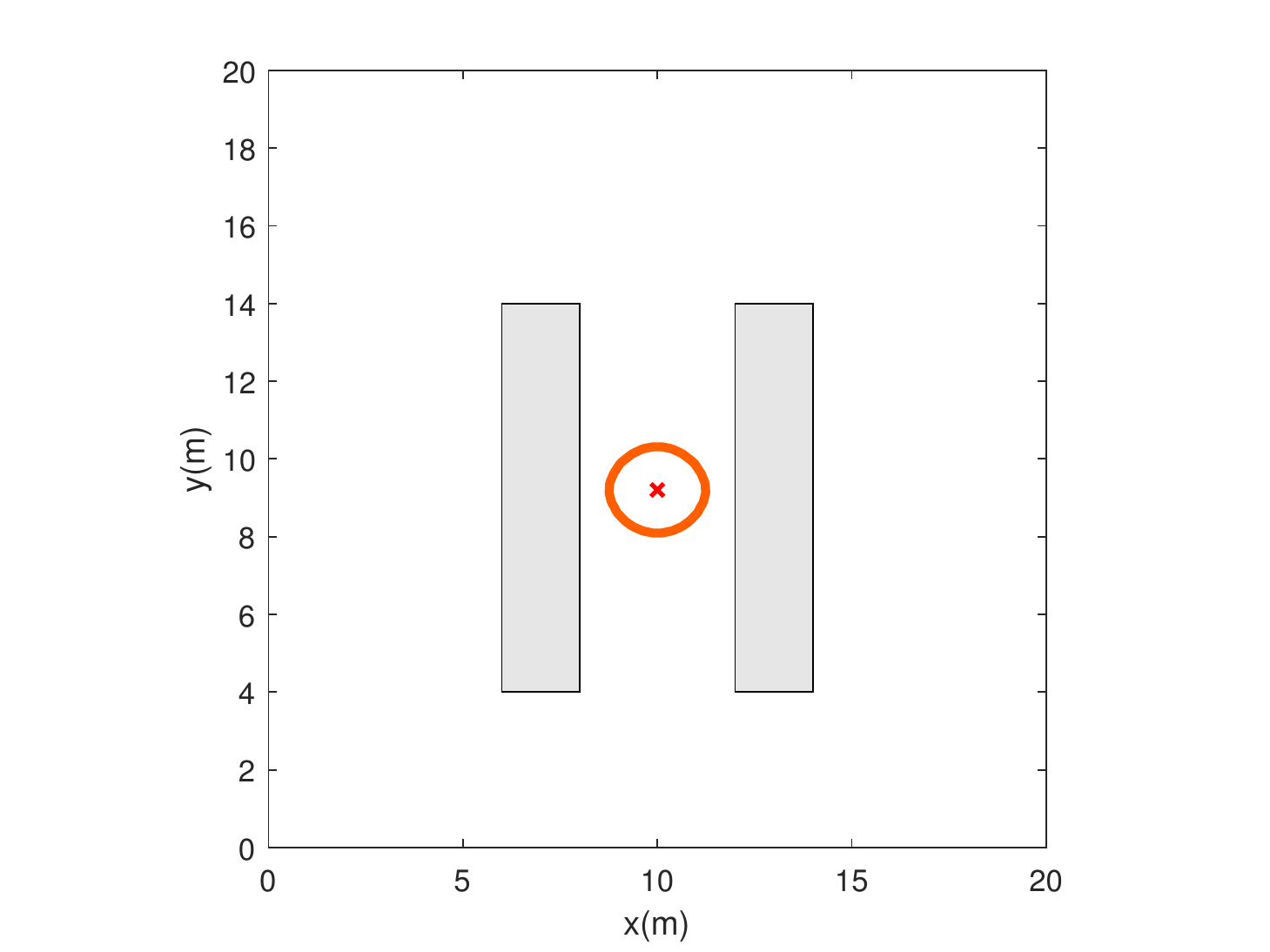}\\
			\centering{\scriptsize{(d)}}
		\end{minipage}
		\begin{minipage}[t]{0.32\linewidth}
			\includegraphics[width = 6.8cm, height = 5.8cm]{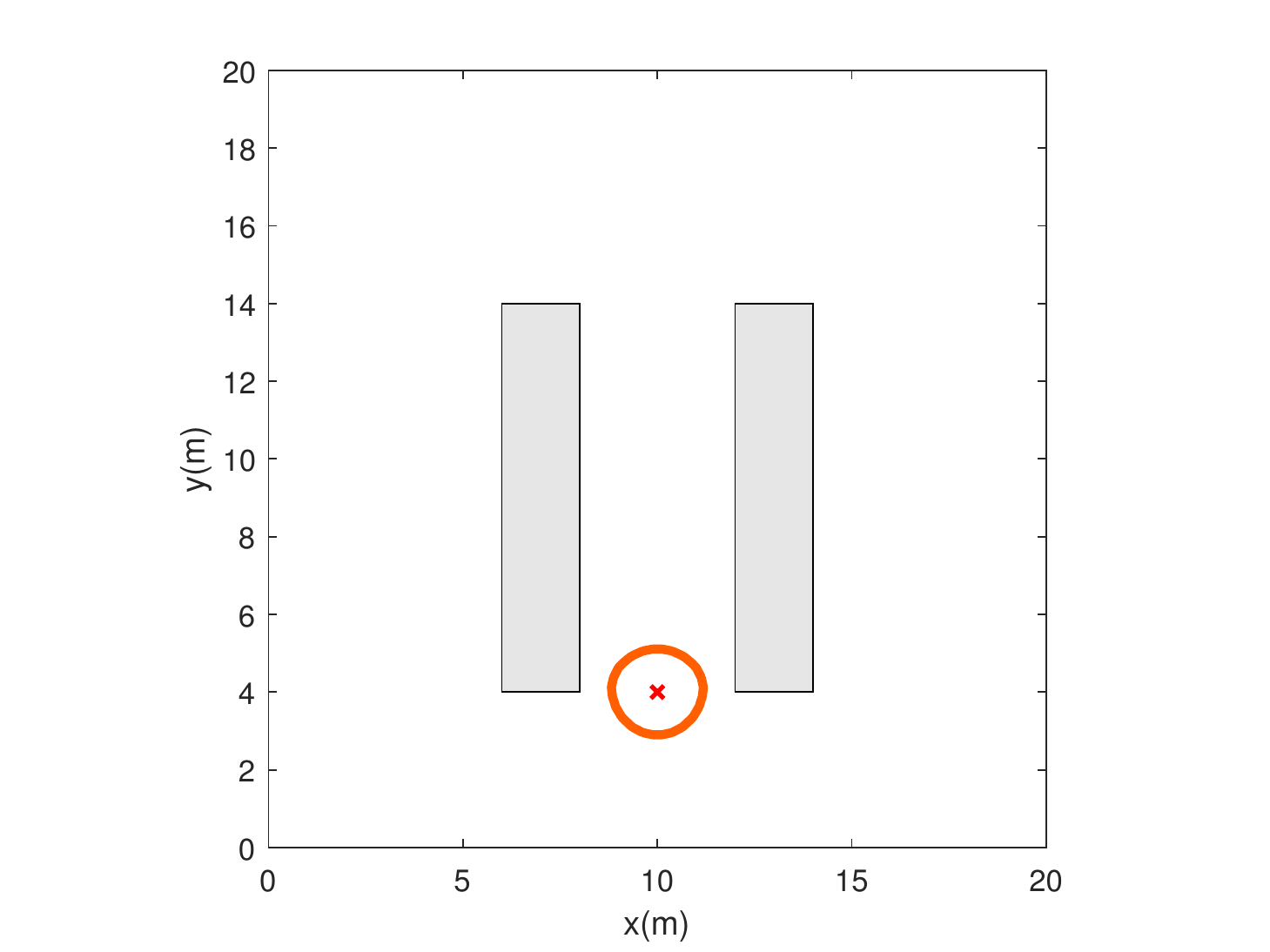}\\
			\centering{\scriptsize{(e)}}
		\end{minipage}
        \begin{minipage}[t]{0.32\linewidth}
			\includegraphics[width = 6.8cm, height = 5.8cm]{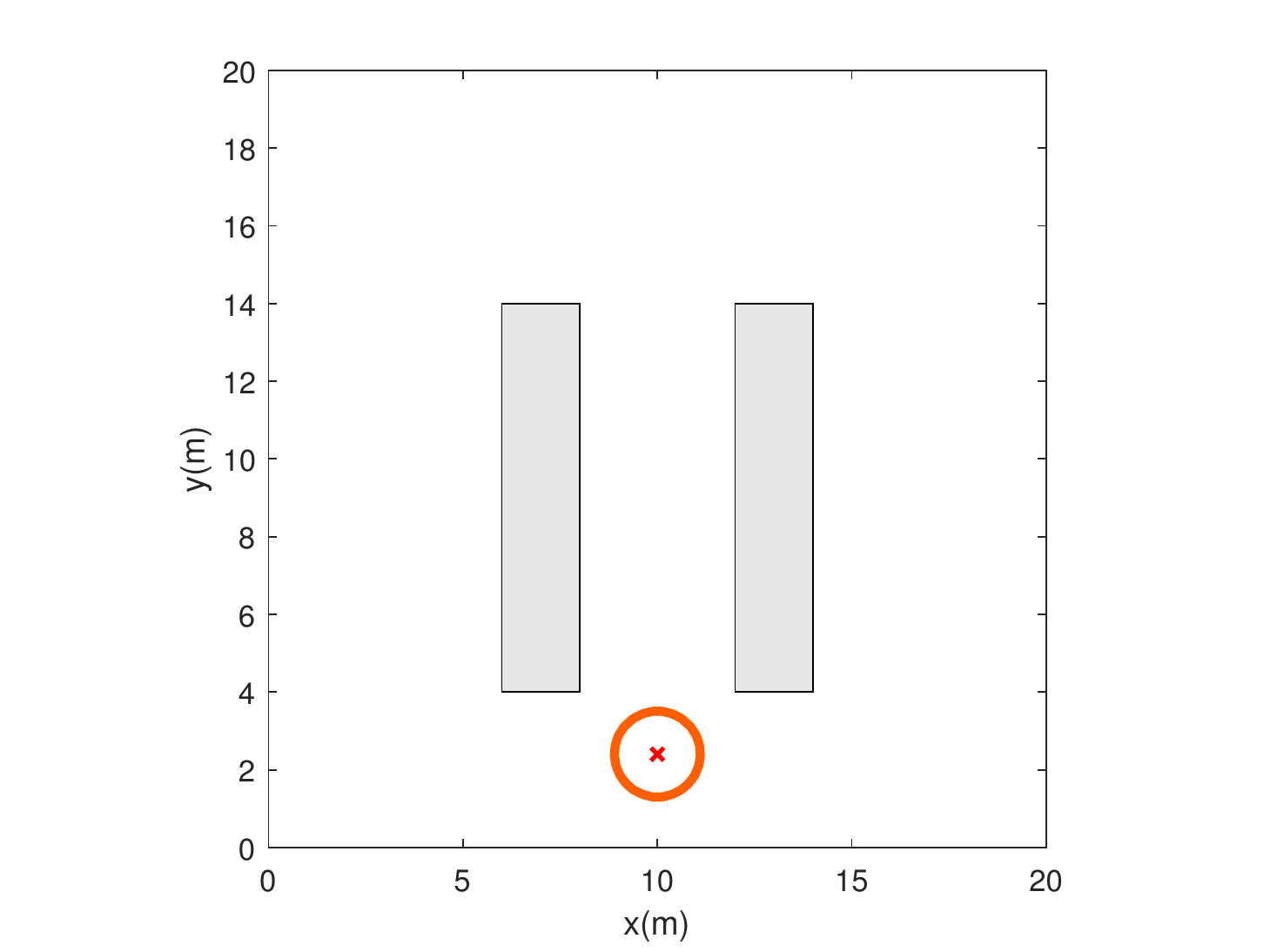}\\
			\centering{\scriptsize{(f)}}
		\end{minipage}
	\end{tabular}
	\caption{\label{fig:case1_EHGRN_one} When a target pass through a channel, the swarm pattern is formed by an EH-GRN structure, which encircles a target without colliding the channel. (a)-(f) show that the swarm pattern encircle the target in an elliptical shape way, when the target passes through the channel.}
\end{figure*}

Fig. \ref{fig:case1_EHGRN_one} shows that the swarm pattern that optimized to achieve by EH-GRN encircles a target across a channel without colliding the channel. In particular, Fig. \ref{fig:case1_EHGRN_one} (a) and (f) show if the target does not enter the channel, then the swarm pattern is a circular shape way to encircle the target. In addition, the biggest difference between Fig. \ref{fig:case1_GP_one} and Fig. \ref{fig:case1_EHGRN_one} is that the swarm pattern passes through the channel in different patterns. That is, Fig. \ref{fig:case1_EHGRN_one} (b)-(d) show that the swarm pattern encircle a target in an elliptical shape way to cross the channel and does not collide the channel. This is because the restricted scenario is relatively simple and has little impact on swarm pattern formation.

\begin{figure*}[ht]
	\begin{tabular}{cc}
		\begin{minipage}[t]{0.32\linewidth}  %  width = 4.5cm,height = 3.6cm
			\includegraphics[width = 6.8cm, height = 5.8cm]{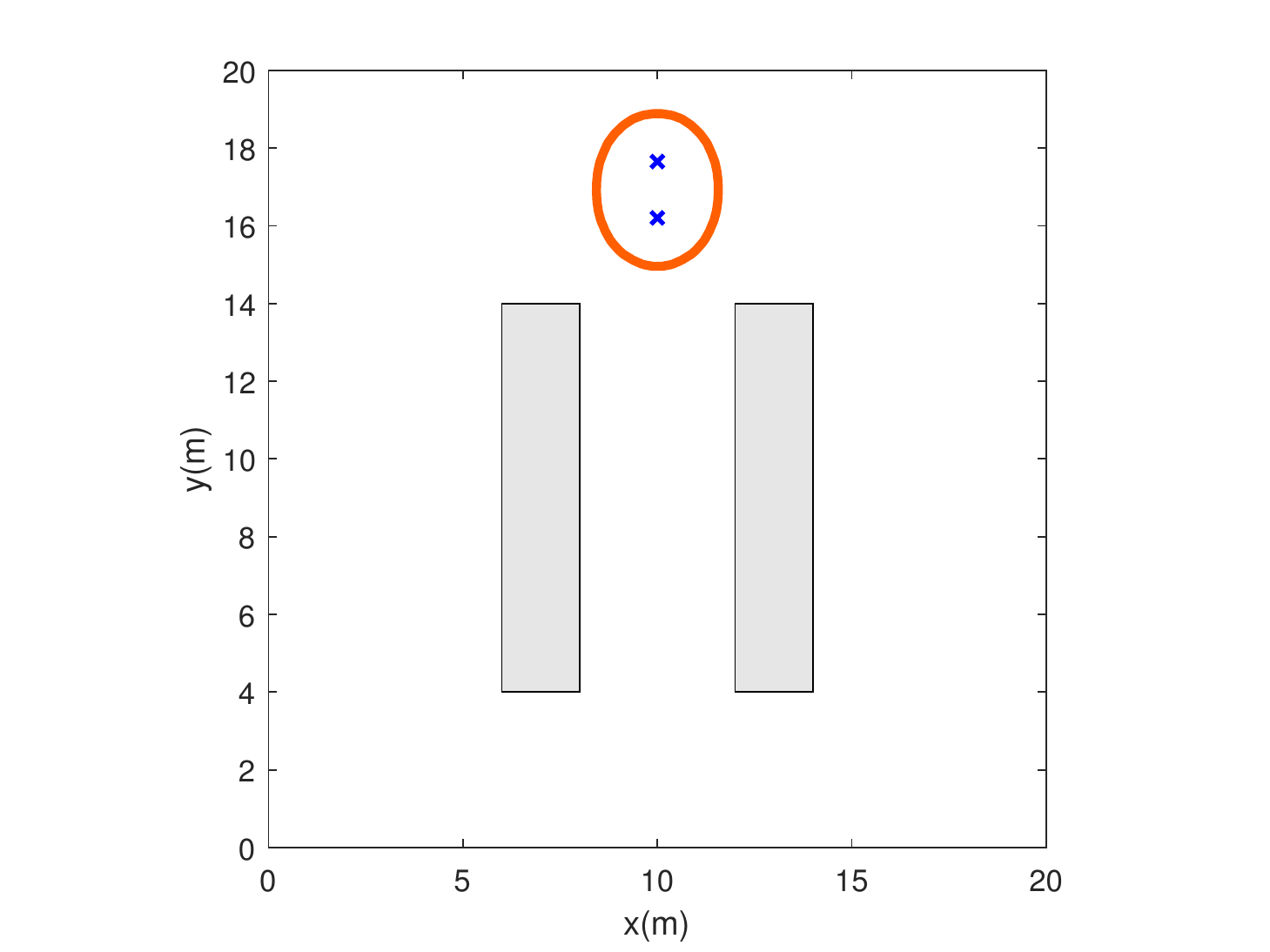}\\
			\centering{\scriptsize{(a)}}
		\end{minipage}
        \begin{minipage}[t]{0.32\linewidth}
			\includegraphics[width = 6.8cm, height = 5.8cm]{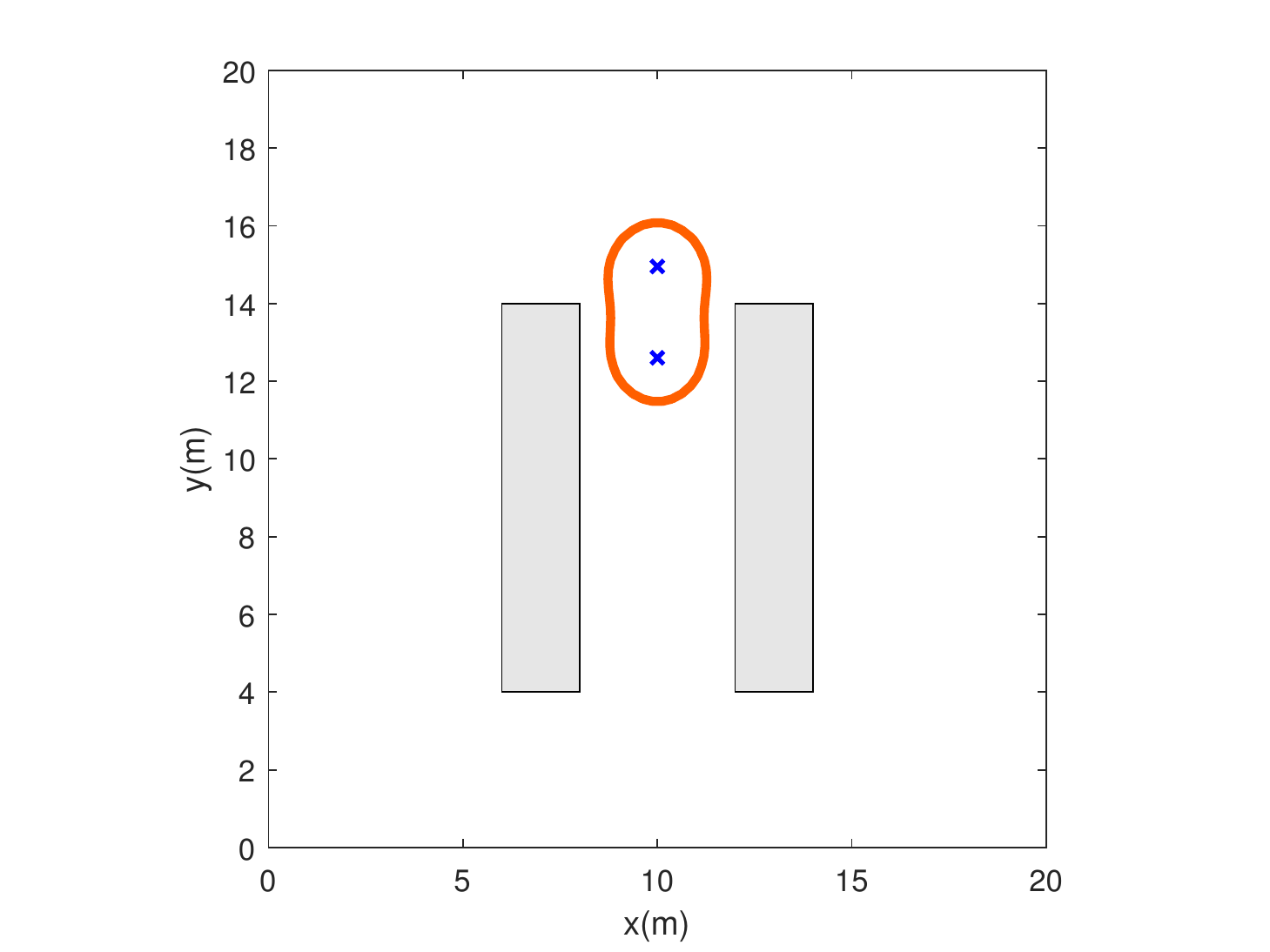}\\
			\centering{\scriptsize{(b)}}
		\end{minipage}
        \begin{minipage}[t]{0.32\linewidth}
			\includegraphics[width = 6.8cm, height = 5.8cm]{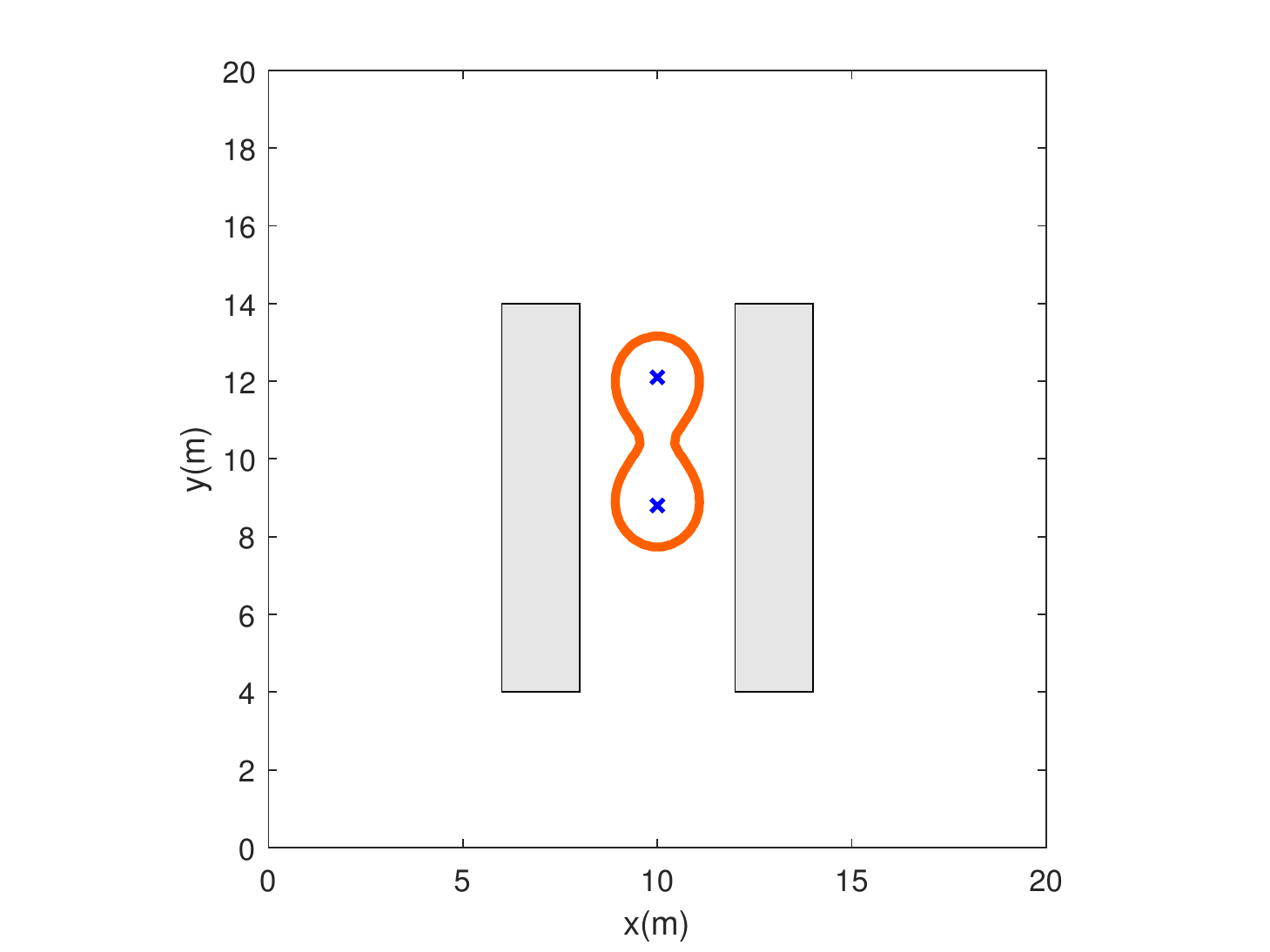}\\
			\centering{\scriptsize{(c)}}
		\end{minipage}
	\end{tabular}
	
    \begin{tabular}{cc}	
        \begin{minipage}[t]{0.32\linewidth}
			\includegraphics[width = 6.8cm, height = 5.8cm]{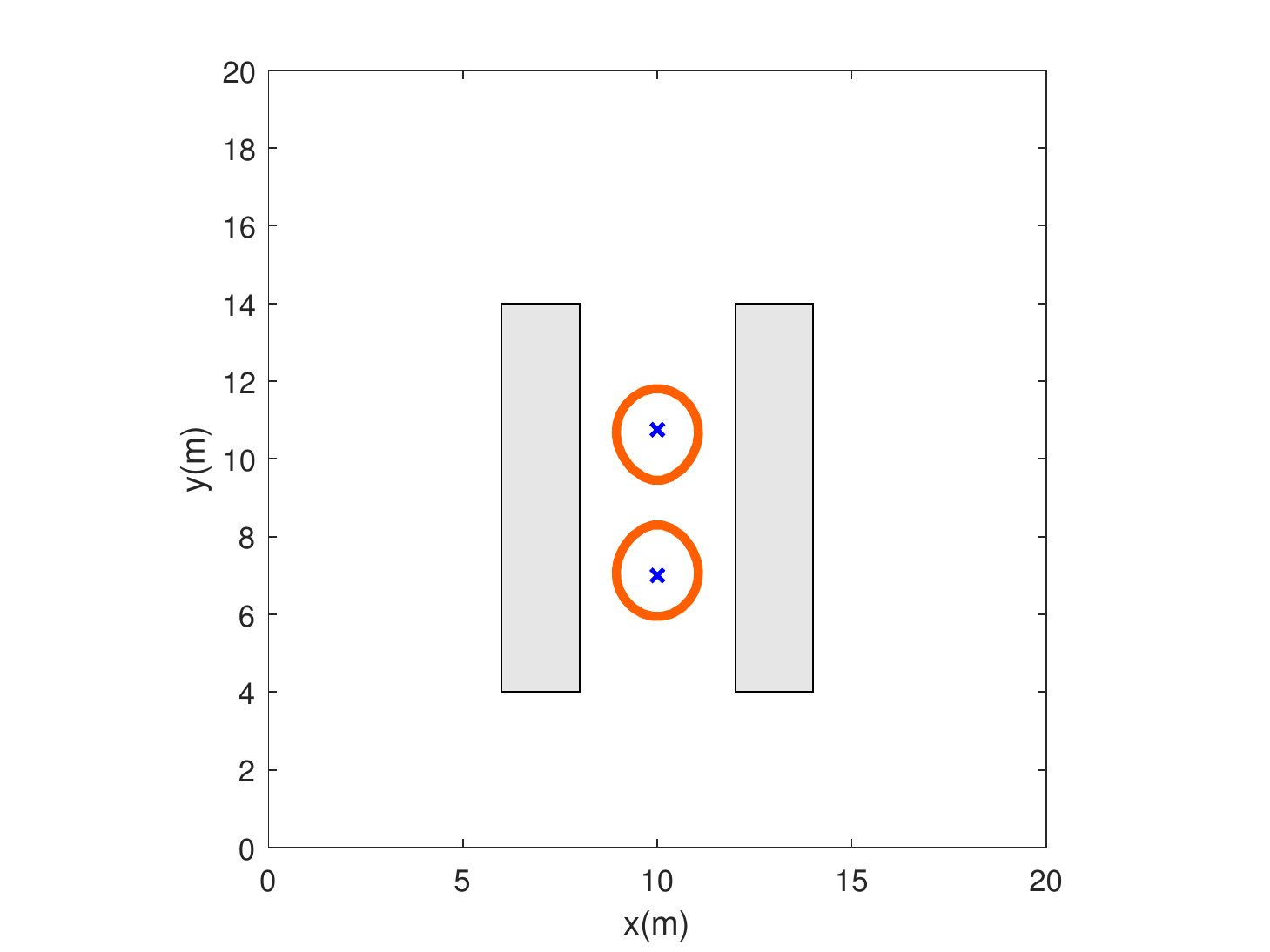}\\
			\centering{\scriptsize{(d)}}
		\end{minipage}
		\begin{minipage}[t]{0.32\linewidth}
			\includegraphics[width = 6.8cm, height = 5.8cm]{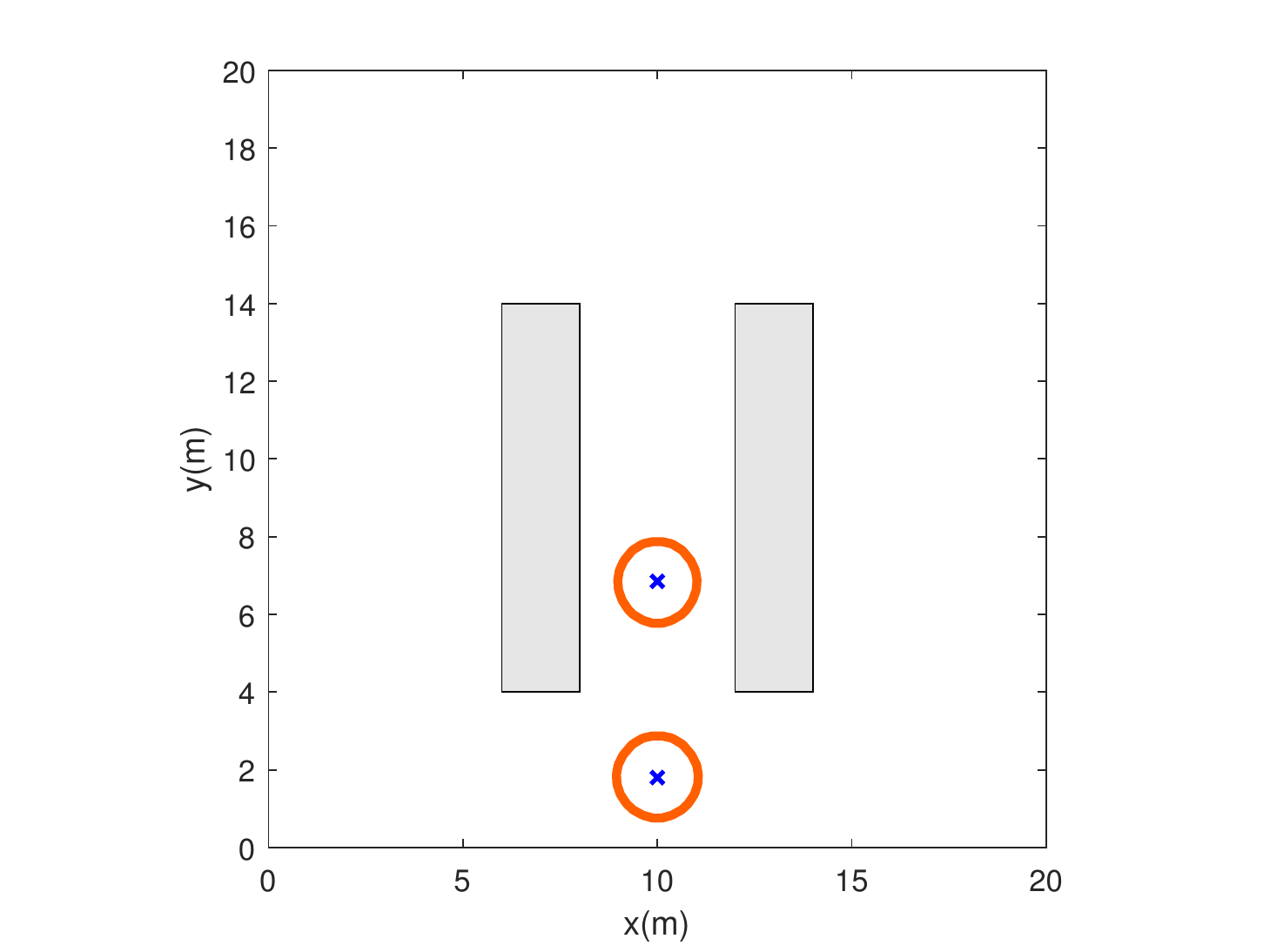}\\
			\centering{\scriptsize{(e)}}
		\end{minipage}
		\begin{minipage}[t]{0.32\linewidth}
			\includegraphics[width = 6.8cm, height = 5.8cm]{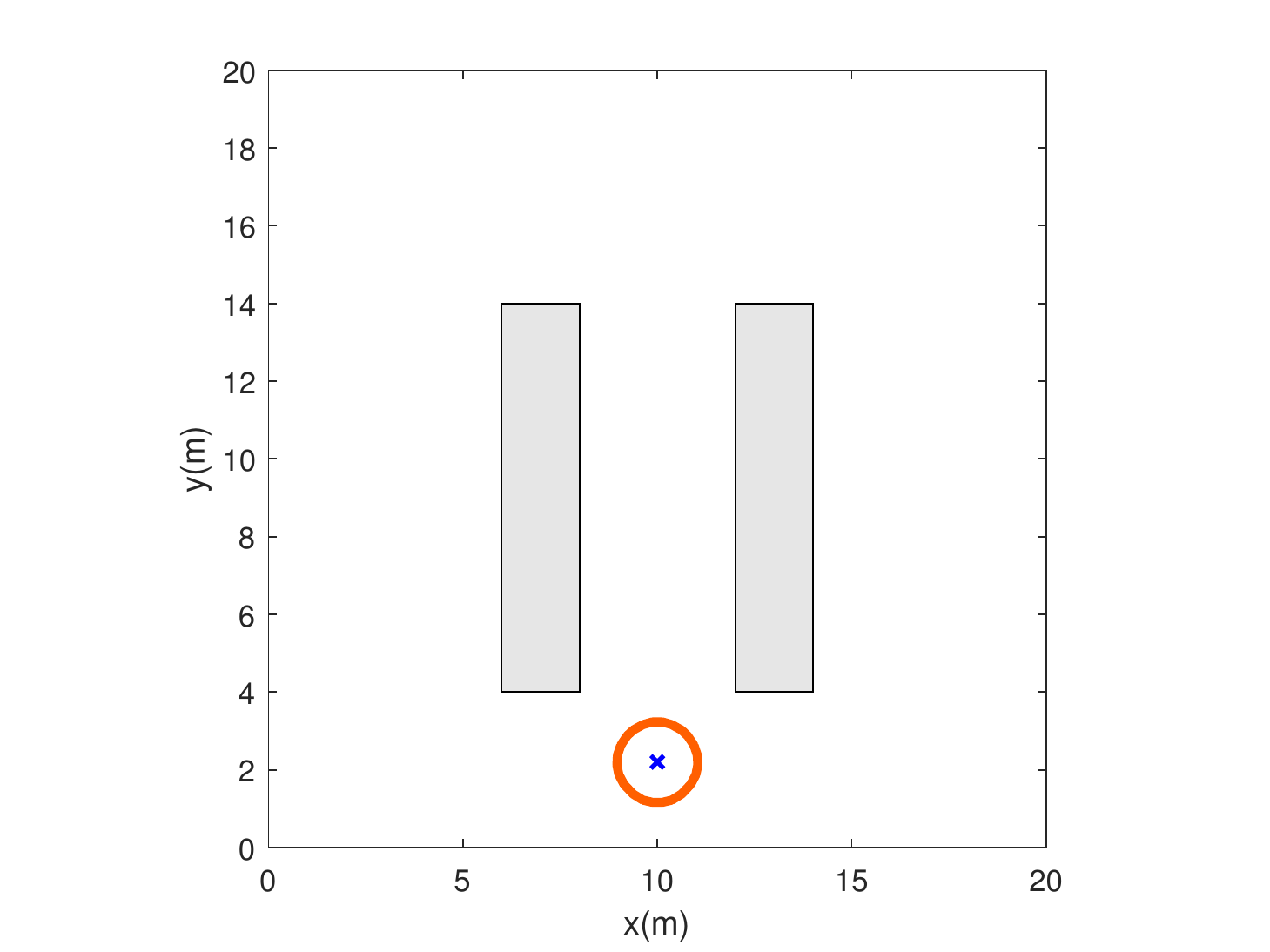}\\
			\centering{\scriptsize{(f)}}
		\end{minipage}
	\end{tabular}
	\caption{\label{fig:case1_GP_two} When two targets pass through a channel, the swarm pattern that formed by the proposed automatic design framework encircles two targets without colliding the channel. (a)-(e) show that the swarm pattern encircle the two targets in an elliptical shape way, when the targets passes through the channel.}
\end{figure*}

\begin{figure*}[ht]
	\begin{tabular}{cc}
		\begin{minipage}[t]{0.32\linewidth}  %  width = 4.5cm,height = 3.6cm
			\includegraphics[width = 6.8cm, height = 5.8cm]{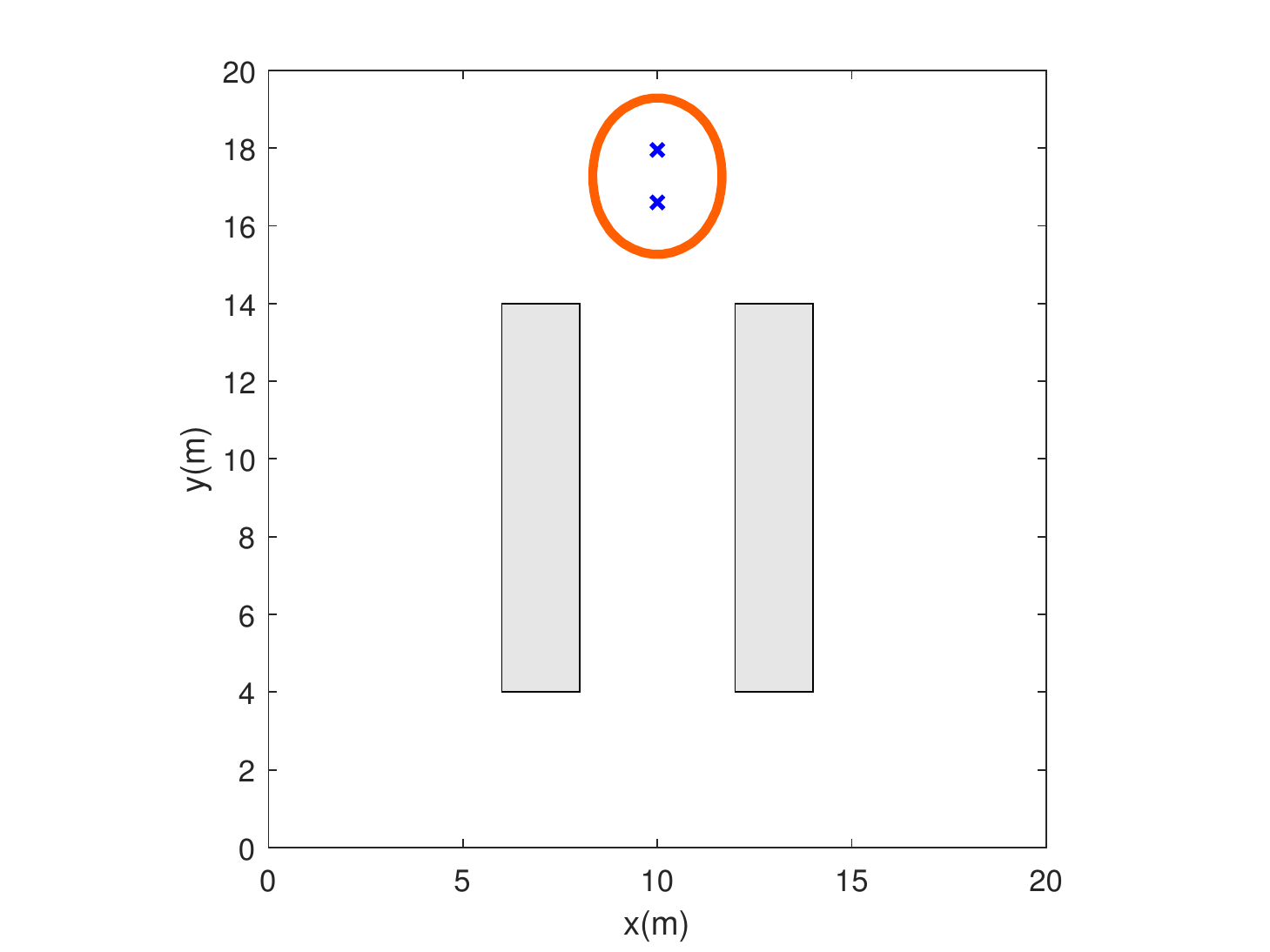}\\
			\centering{\scriptsize{(a)}}
		\end{minipage}
        \begin{minipage}[t]{0.32\linewidth}
			\includegraphics[width = 6.8cm, height = 5.8cm]{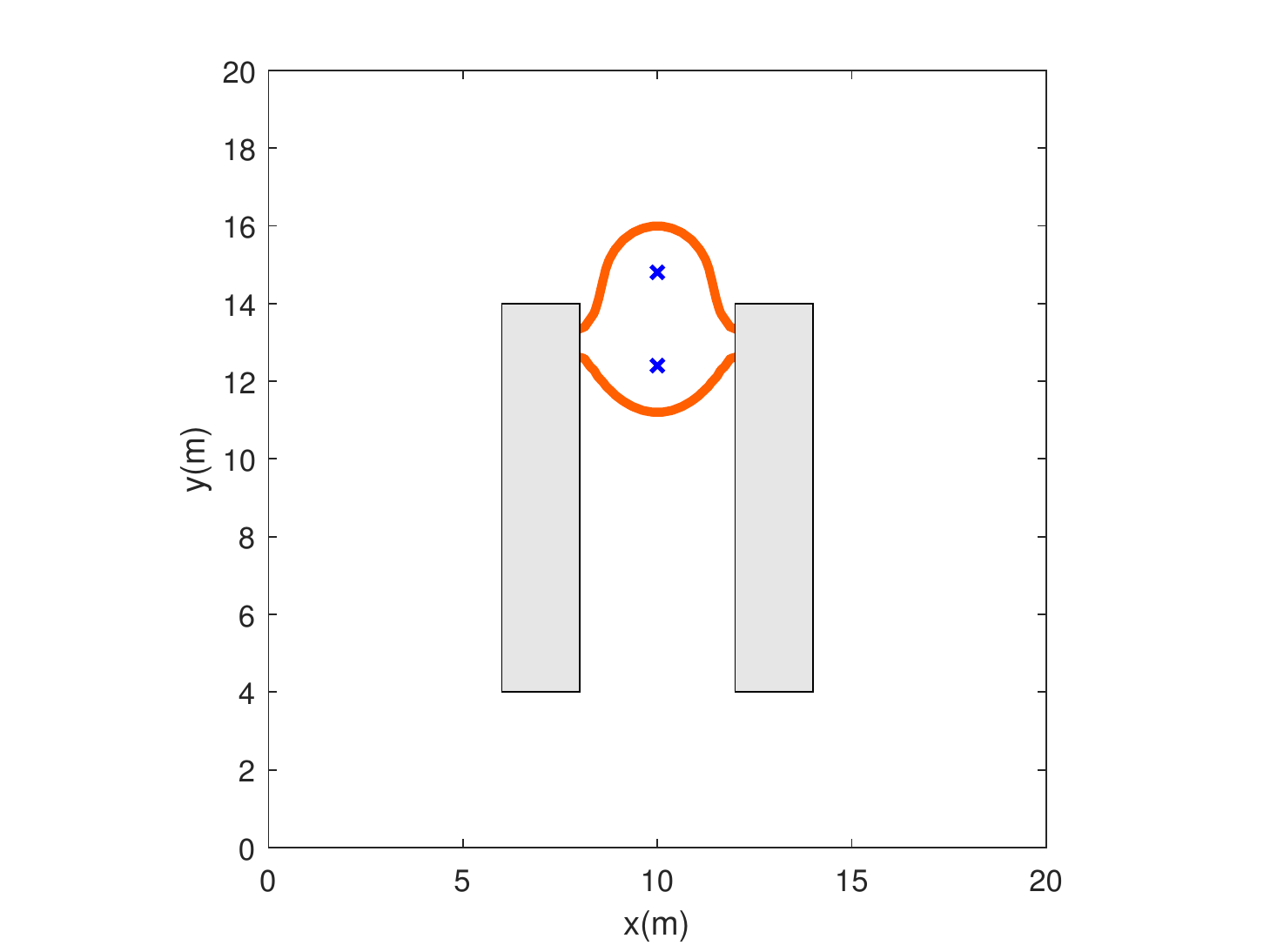}\\
			\centering{\scriptsize{(b)}}
		\end{minipage}
        \begin{minipage}[t]{0.32\linewidth}
			\includegraphics[width = 6.8cm, height = 5.8cm]{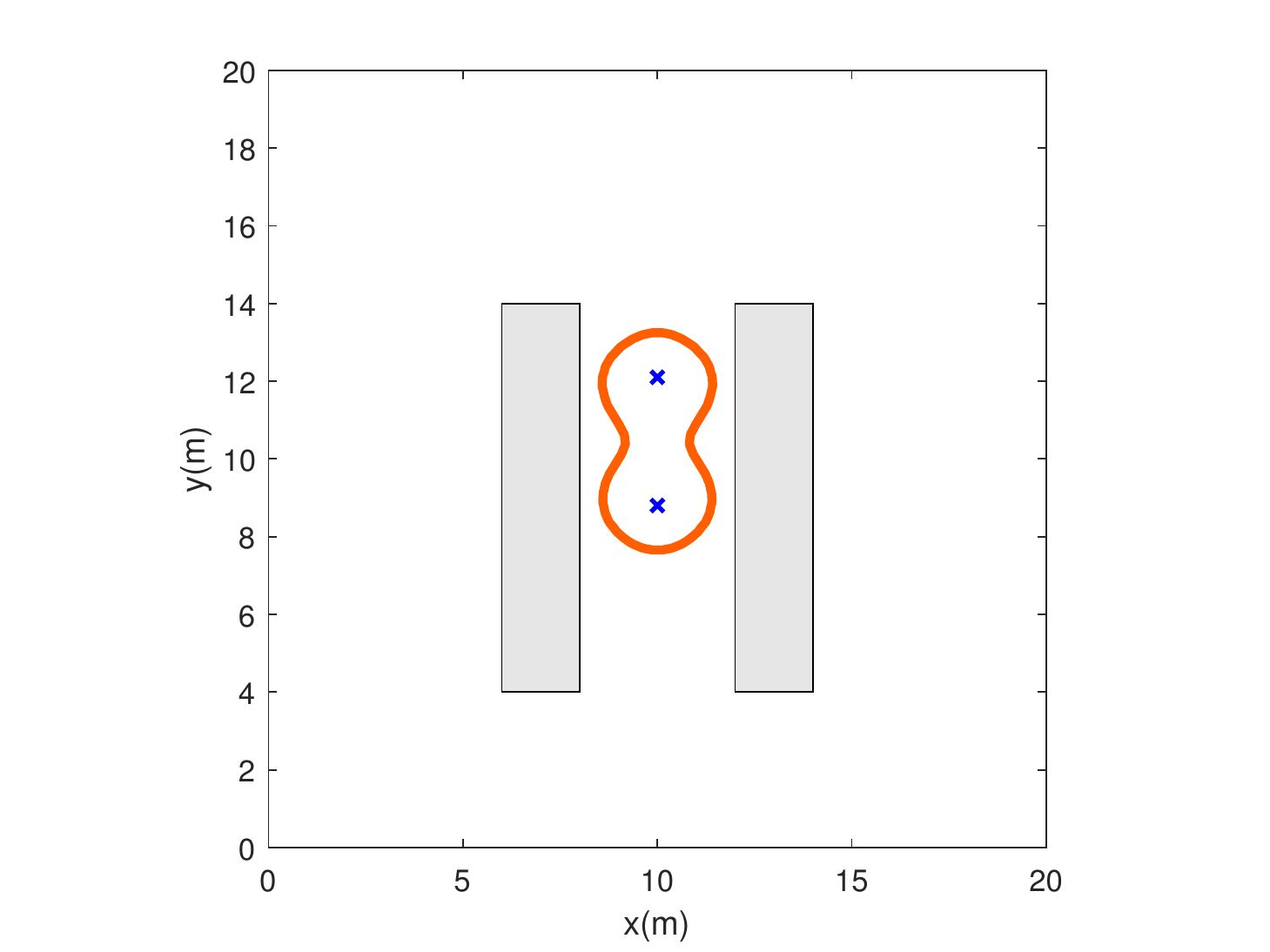}\\
			\centering{\scriptsize{(c)}}
		\end{minipage}
	\end{tabular}
	
    \begin{tabular}{cc}	
        \begin{minipage}[t]{0.32\linewidth}
			\includegraphics[width = 6.8cm, height = 5.8cm]{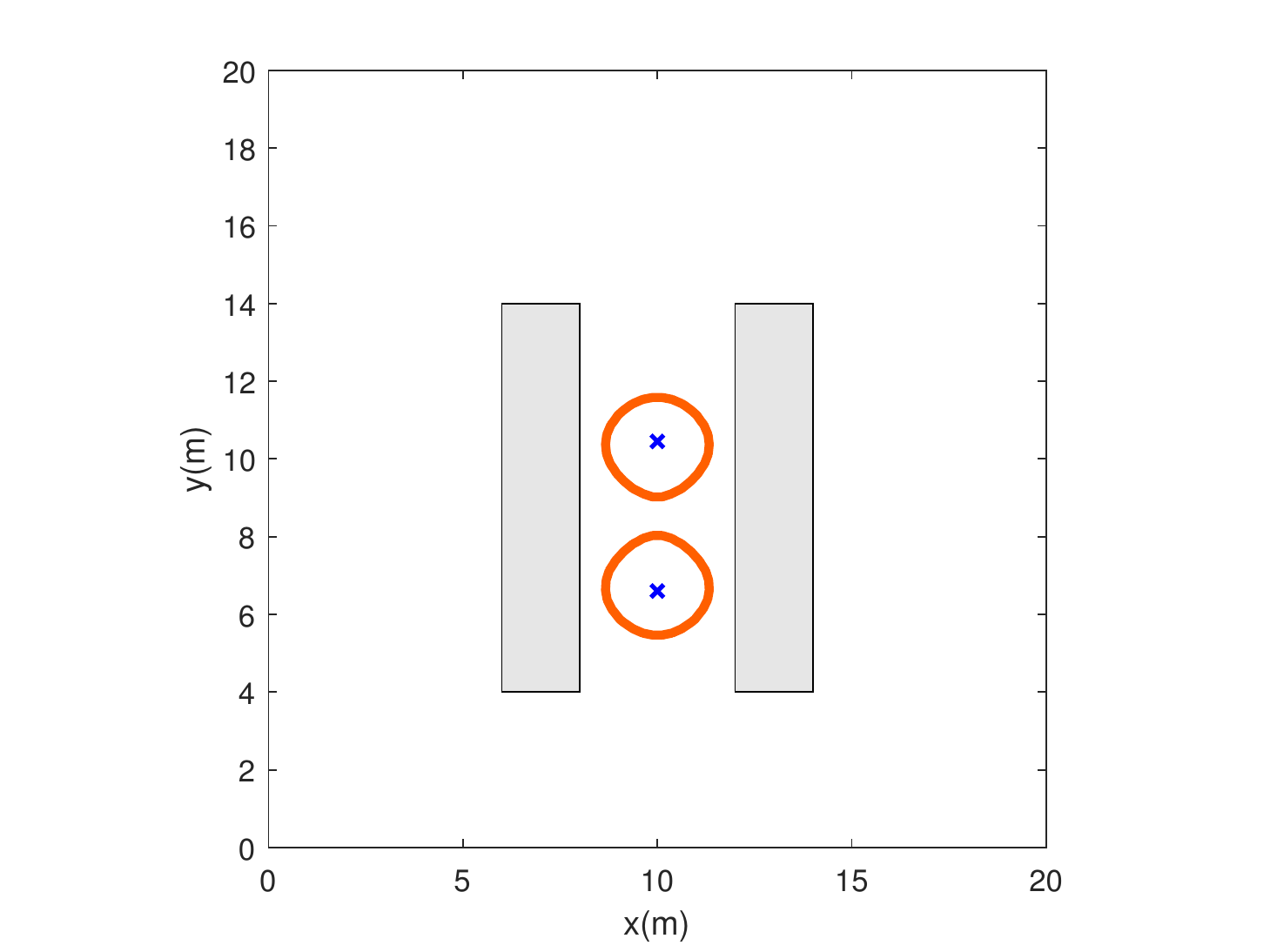}\\
			\centering{\scriptsize{(d)}}
		\end{minipage}
		\begin{minipage}[t]{0.32\linewidth}
			\includegraphics[width = 6.8cm, height = 5.8cm]{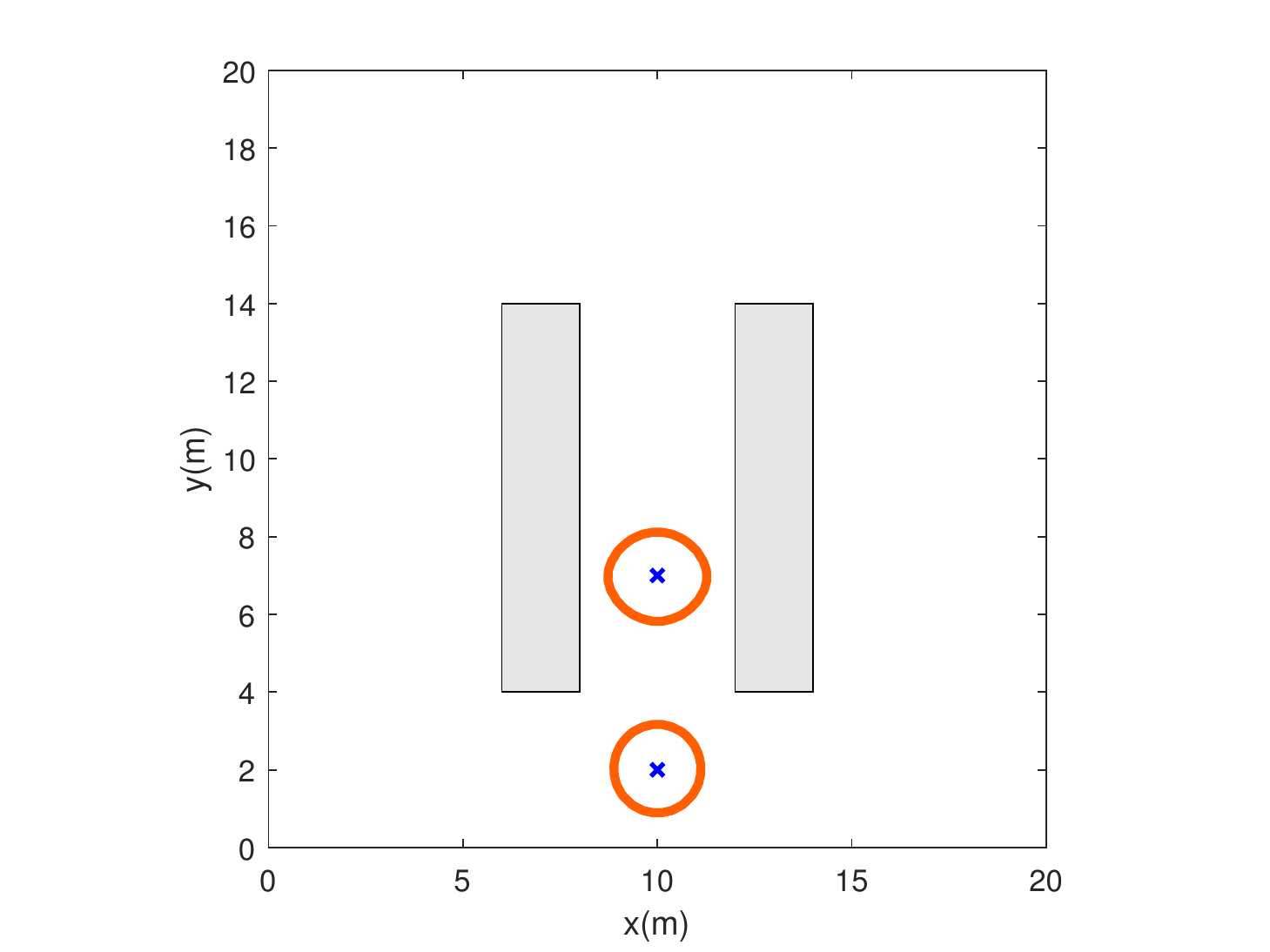}\\
			\centering{\scriptsize{(e)}}
		\end{minipage}
		\begin{minipage}[t]{0.32\linewidth}
			\includegraphics[width = 6.8cm, height = 5.8cm]{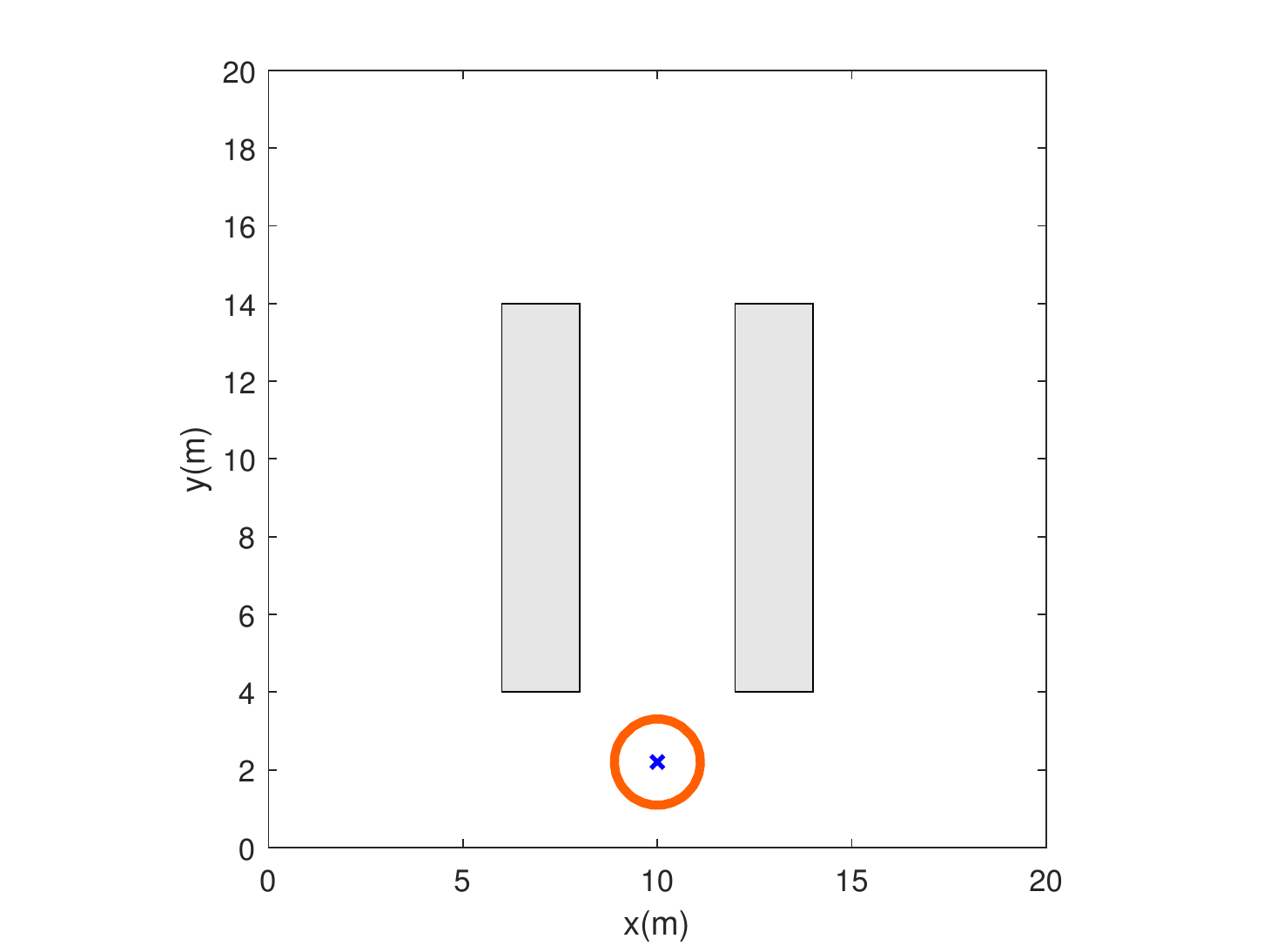}\\
			\centering{\scriptsize{(f)}}
		\end{minipage}
	\end{tabular}
	\caption{\label{fig:case1_EHGRN_two} When two targets pass through a channel, the swarm pattern that formed by an EH-GRN structure encircles two targets. (a)-(f) show that the swarm pattern encircle two targets across a channel.}
\end{figure*}

Note that, the GRN-based model of the two swarm pattern are optimized to achieve by the proposed automatic design framework and EH-GRN when one target is encircled, which directly applies to encircle two targets in the channel. Fig. \ref{fig:case1_GP_two} and Fig. \ref{fig:case1_EHGRN_two} show two different swarm patterns encircle two targets across the channel. For the proposed automatic design framework, Fig. \ref{fig:case1_GP_two} (a) shows that the swarm pattern encircles two targets in an elliptical shape and does not collide the channel when two targets are not far away. As the two targets gradually move away, the swarm pattern encircles the two targets in a gourd shape, as shown Fig. \ref{fig:case1_GP_two} (b) and (c). Finally, swarm pattern forms two separate circular shape encircling two targets respectively, as shown Fig. \ref{fig:case1_GP_two} (d) and (f). For EH-GRN, the most significant difference is when the two targets enter the channel, the swarm pattern collides with the obstacles to form two separate parts, as shown Fig. \ref{fig:case1_EHGRN_two} (b). In Fig. \ref{fig:case1_EHGRN_two} (e), swarm patterns encircle two targets in two different shapes. That is, if the target is in the channel, swarm pattern encircles a target in an elliptical shape. If the target is outside the channel, swarm pattern encircles a target as a circle shape. One reason is that the structure of the swarm pattern has been predefined, which can not find an optimize solution in the restricted scenario, as so to the width of the channel is smaller than the maximum radius of compound swarm pattern formed by surrounding two targets at the same time. For the proposed automatic design framework, it applies some efficient basic network motifs, which makes swarm pattern adapt the considered scenario.

In summary, the proposed automatic design framework can find a GRN-based model, which encircles a or two targets without colliding the channel and verifies the feasibility of the proposed framework.

\subsection{Pattern Formation in a compound channel}

\begin{figure}[!t]
\centering
\includegraphics[width=2.5in]{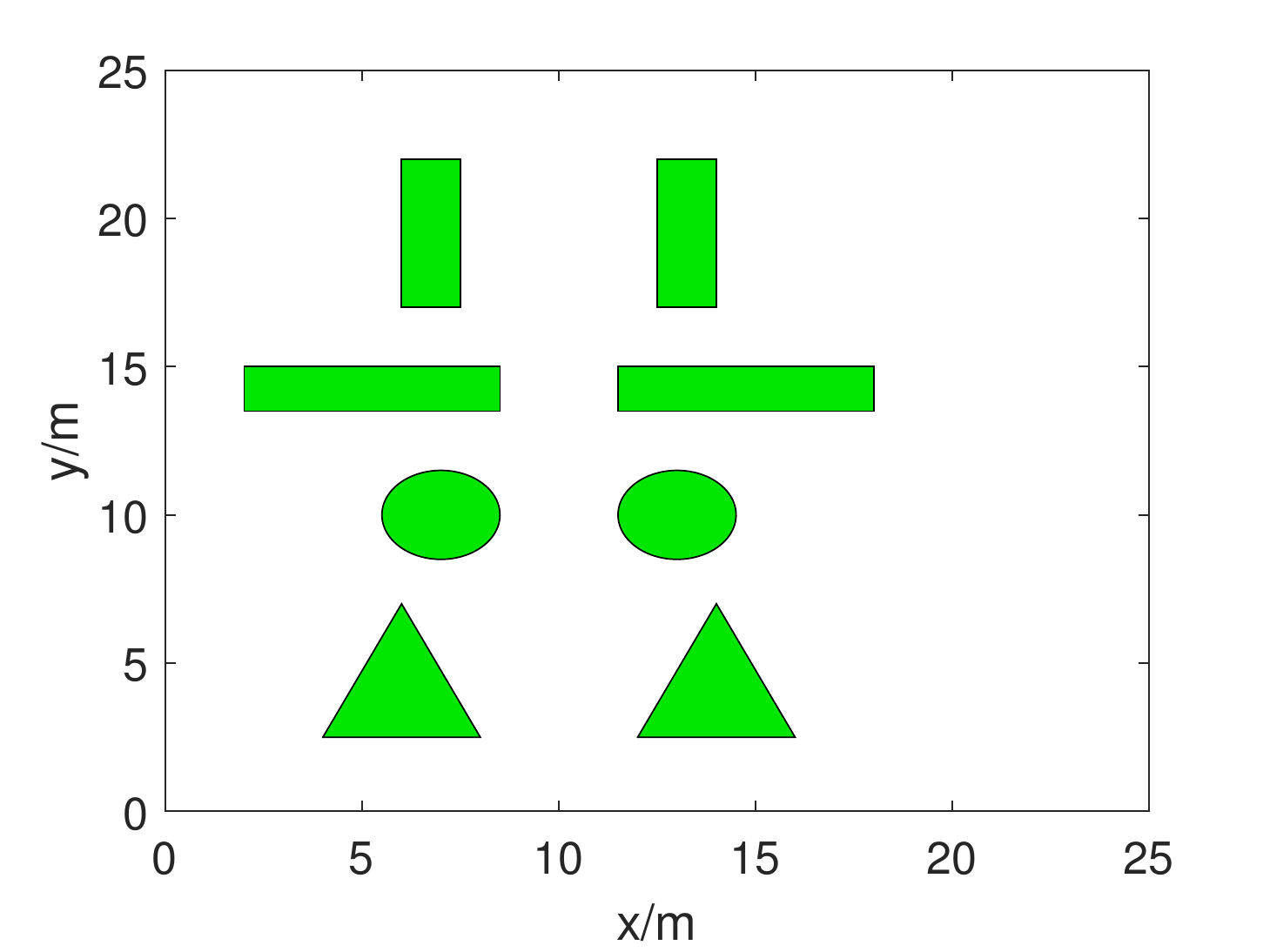}%
\hfil
\caption{A compound channel is in a region of 25 by 25 meters, and green shadows represent the compound channel.}
\label{fig:task_two}
\end{figure}

To verify the adaptability of the proposed automatic design framework, the framework is tested in a compound channel. The four parts, a channel, a narrow channel, a circular narrow channel and a T-shape channel, constitute the compound channel, as shown in Fig. \ref{fig:task_two}. In addition, the restricted scenario is in a region of 25 by 25 meters. When a target pass through the restricted scenario, the swarm pattern needs to generate various shapes to adapt to the restricted scenario. An evaluation criterion is that swarm pattern can not collide with the compound channel while encircling a target.

\begin{figure}[!t]
\centering
\includegraphics[width=3.5in]{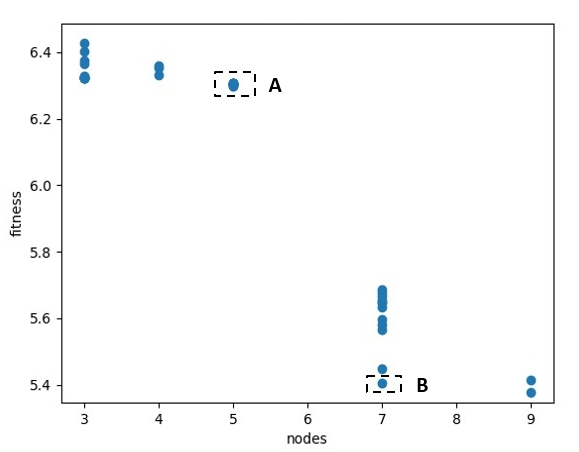}%
\hfil
\caption{The non-dominated solutions are achieved by MOGP-NSGA-II. The point A and B are called knee points. }
\label{fig:task2_PF_solution}
\end{figure}

\begin{figure}[!t]
\centering
\includegraphics[width=2in]{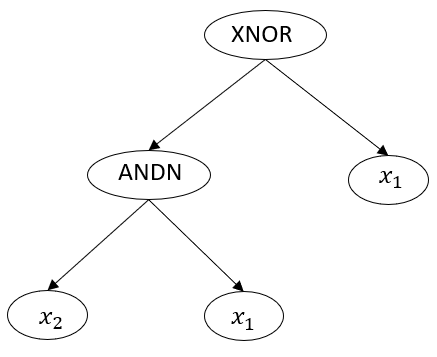}%
\hfil
\caption{The syntax tree of point A is optimized to achieve by MOGP-NSGA-II. $x_{1}$ is that the location information of target forms a morphogen gradient space. $x_{2}$ is that the location information of obstacle forms a morphogen gradient space. $ANDN$ and $XNOR$ are two basic network motifs.}
\label{fig:task2_tree_1}
\end{figure}

\begin{figure}[!t]
\centering
\includegraphics[width=2in]{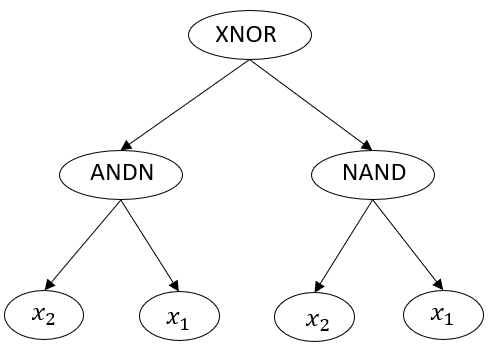}%
\hfil
\caption{The syntax tree of point B is optimized to achieve by MOGP-NSGA-II. $x_{1}$ is that the location information of target forms a morphogen gradient space. $x_{2}$ is that the location information of obstacle forms a morphogen gradient space. $ANDN$, $NAND$ and $XNOR$ are three basic network motifs.}
\label{fig:task2_tree_2}
\end{figure}

\begin{figure}[!t]
\centering
\includegraphics[width=2in]{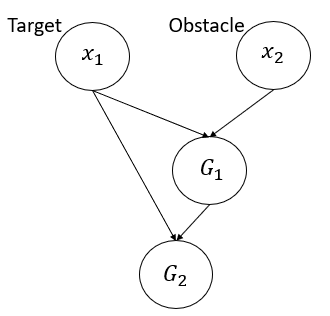}%
\hfil
\caption{The evolutionary design framework gets GRN. $x_{1}$ is that the location information of target forms a morphogen gradient space. $x_{2}$ is that the location information of obstacle forms a morphogen gradient space. $G_{1}$ and $G_{2}$ are activate genes, and the concentration of $G_{2}$ represent a morphogen gradient space to form the desired swarm pattern.}
\label{fig:task2_GRN_1}
\end{figure}

\begin{figure}[!t]
\centering
\includegraphics[width=2in]{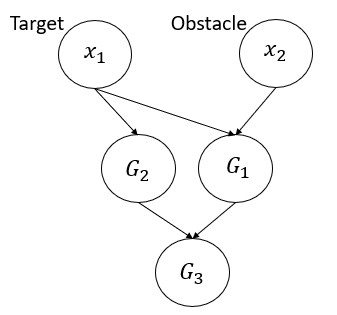}%
\hfil
\caption{The evolutionary design framework gets GRN. $x_{1}$ is that the location information of target forms a morphogen gradient space. $x_{2}$ is that the location information of obstacle forms a morphogen gradient space. $G_{1}$, $G_{2}$ and $G_{3}$ are activate genes, and the concentration of $G_{3}$ represent a morphogen gradient space to form the desired swarm pattern.}
\label{fig:task2_GRN_2}
\end{figure}

In evolution process, except for evaluation number, all the other parameters are the same as the previous simulation experiment. The evaluation number of the proposed automatic framework and EH-GRN are both set to 8000. In Fig. \ref{fig:task2_PF_solution}, the non-dominated solutions are achieved by MOGP-NSGA-II when the evaluation number reaches 8000. Fig. \ref{fig:task2_PF_solution} shows that point A and B are selected as knee points. That is, the number of nodes at point A and point B are 5 and 7, respectively. In other words, they all have a simple the structure of GRN-based model. Fig. \ref{fig:task2_tree_1} and Fig. \ref{fig:task2_tree_2} show the syntax tree of point A and point B, respectively. For Fig. \ref{fig:task2_tree_1} and Fig. \ref{fig:task2_tree_2}, there is a difference: the right structure of the syntax tree at point B is more complex than that at point A. Because of this, the fitness value of point B is lower than that of point A. That is, the GRN-based model of the point B is more suitable for the restricted scenario than that of point A. Fig. \ref{fig:task2_GRN_1} and Fig. \ref{fig:task2_GRN_2} show the GRN structure of point A and point B, respectively.

In addition, the mathematical description of the point A is as follows:
\begin{equation}
\label{equ:tree_3}
\begin{split}
\frac{dy_{1}}{dt} = &- y_{1} + sig(x_{2} \ast (1 - x_{1}) ,\theta_{1},k)) \\
\frac{dy_{2}}{dt} = & - y_{2} + 1 - sig(y_{1} \ast (1 - x_{1}) ,\theta_{2},k))\\
& - sig((1 - y_{1}) \ast x_{1} ,\theta_{2},k))
\end{split}
\end{equation}
where $\theta_{1}$ and $\theta_{2}$ are 1.3072 and 0.7472, respectively. Eq.\eqref{equ:tree_3} is a mathematical description of $ANDN$ and $XNOR$. $y_{1}$ and $y_{2}$ are morphogen concentrations, where $y_{2}$ is a morphogen gradient that defines the swarm pattern.

In addition, the mathematical description of the point B is as follows:
\begin{equation}
\label{equ:tree_5}
\begin{split}
\frac{dy_{1}}{dt} = &- y_{1} + sig(x_{2} \ast (1 - x_{1}) ,\theta_{1},k)) \\
\frac{dy_{2}}{dt} = &- y_{2} + 1 - sig(x_{1} \ast x_{1},\theta_{2},k)) \\
\frac{dy_{3}}{dt} = & - y_{3} + 1 - sig(y_{1} \ast (1 - y_{2}) ,\theta_{3},k))\\
& - sig((1 - y_{1}) \ast y_{2} ,\theta_{3},k))
\end{split}
\end{equation}
%\begin{eqnarray}
%\label{equ:tree_5}
%\frac{dy_{1}}{dt} = - y_{1} + sig(x_{2} \ast (1 - x_{1}) ,\theta_{1},k))
%\end{eqnarray}
%\begin{eqnarray}
%\label{equ:tree_6}
%\frac{dy_{2}}{dt} = - y_{2} + 1 - sig(x_{1} \ast x_{1},\theta_{2},k))
%\end{eqnarray}
%\begin{equation}
%\label{equ:tree_7}
%\begin{split}
%\frac{dy_{3}}{dt} = & - y_{3} + 1 - sig(y_{1} \ast (1 - y_{2}) ,\theta_{3},k))\\
%& - sig((1 - y_{1}) \ast y_{2} ,\theta_{3},k))
%\end{split}
%\end{equation}
where $\theta_{1}$, $\theta_{2}$ and $\theta_{3}$ are 1.5441, 0.0904 and 0.2414, respectively. Eq.\eqref{equ:tree_5} is a mathematical description of $ANDN$, $NAND$, and $XNOR$. $y_{1}$, $y_{2}$ and $y_{3}$ are morphogen concentrations, where $y_{3}$ is a morphogen gradient that defines the swarm pattern.

\begin{figure}[!t]
\centering
\includegraphics[width=2.5in]{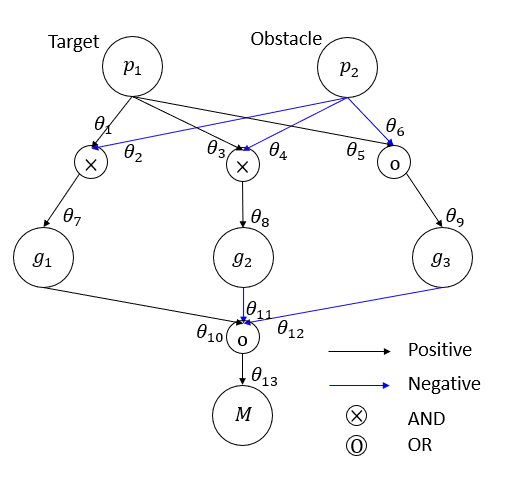}%
\hfil
\caption{Illustration of an EH-GRN structure for swarm pattern formation. The structure of model is predefined, and CMA-ES is applied to optimise the regulatory parameters. $g_{1}$, $g_{2}$ and $g_{3}$ are activate genes. The concentration of $M$ represent a morphogen gradient space to form the desired swarm pattern.}
\label{fig:task2_EH}
\end{figure}

To measure the adaptability of the proposed automatic design framework, EH-GRN is tested in the compound channel. Fig. \ref{fig:task2_EH} shows an EH-GRN structure for swarm pattern formation, which is optimized to achieve by CMA-ES in the restricted scenario. In addition, $p_{1}$ and $p_{2}$ represent a morphogen concentration produced by the environmental inputs in Fig. \ref{fig:task2_EH}. In other words, $p_{1}$ and $p_{2}$ represent the morphogen concentrations produced by the environmental inputs. $g_{1}$, $g_{2}$ and $g_{3}$ are activate genes. The concentration of $M$ represent a morphogen gradient space to form the desired swarm pattern. $\theta_{i}$ ($i=1,...,13$) is a regulatory parameter.

For an EH-GRN structure for swarm pattern formation in the considered scenario, each swarm robots follows the following dynamic equations to generate a morphogen gradient space that define the swarm pattern.
\begin{align}
& \frac{dy_{1}}{dt} = - y_{1} + sig(p_{1},\theta_{1},k))  \label{eq:HGRN2_1} \\
& \frac{dy_{2}}{dt} = - y_{2} + 1 - sig(p_{2},\theta_{2},k))  \label{eq:HGRN2_2} \\
& \frac{dg_{1}}{dt} = - g_{1} + sig(y_{1} \ast y_{2},\theta_{7},k)) \label{eq:HGRN2_3} \\
& \frac{dy_{3}}{dt} = - y_{3} + sig(p_{1},\theta_{3},k)) \label{eq:HGRN2_4} \\
& \frac{dy_{4}}{dt} = - y_{4} + 1 - sig(p_{2},\theta_{4},k)) \label{eq:HGRN2_5} \\
& \frac{dg_{2}}{dt} = - g_{2} + sig(y_{3} \ast y_{4},\theta_{8},k)) \label{eq:HGRN2_6} \\
& \frac{dy_{5}}{dt} = - y_{5} + sig(p_{1},\theta_{5},k))  \label{eq:HGRN2_7} \\
& \frac{dg_{2}}{dt} = - g_{2} + sig(y_{3} \ast y_{4},\theta_{8},k)) \label{eq:HGRN2_8} \\
& \frac{dy_{5}}{dt} = - y_{5} + sig(p_{1},\theta_{5},k)) \label{eq:HGRN2_9} \\
& \frac{dy_{7}}{dt} = - y_{7} + sig(g_{1},\theta_{10},k)) \label{eq:HGRN2_10}
\end{align}
\begin{align}
& \frac{dy_{8}}{dt} = - y_{8} + sig(g_{2},\theta_{11},k)) \label{eq:HGRN2_11} \\
& \frac{dy_{9}}{dt} = - y_{9} + 1 - sig(g_{3},\theta_{12},k)) \label{eq:HGRN2_12} \\
& \frac{dM}{dt} = - M + sig(y_{7} + y_{8} + y_{9},\theta_{13},k)) \label{eq:HGRN2_13}
\end{align}
%\begin{eqnarray}
%\label{equ:HGRN2_1}
%\frac{dy_{1}}{dt} = - y_{1} + sig(p_{1},\theta_{1},k))
%\end{eqnarray}
%\begin{eqnarray}
%\label{equ:HGRN2_2}
%\frac{dy_{2}}{dt} = - y_{2} + 1 - sig(p_{2},\theta_{2},k))
%\end{eqnarray}
%\begin{eqnarray}
%\label{equ:HGRN2_3}
%\frac{dg_{1}}{dt} = - g_{1} + sig(y_{1} \ast y_{2},\theta_{7},k))
%\end{eqnarray}
%\begin{eqnarray}
%\label{equ:HGRN2_4}
%\frac{dy_{3}}{dt} = - y_{3} + sig(p_{1},\theta_{3},k))
%\end{eqnarray}
%\begin{eqnarray}
%\label{equ:HGRN2_5}
%\frac{dy_{4}}{dt} = - y_{4} + 1 - sig(p_{2},\theta_{4},k))
%\end{eqnarray}
%\begin{eqnarray}
%\label{equ:HGRN2_6}
%\frac{dg_{2}}{dt} = - g_{2} + sig(y_{3} \ast y_{4},\theta_{8},k))
%\end{eqnarray}
%\begin{eqnarray}
%\label{equ:HGRN2_7}
%\frac{dy_{5}}{dt} = - y_{5} + sig(p_{1},\theta_{5},k))
%\end{eqnarray}
%\begin{eqnarray}
%\label{equ:HGRN2_8}
%\frac{dy_{6}}{dt} = - y_{6} + 1 - sig(p_{2},\theta_{6},k))
%\end{eqnarray}
%\begin{eqnarray}
%\label{equ:HGRN2_9}
%\frac{dg_{3}}{dt} = - g_{3} + sig(y_{5} + y_{6},\theta_{9},k))
%\end{eqnarray}
%\begin{eqnarray}
%\label{equ:HGRN2_10}
%\frac{dy_{7}}{dt} = - y_{7} + sig(g_{1},\theta_{10},k))
%\end{eqnarray}
%\begin{eqnarray}
%\label{equ:HGRN2_11}
%\frac{dy_{8}}{dt} = - y_{8} + sig(g_{2},\theta_{11},k))
%\end{eqnarray}
%\begin{eqnarray}
%\label{equ:HGRN2_12}
%\frac{dy_{9}}{dt} = - y_{9} + 1 - sig(g_{3},\theta_{12},k))
%\end{eqnarray}
%\begin{eqnarray}
%\label{equ:HGRN2_13}
%\frac{dM}{dt} = - M + sig(y_{7} + y_{8} + y_{9},\theta_{13},k))
%\end{eqnarray}
where the following parameter values: $\theta_{1} = 0.1438$, $\theta_{2} = 1$, $\theta_{3} = 0.3457$, $\theta_{4} = 0.8571$, $\theta_{5} = 0.3827$, $\theta_{6} = 1$, $\theta_{7} = 1.5841$, $\theta_{8} = 1.1972$, $\theta_{9} = 0.4208$, $\theta_{10} = 0$, $\theta_{11} = 0$, $\theta_{12} = 0.5977$, $\theta_{13} = 0.6777$.

\begin{figure*}[ht]
	\begin{tabular}{cc}
		\begin{minipage}[t]{0.32\linewidth}  %  width = 4.5cm,height = 3.6cm
			\includegraphics[width = 6.8cm, height = 5.8cm]{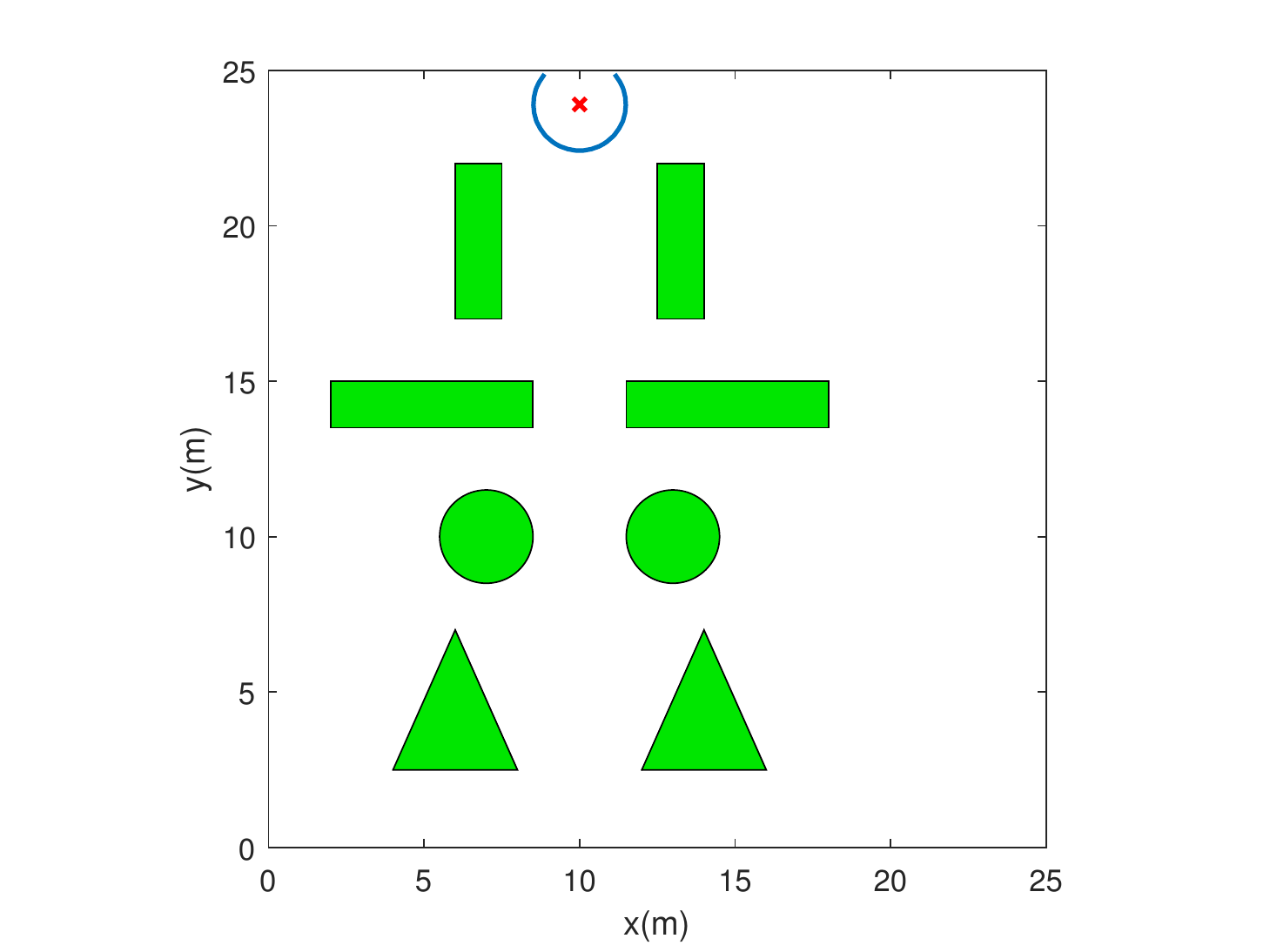}\\
			\centering{\scriptsize{(a)}}
		\end{minipage}
        \begin{minipage}[t]{0.32\linewidth}
			\includegraphics[width = 6.8cm, height = 5.8cm]{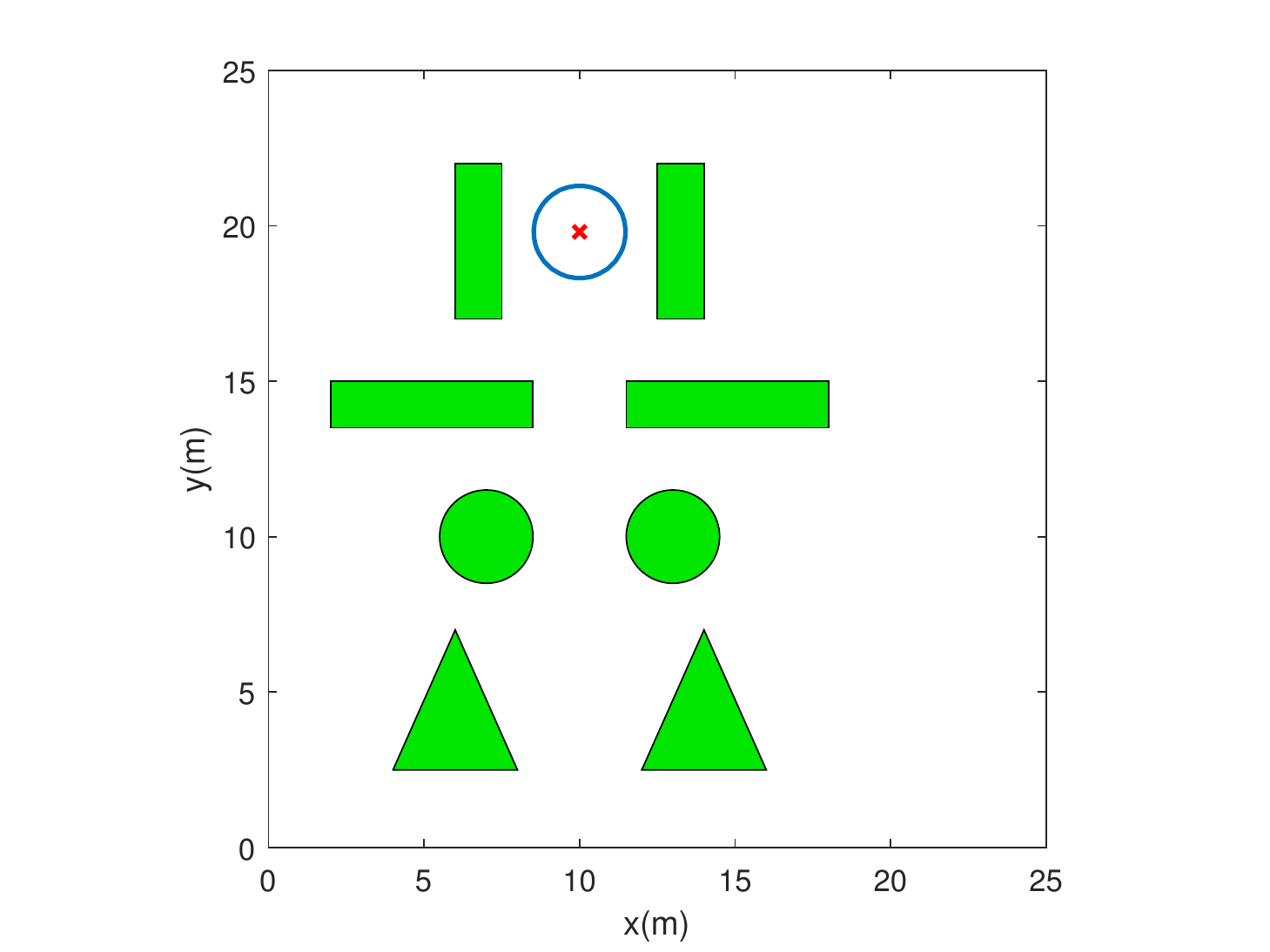}\\
			\centering{\scriptsize{(b)}}
		\end{minipage}
        \begin{minipage}[t]{0.32\linewidth}
			\includegraphics[width = 6.8cm, height = 5.8cm]{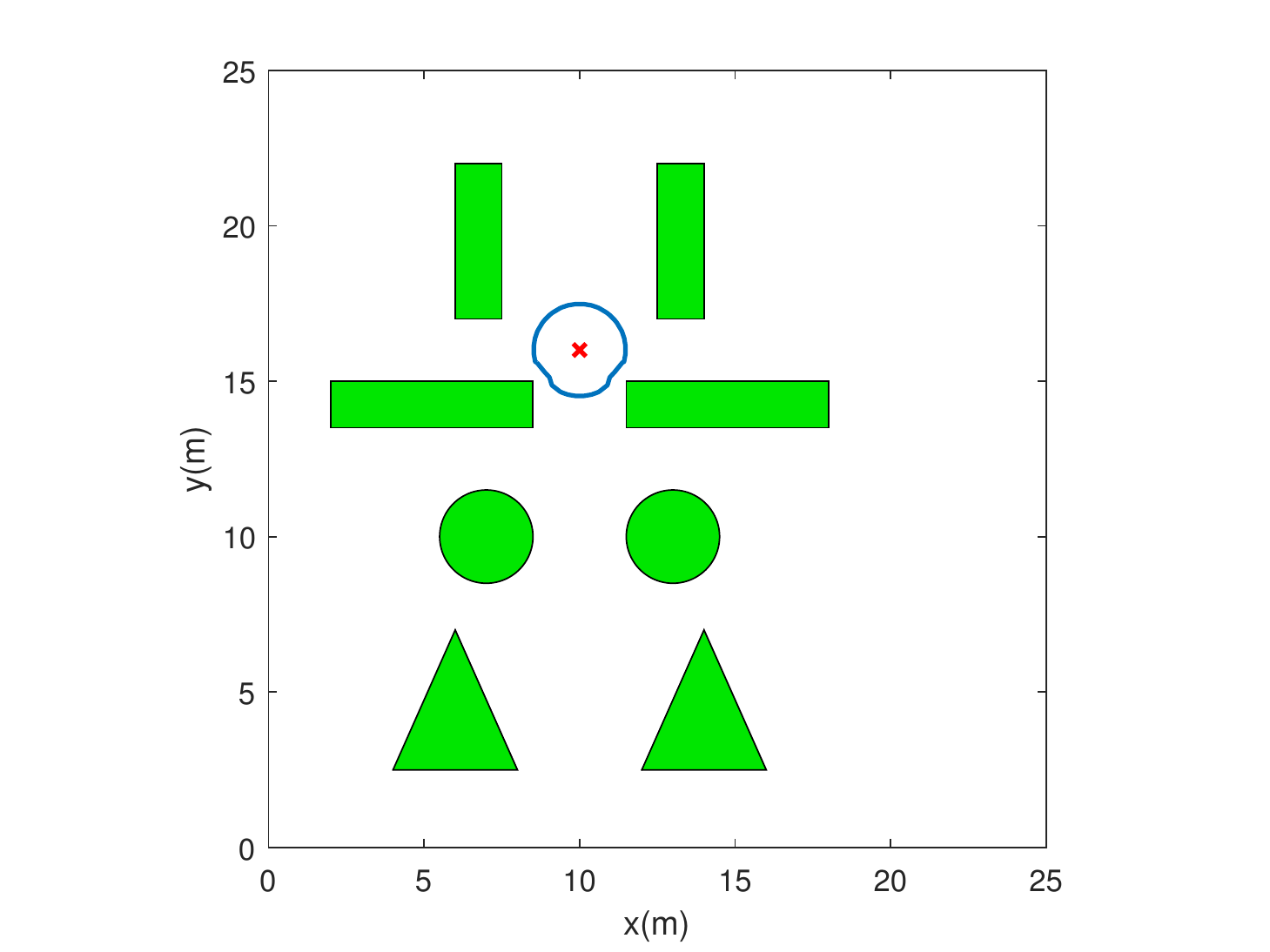}\\
			\centering{\scriptsize{(c)}}
		\end{minipage}
	\end{tabular}
	\hspace{0.5cm}
    \begin{tabular}{cc}	
        \begin{minipage}[t]{0.32\linewidth}
			\includegraphics[width = 6.8cm, height = 5.8cm]{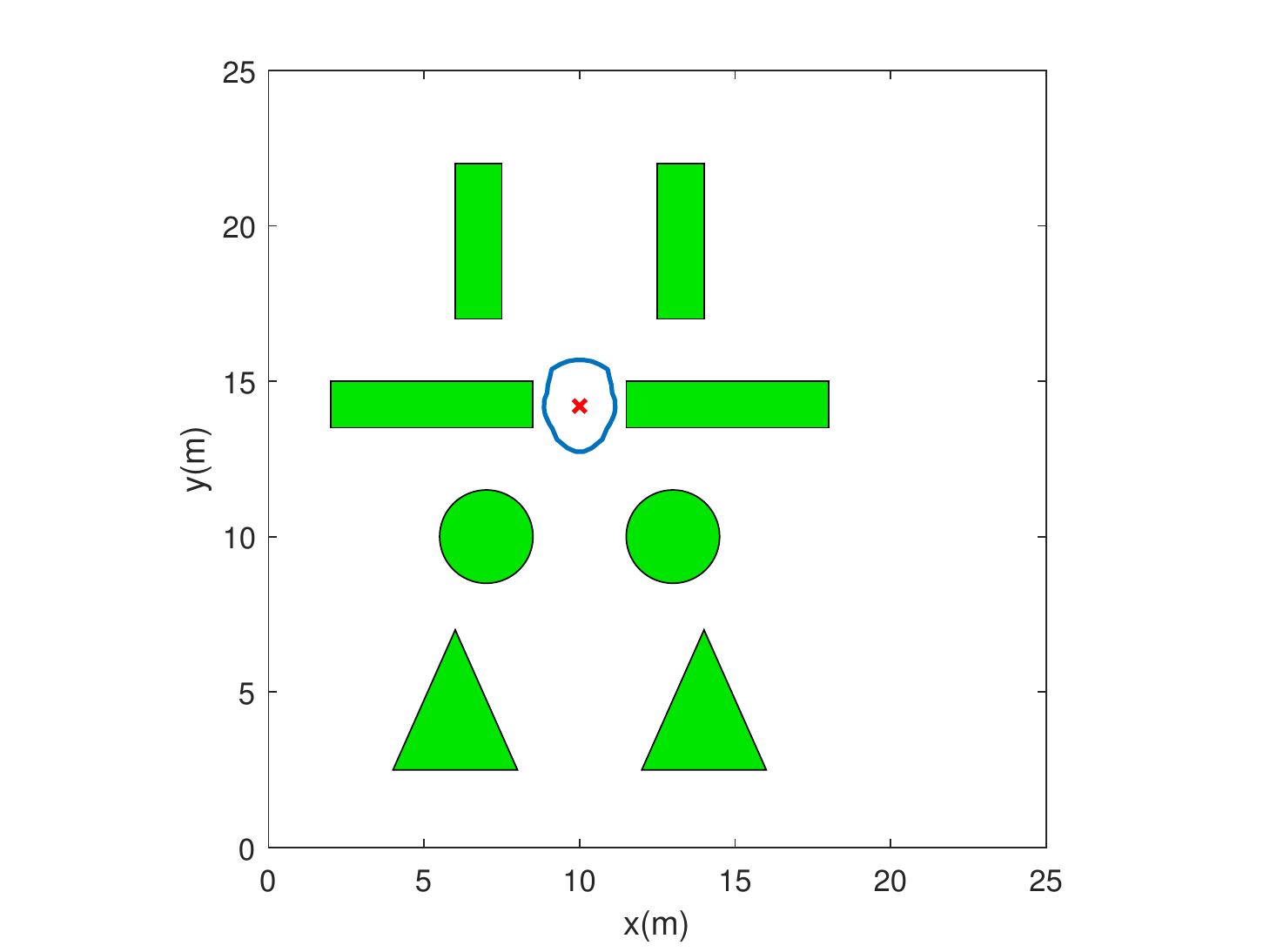}\\
			\centering{\scriptsize{(d)}}
		\end{minipage}
		\begin{minipage}[t]{0.32\linewidth}
			\includegraphics[width = 6.8cm, height = 5.8cm]{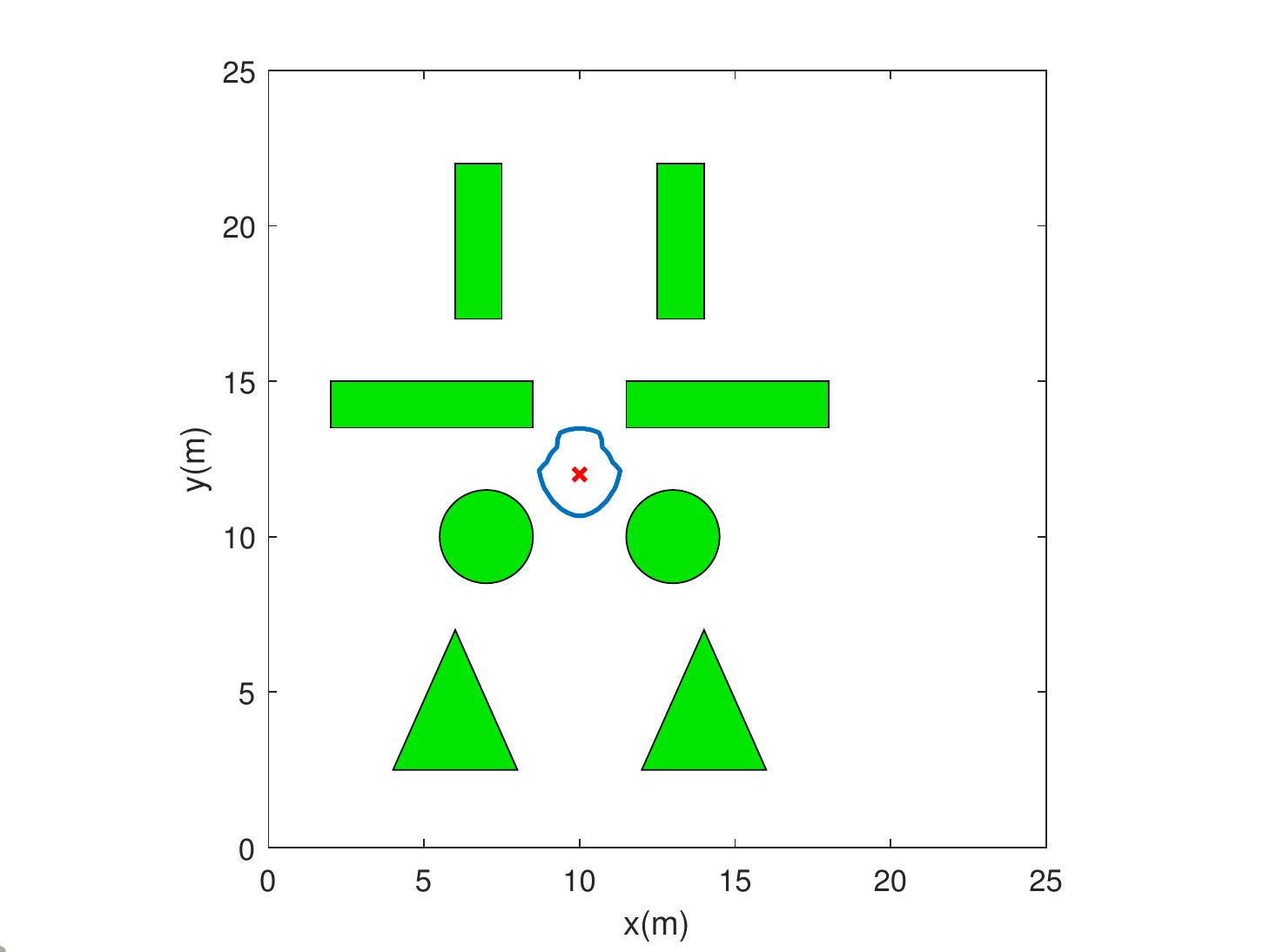}\\
			\centering{\scriptsize{(e)}}
		\end{minipage}
        \begin{minipage}[t]{0.32\linewidth}
			\includegraphics[width = 6.8cm, height = 5.8cm]{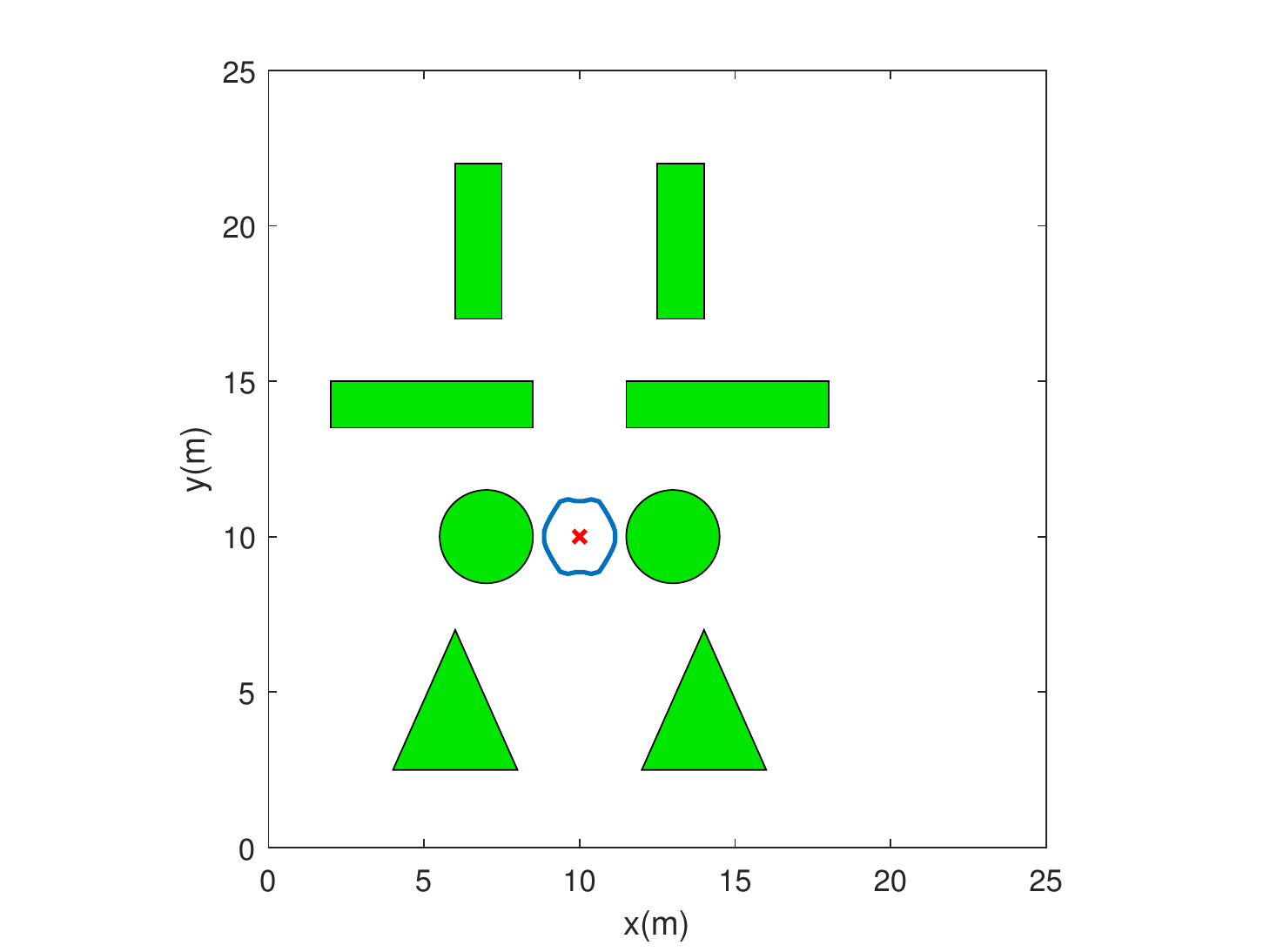}\\
			\centering{\scriptsize{(f)}}
		\end{minipage}
	\end{tabular}
    \hspace{0.5cm}
    \begin{tabular}{cc}	
        \begin{minipage}[t]{0.32\linewidth}
			\includegraphics[width = 6.8cm, height = 5.8cm]{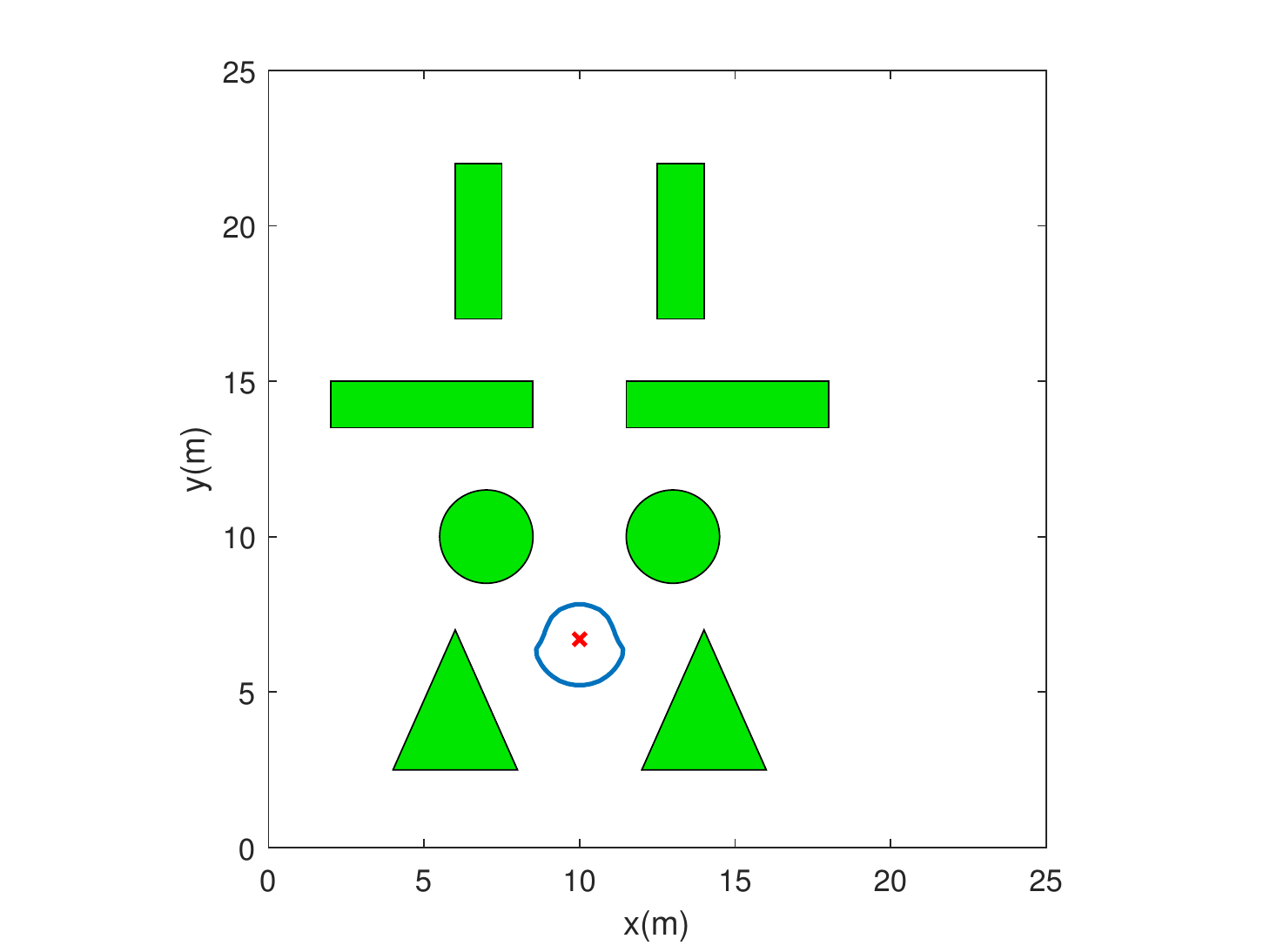}\\
			\centering{\scriptsize{(g)}}
		\end{minipage}
		\begin{minipage}[t]{0.32\linewidth}
			\includegraphics[width = 6.8cm, height = 5.8cm]{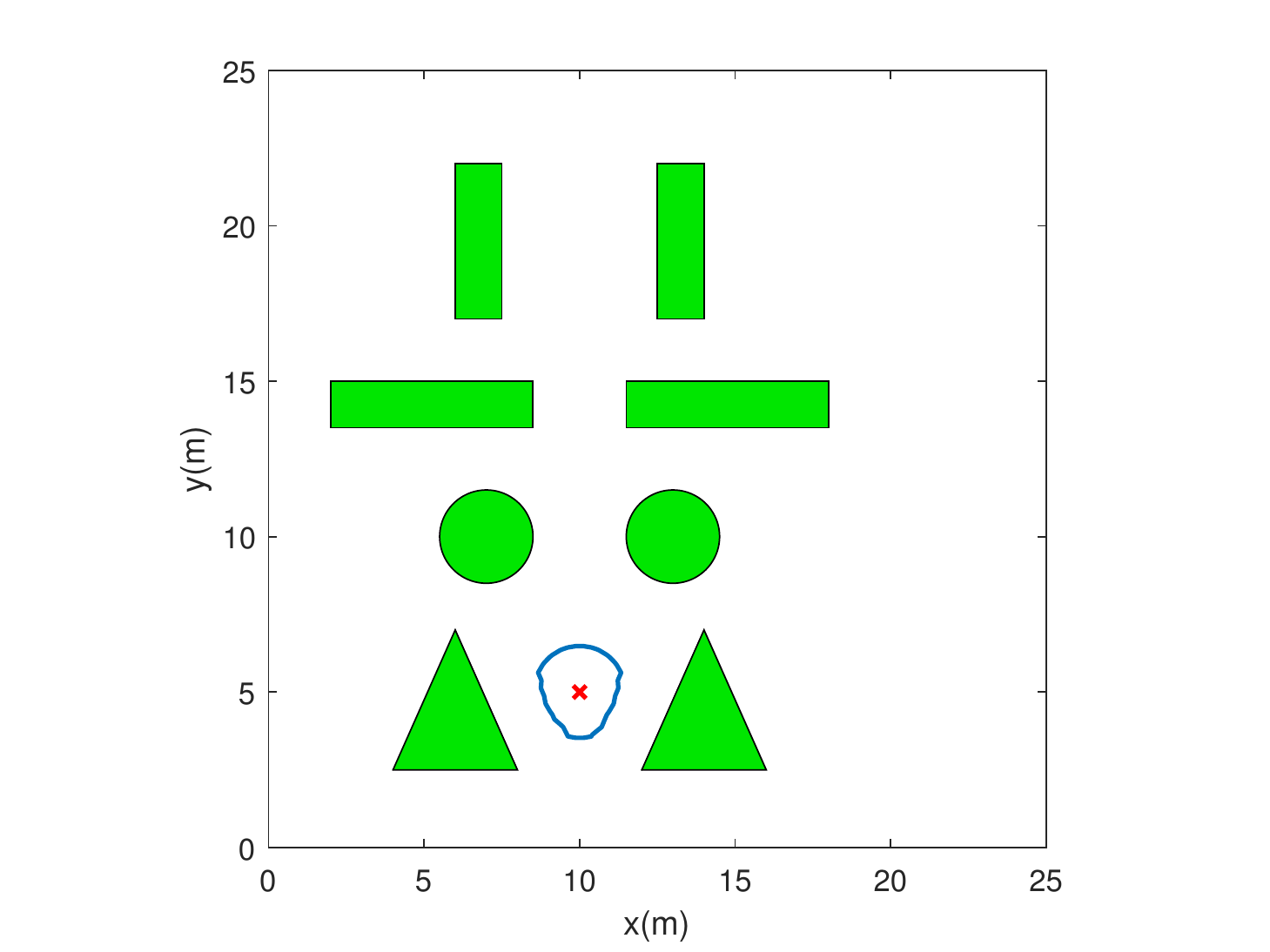}\\
			\centering{\scriptsize{(h)}}
		\end{minipage}
        \begin{minipage}[t]{0.32\linewidth}
			\includegraphics[width = 6.8cm, height = 5.8cm]{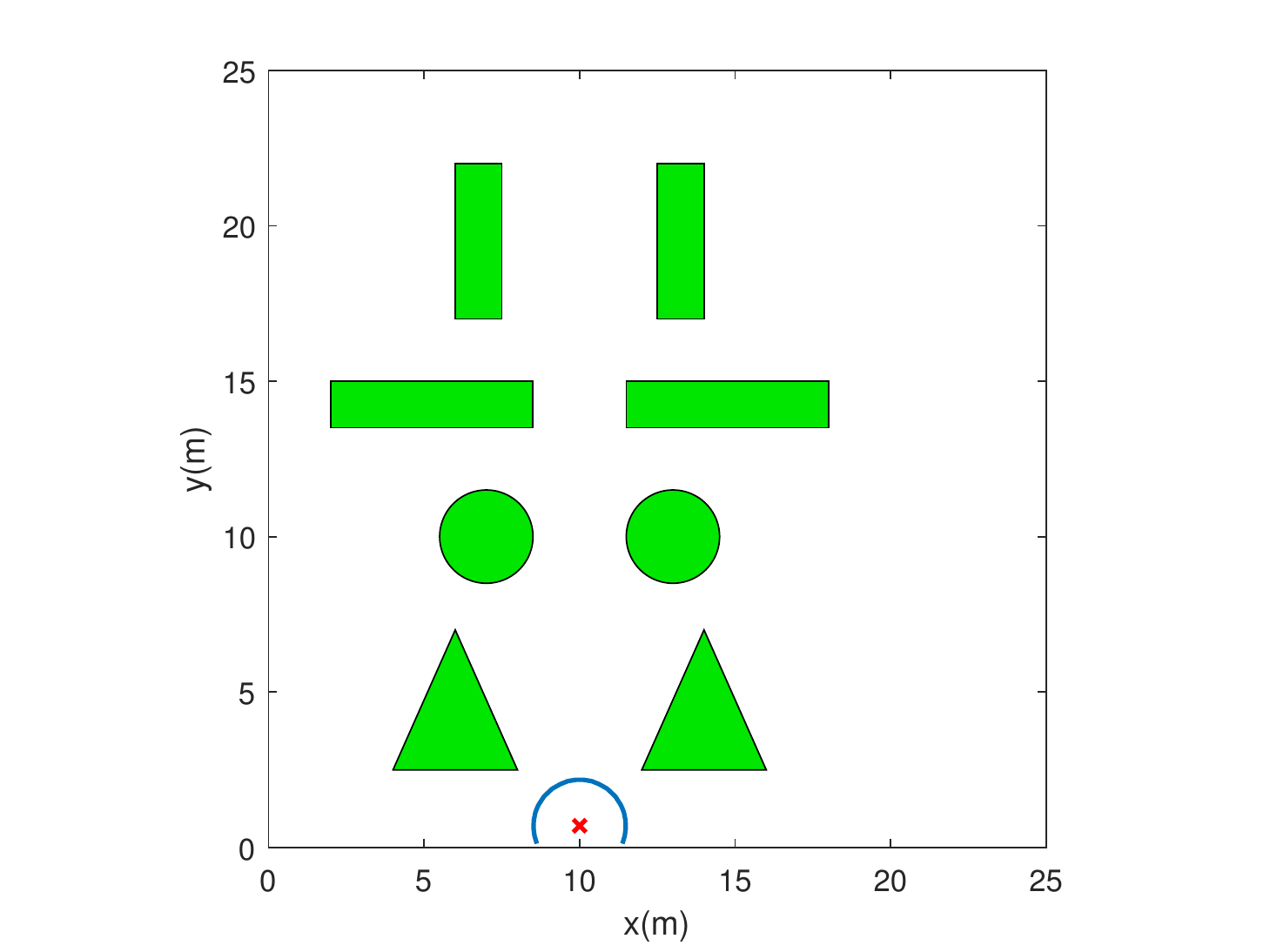}\\
			\centering{\scriptsize{(i)}}
		\end{minipage}
	\end{tabular}
    \hspace{0.5cm}
	\caption{\label{fig:case2_GPGRN_one} When a target pass through a compound channel, the swarm pattern of point A is formed by MOGP-NSGA-II, which encircles a target without colliding the channel. (a)-(e) show that the swarm pattern encircle the target in different shapes way, when the target passes through the channel.}
\end{figure*}

Fig. \ref{fig:case2_GPGRN_one} shows that the swarm pattern of point A that optimized to achieve by the proposed automatic design framework encircles a target across a channel without colliding the channel.  In
particular, when the target does not enter the channel, the swarm pattern is a circular shape way to encircle the target, as illustrated by Fig. \ref{fig:case2_GPGRN_one} (a) and (i).  Fig. \ref{fig:case2_GPGRN_one} (b)-(h) show the swarm pattern encircles the target in different shapes way. For example, when the target moves from the channel to the narrow channel, the shape of swarm pattern gradually changes from elliptocytosis to circular, as shown Fig. \ref{fig:case2_GPGRN_one} (b)-(c). As the target moves from the circular narrow channel to the T-channel, the shape of swarm pattern gradually becomes trapezoid shape, as shown Fig. \ref{fig:case2_GPGRN_one} (f)-(h). The reason for the dynamic change of swarm pattern with different target position is MOGP-NSGA-II can automatically optimize the optimal GRN-based model according to different restricted  environments.

\begin{figure*}[ht]
	\begin{tabular}{cc}
		\begin{minipage}[t]{0.32\linewidth}  %  width = 4.5cm,height = 3.6cm
			\includegraphics[width = 6.8cm, height = 5.8cm]{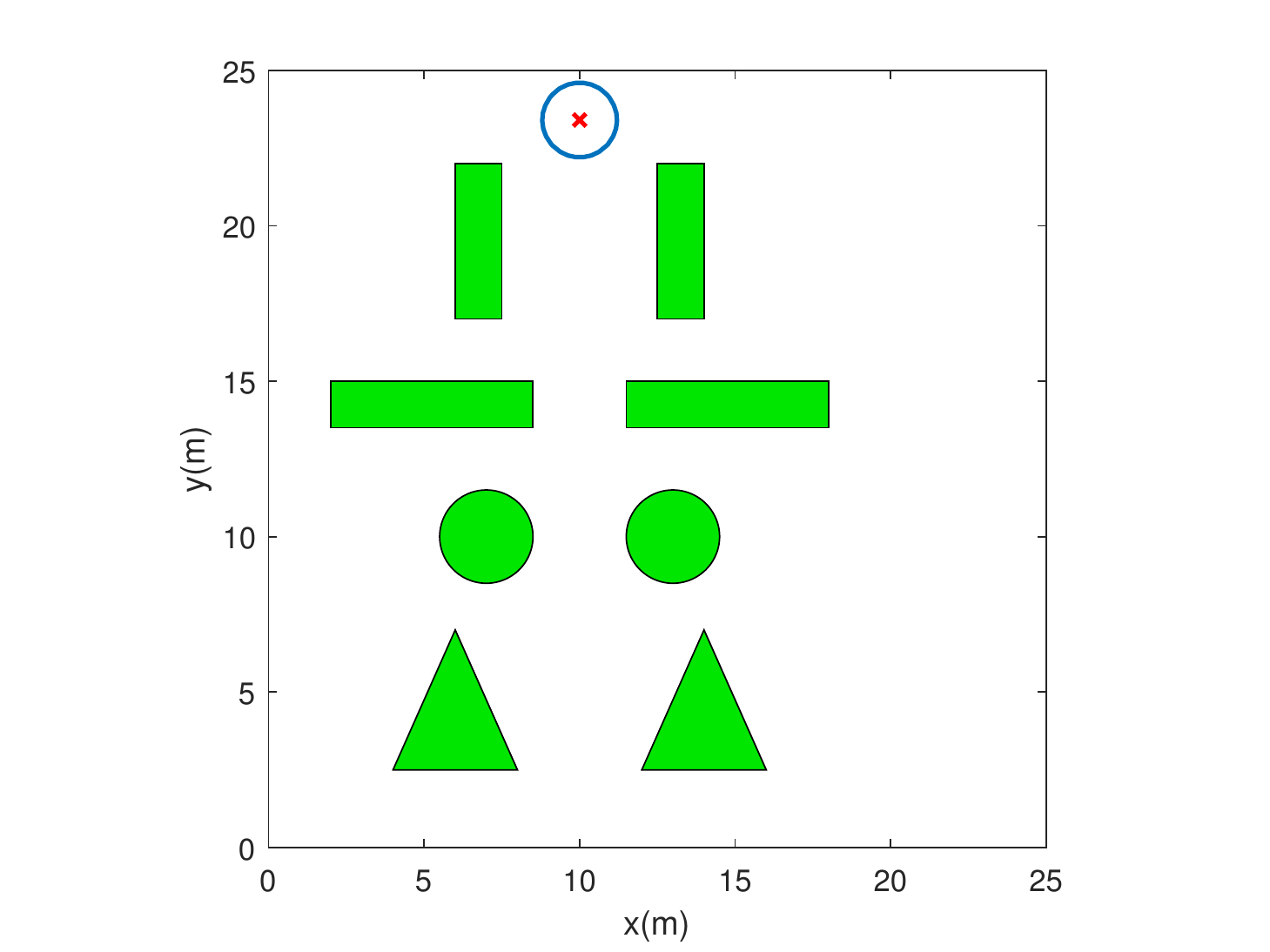}\\
			\centering{\scriptsize{(a)}}
		\end{minipage}
        \begin{minipage}[t]{0.32\linewidth}
			\includegraphics[width = 6.8cm, height = 5.8cm]{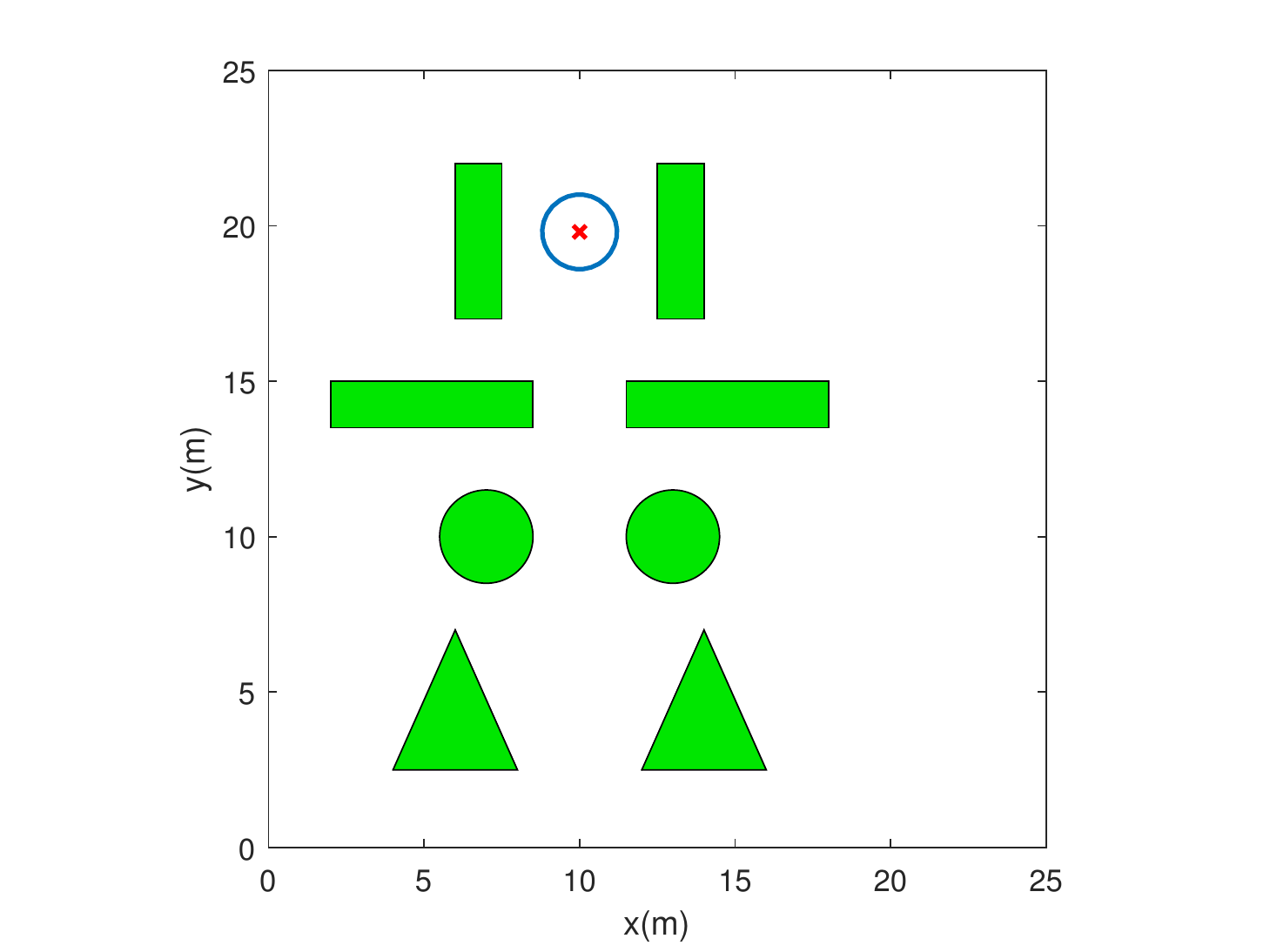}\\
			\centering{\scriptsize{(b)}}
		\end{minipage}
        \begin{minipage}[t]{0.32\linewidth}
			\includegraphics[width = 6.8cm, height = 5.8cm]{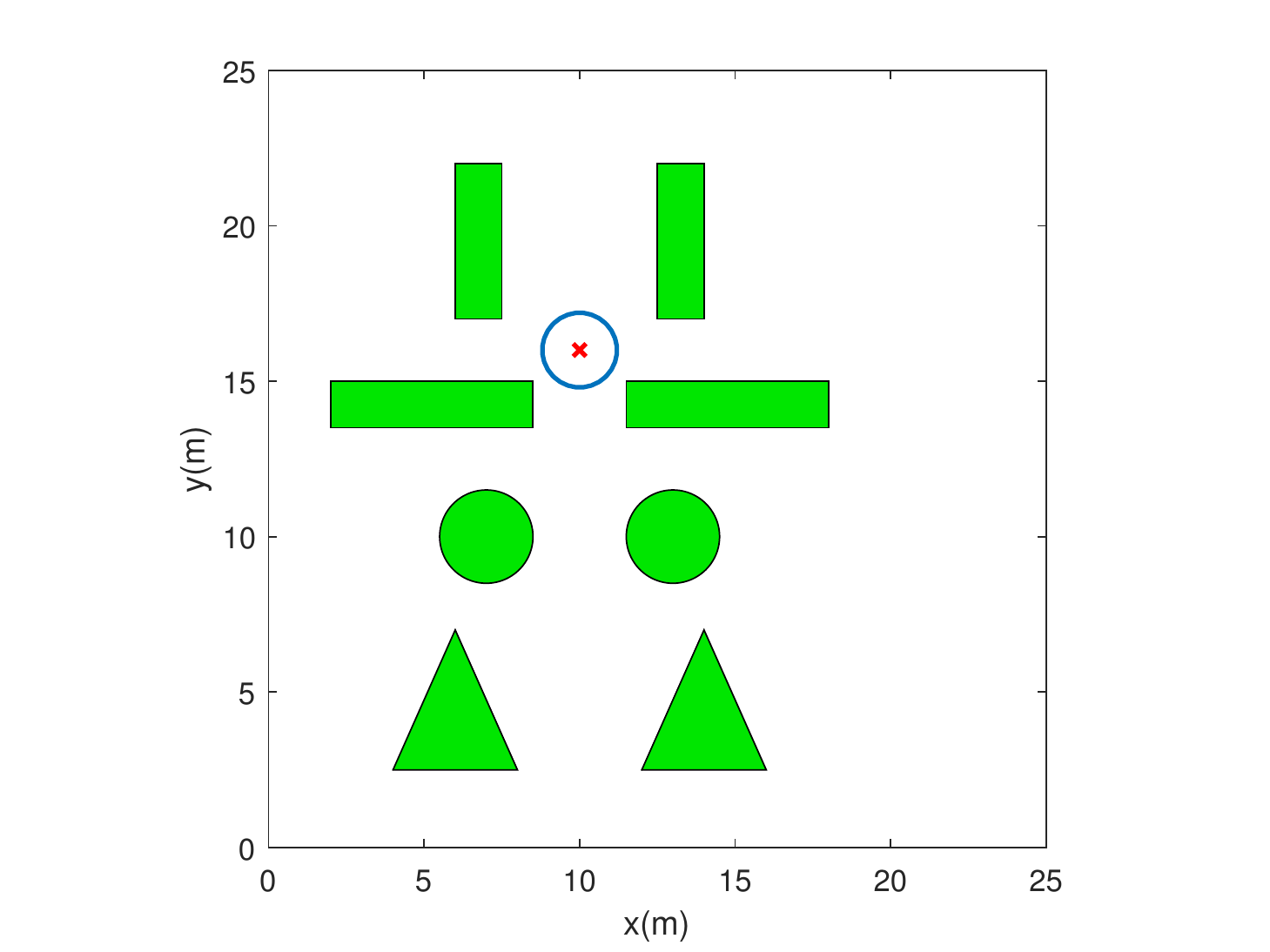}\\
			\centering{\scriptsize{(c)}}
		\end{minipage}
	\end{tabular}
	\hspace{0.5cm}
    \begin{tabular}{cc}	
        \begin{minipage}[t]{0.32\linewidth}
			\includegraphics[width = 6.8cm, height = 5.8cm]{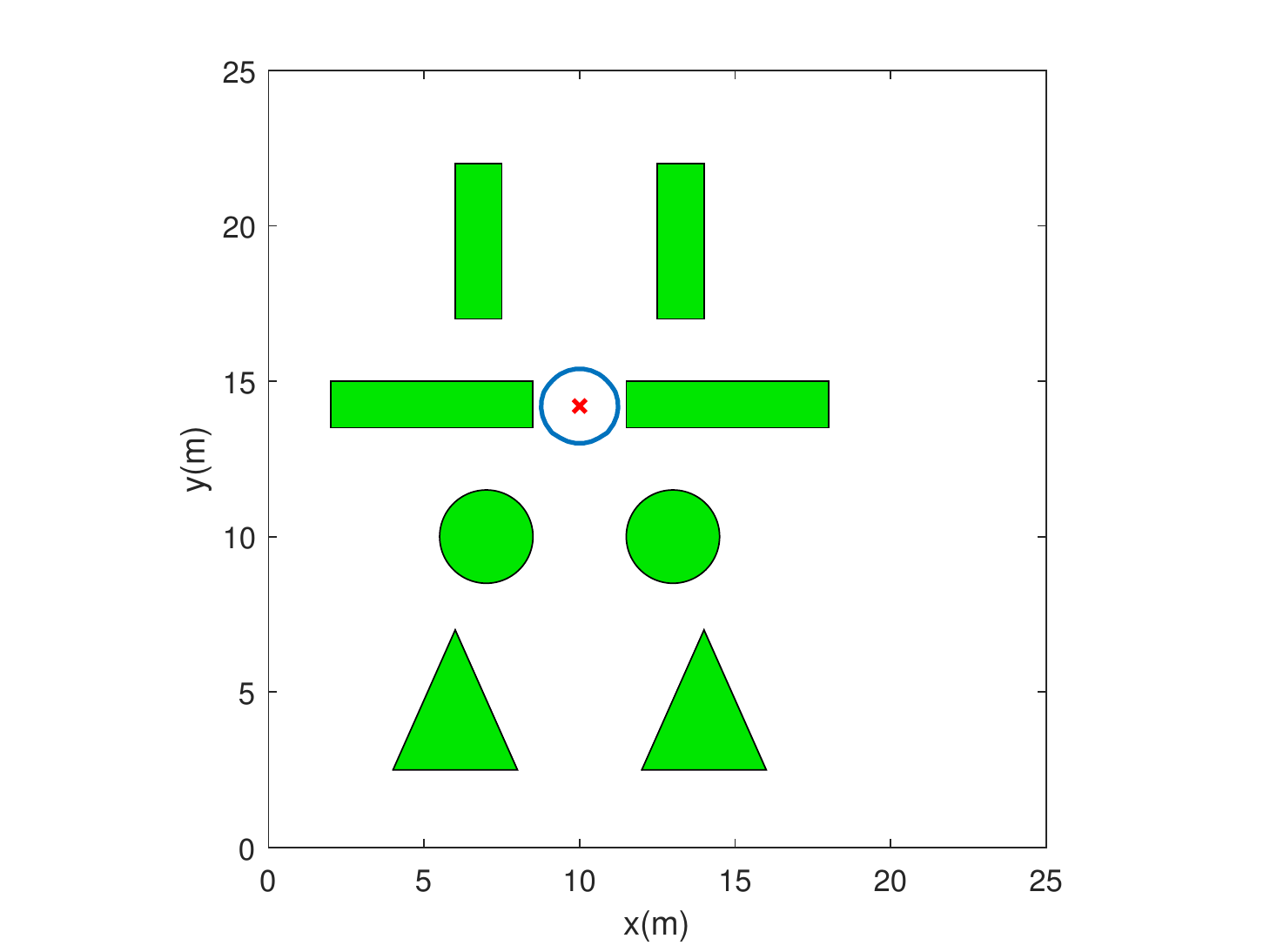}\\
			\centering{\scriptsize{(d)}}
		\end{minipage}
		\begin{minipage}[t]{0.32\linewidth}
			\includegraphics[width = 6.8cm, height = 5.8cm]{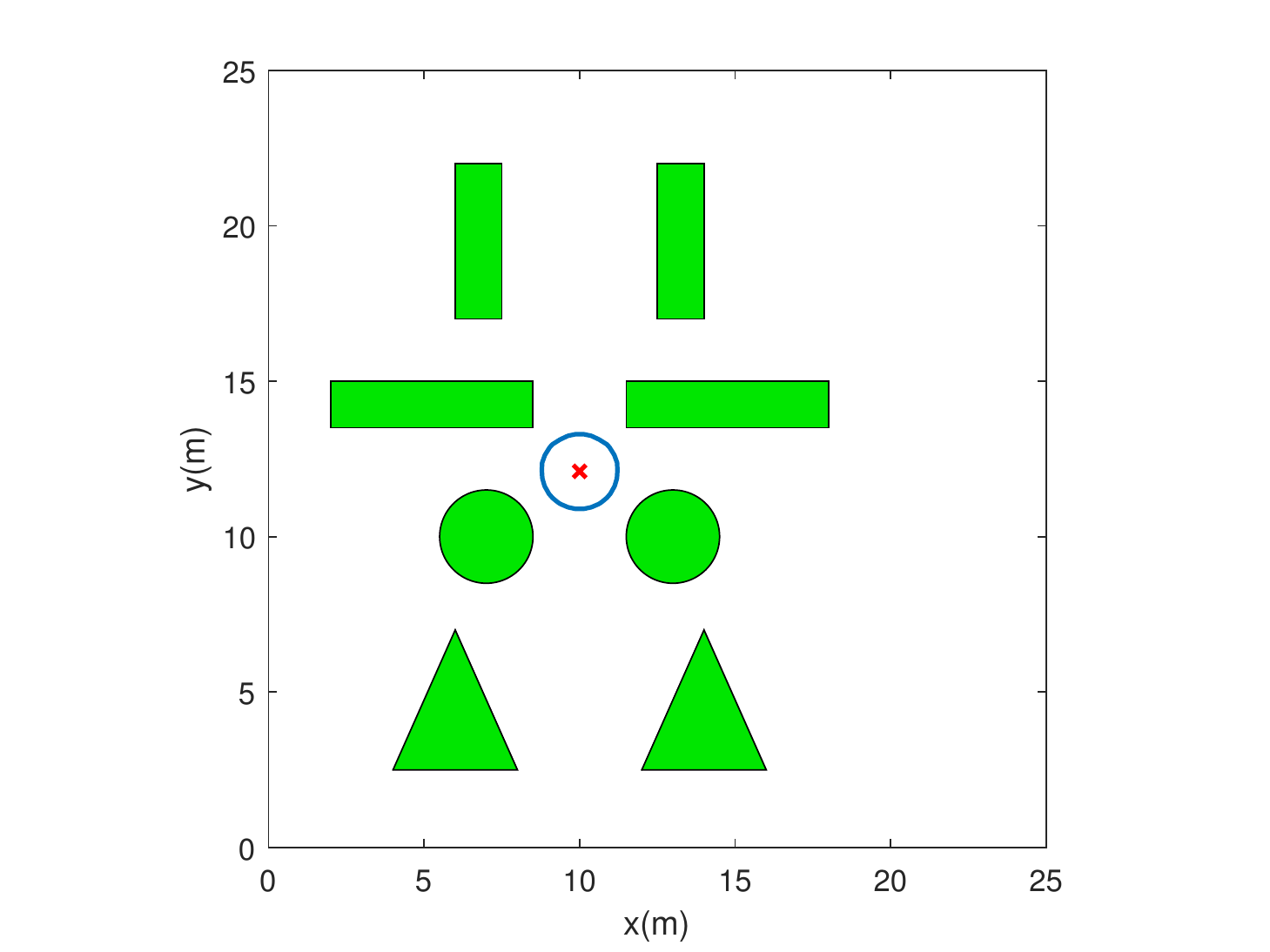}\\
			\centering{\scriptsize{(e)}}
		\end{minipage}
        \begin{minipage}[t]{0.32\linewidth}
			\includegraphics[width = 6.8cm, height = 5.8cm]{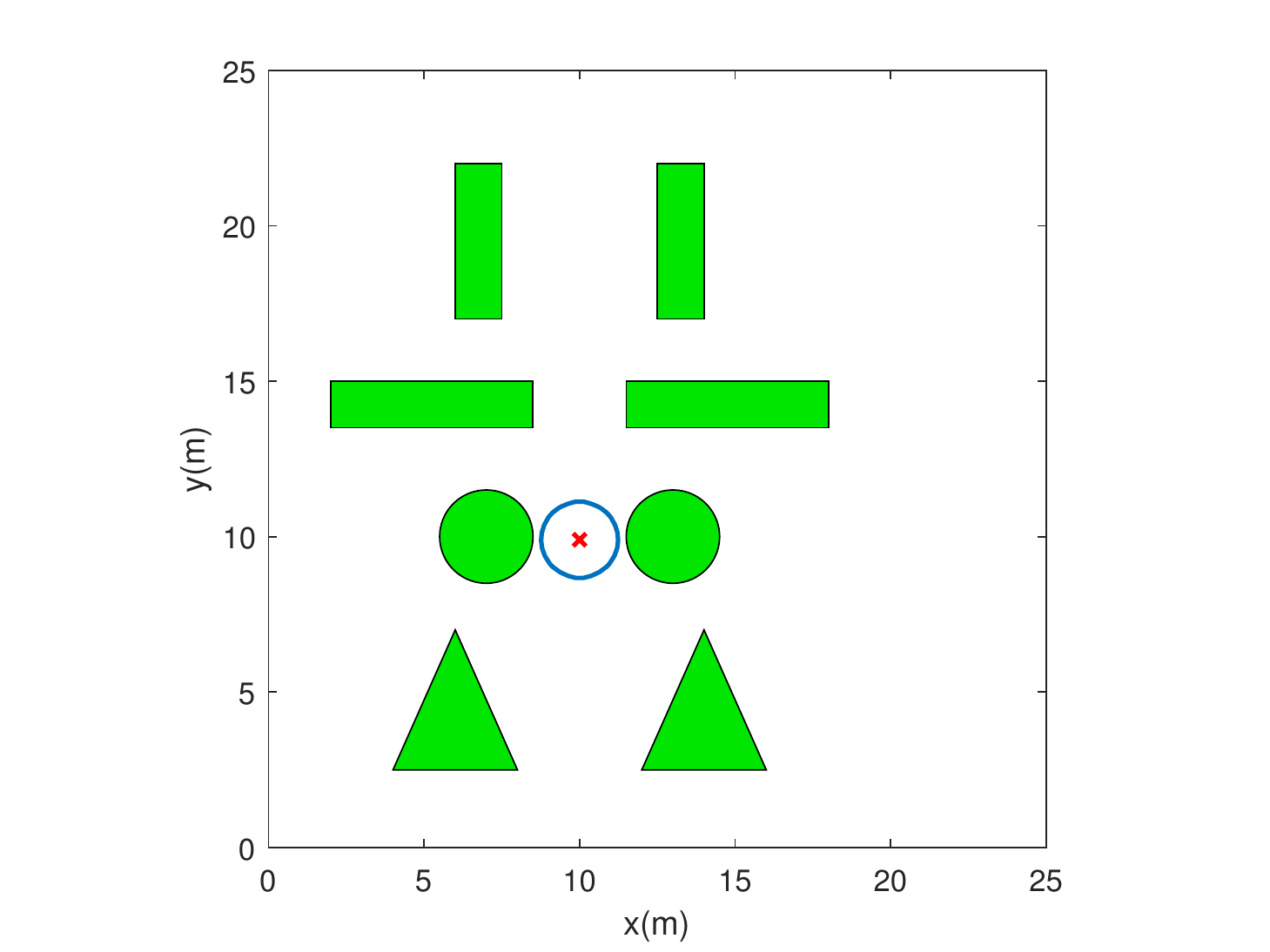}\\
			\centering{\scriptsize{(f)}}
		\end{minipage}
	\end{tabular}
    \hspace{0.5cm}
    \begin{tabular}{cc}	
        \begin{minipage}[t]{0.32\linewidth}
			\includegraphics[width = 6.8cm, height = 5.8cm]{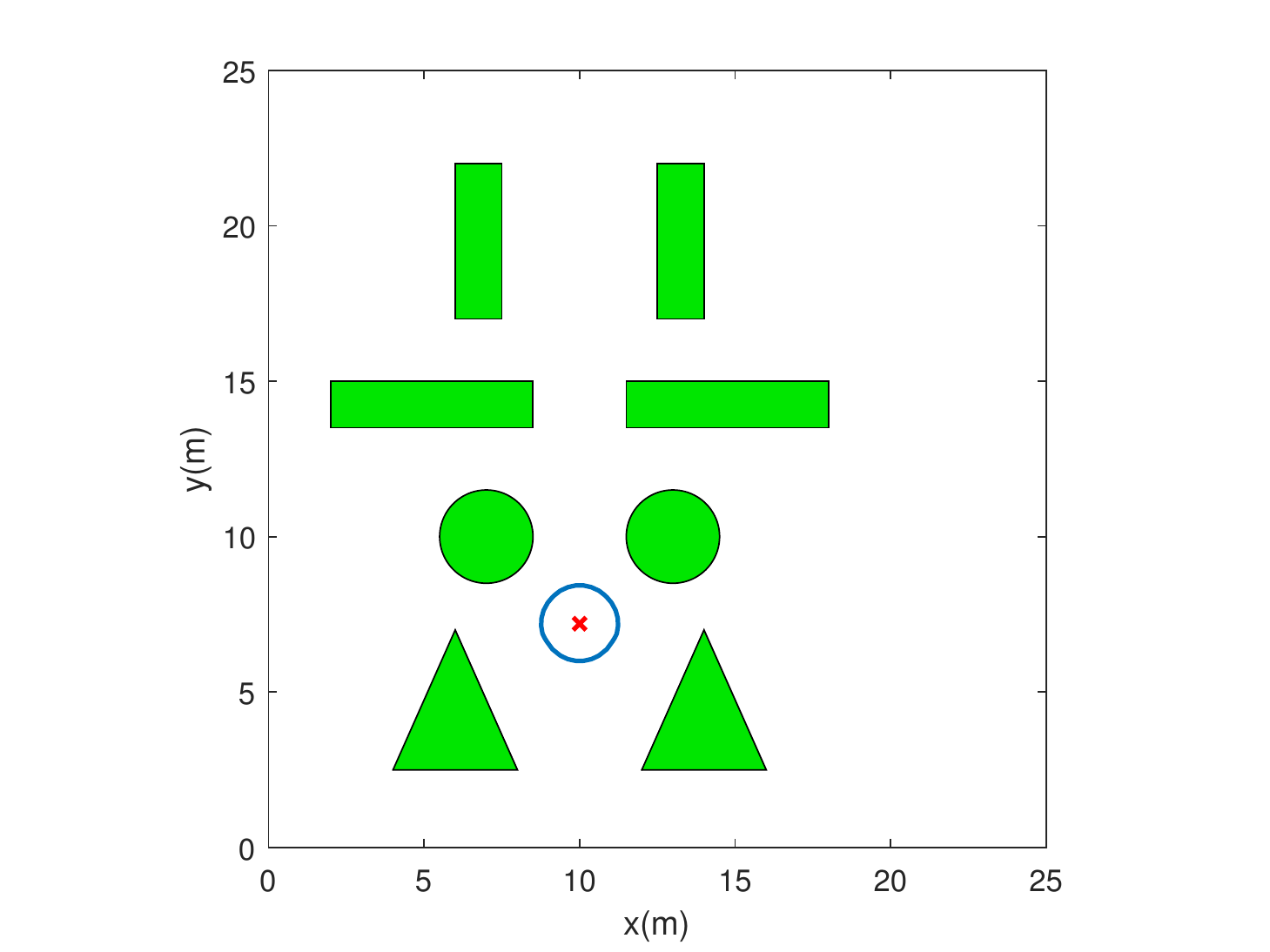}\\
			\centering{\scriptsize{(g)}}
		\end{minipage}
		\begin{minipage}[t]{0.32\linewidth}
			\includegraphics[width = 6.8cm, height = 5.8cm]{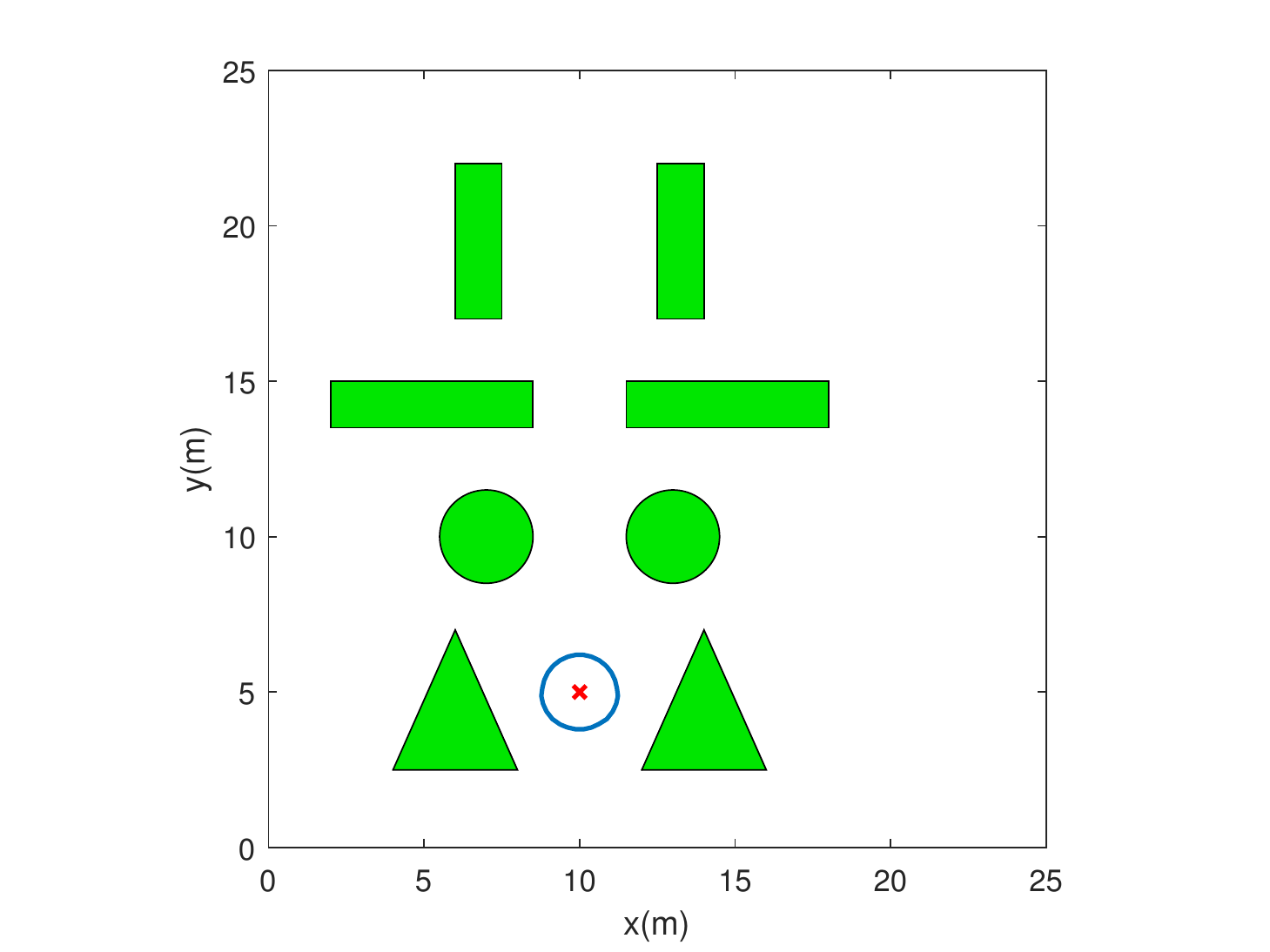}\\
			\centering{\scriptsize{(h)}}
		\end{minipage}
        \begin{minipage}[t]{0.32\linewidth}
			\includegraphics[width = 6.8cm, height = 5.8cm]{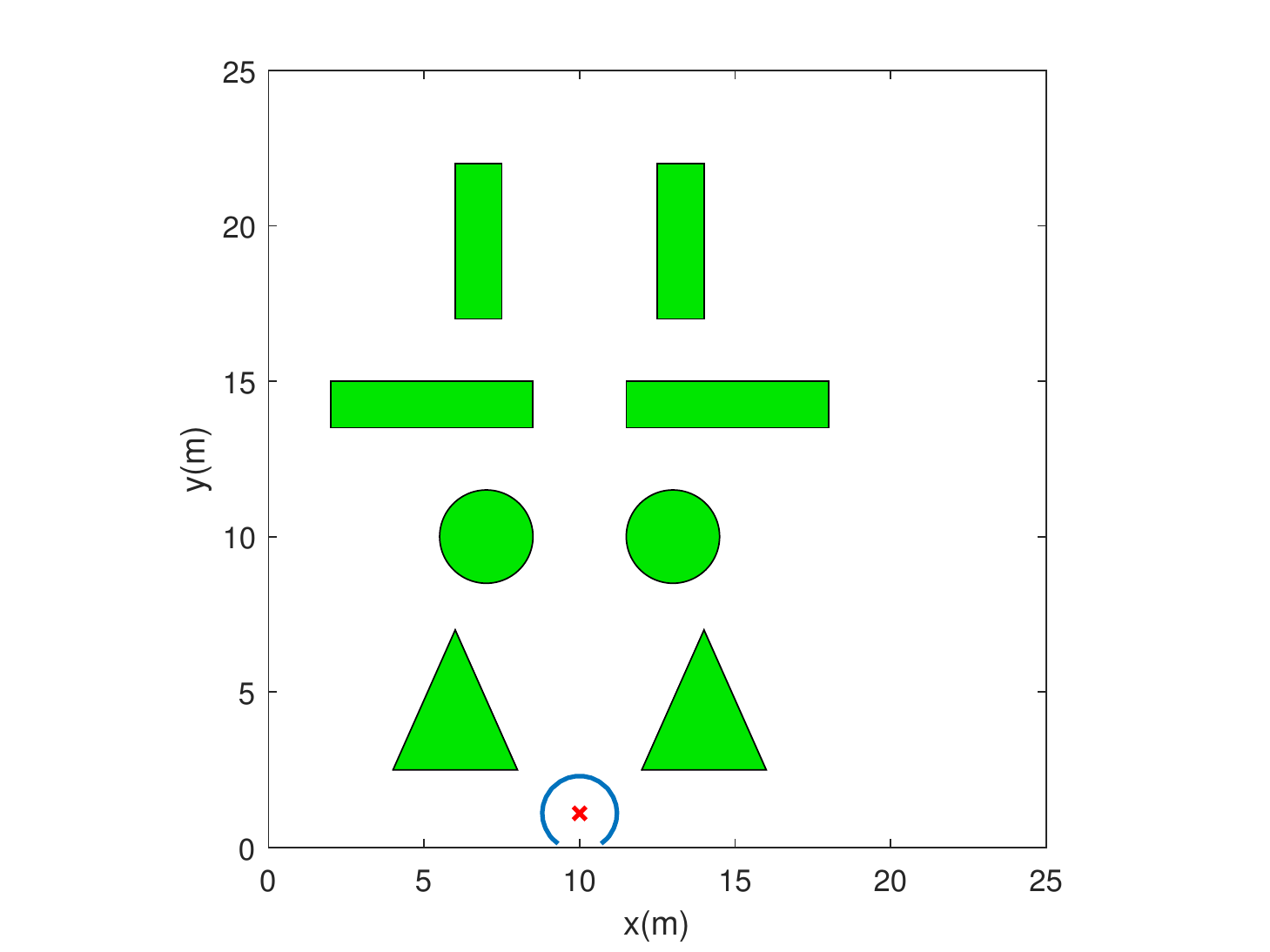}\\
			\centering{\scriptsize{(i)}}
		\end{minipage}
	\end{tabular}
    \hspace{0.5cm}
	\caption{\label{fig:case2_GPGRN_two} When a target pass through a compound channel, the swarm pattern of point B is formed by MOGP-NSGA-II, which encircles a target without colliding the channel. (a)-(e) show that the swarm pattern encircle the target in a  shapes way, when the target passes through the channel.}
\end{figure*}

Fig. \ref{fig:case2_GPGRN_two} shows that the swarm pattern of point B that optimized to achieve by MOGP-NSGA-II encircles a target across a compound channel without colliding the channel. Fig. \ref{fig:case2_GPGRN_two} (a)-(i) show when a target passes through the compound channel, the swarm pattern is a circular shape way to encircle the target. This is because the width of the narrow channel and the width of the circular narrow channel are the narrowest part of the composite channel In Fig. \ref{fig:case2_GPGRN_two} (c)-(g). As long as the swarm pattern can pass through the two parts in a certain shape, it can also pass through the other parts of the composite channel in this shape. The optimized swarm pattern can pass through the narrowest part of the composite channel in a circle shape, so the circle shape can also be used to encircle the target in other parts.

\begin{figure*}[ht]
	\begin{tabular}{cc}
		\begin{minipage}[t]{0.32\linewidth}  %  width = 4.5cm,height = 3.6cm
			\includegraphics[width = 6.8cm, height = 5.8cm]{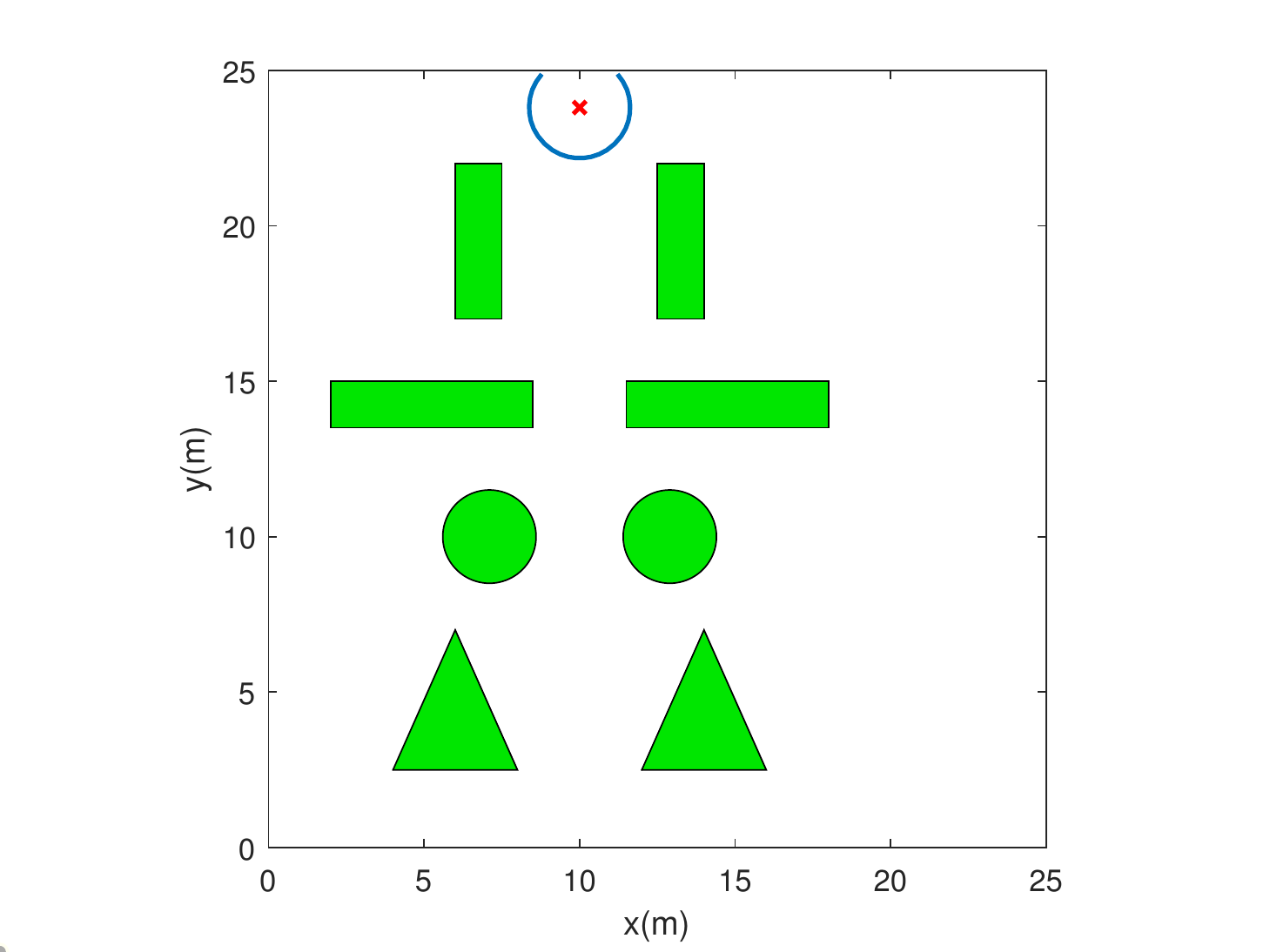}\\
			\centering{\scriptsize{(a)}}
		\end{minipage}
        \begin{minipage}[t]{0.32\linewidth}
			\includegraphics[width = 6.8cm, height = 5.8cm]{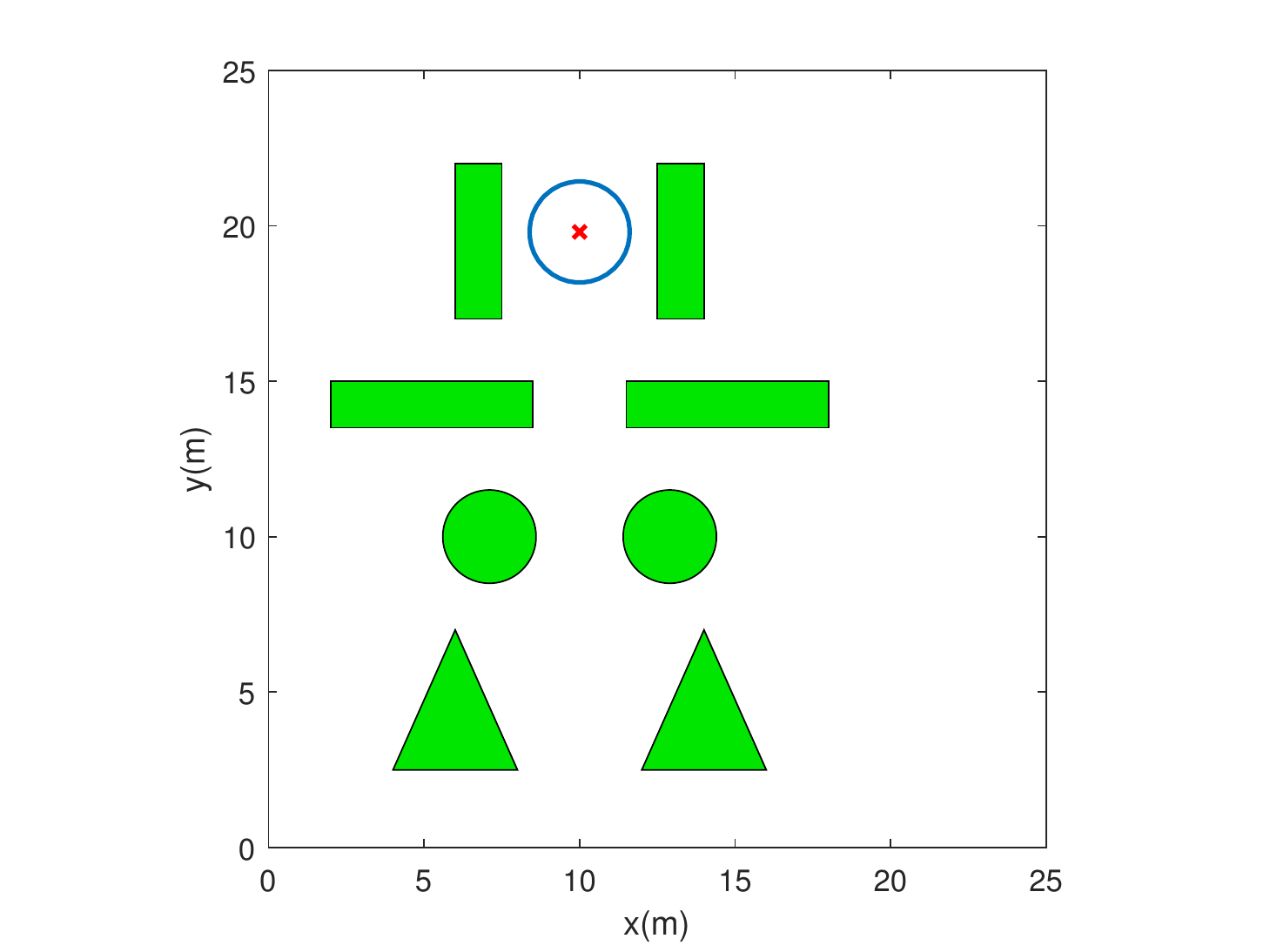}\\
			\centering{\scriptsize{(b)}}
		\end{minipage}
        \begin{minipage}[t]{0.32\linewidth}
			\includegraphics[width = 6.8cm, height = 5.8cm]{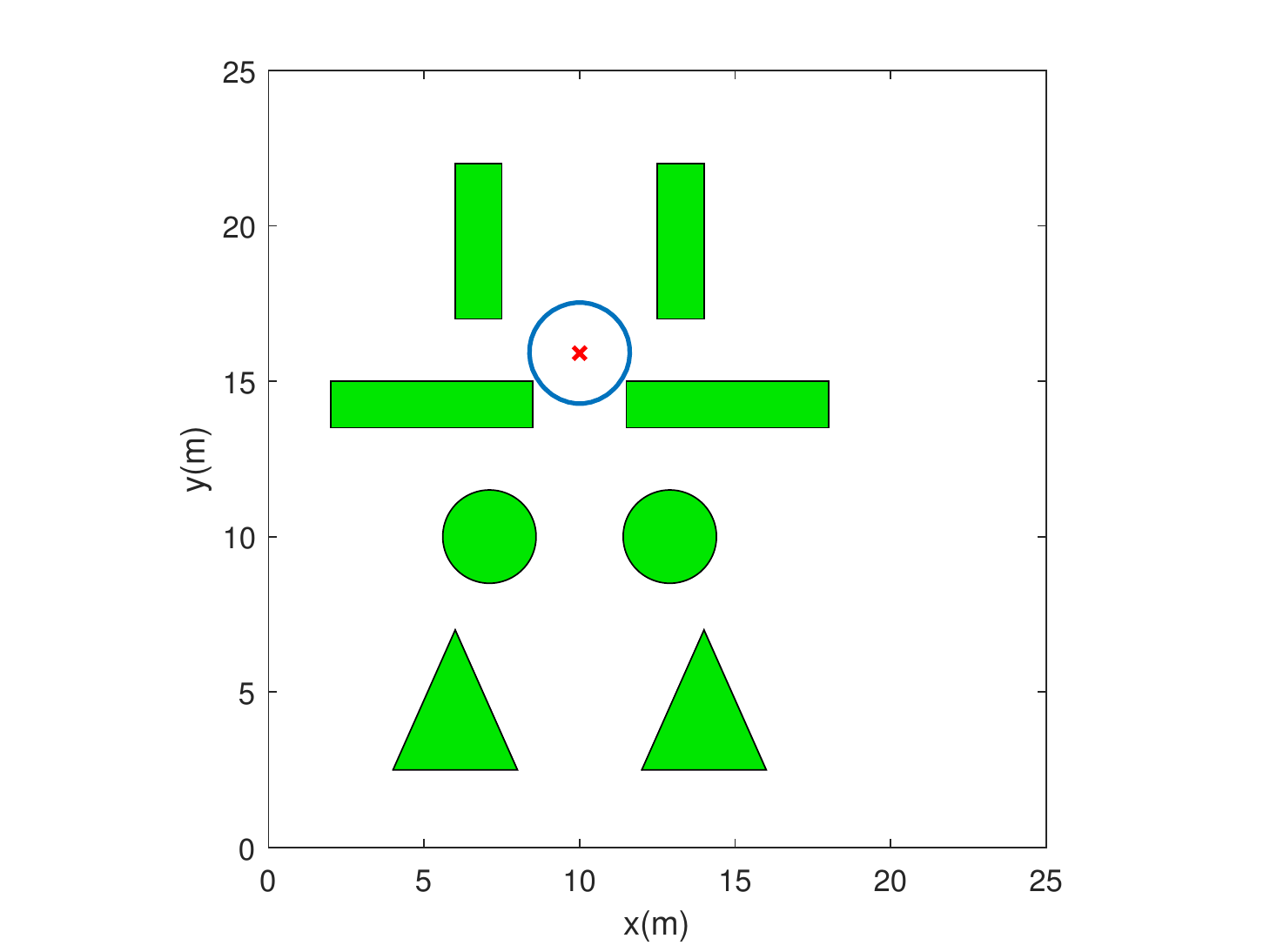}\\
			\centering{\scriptsize{(c)}}
		\end{minipage}
	\end{tabular}
	\hspace{0.5cm}
    \begin{tabular}{cc}	
        \begin{minipage}[t]{0.32\linewidth}
			\includegraphics[width = 6.8cm, height = 5.8cm]{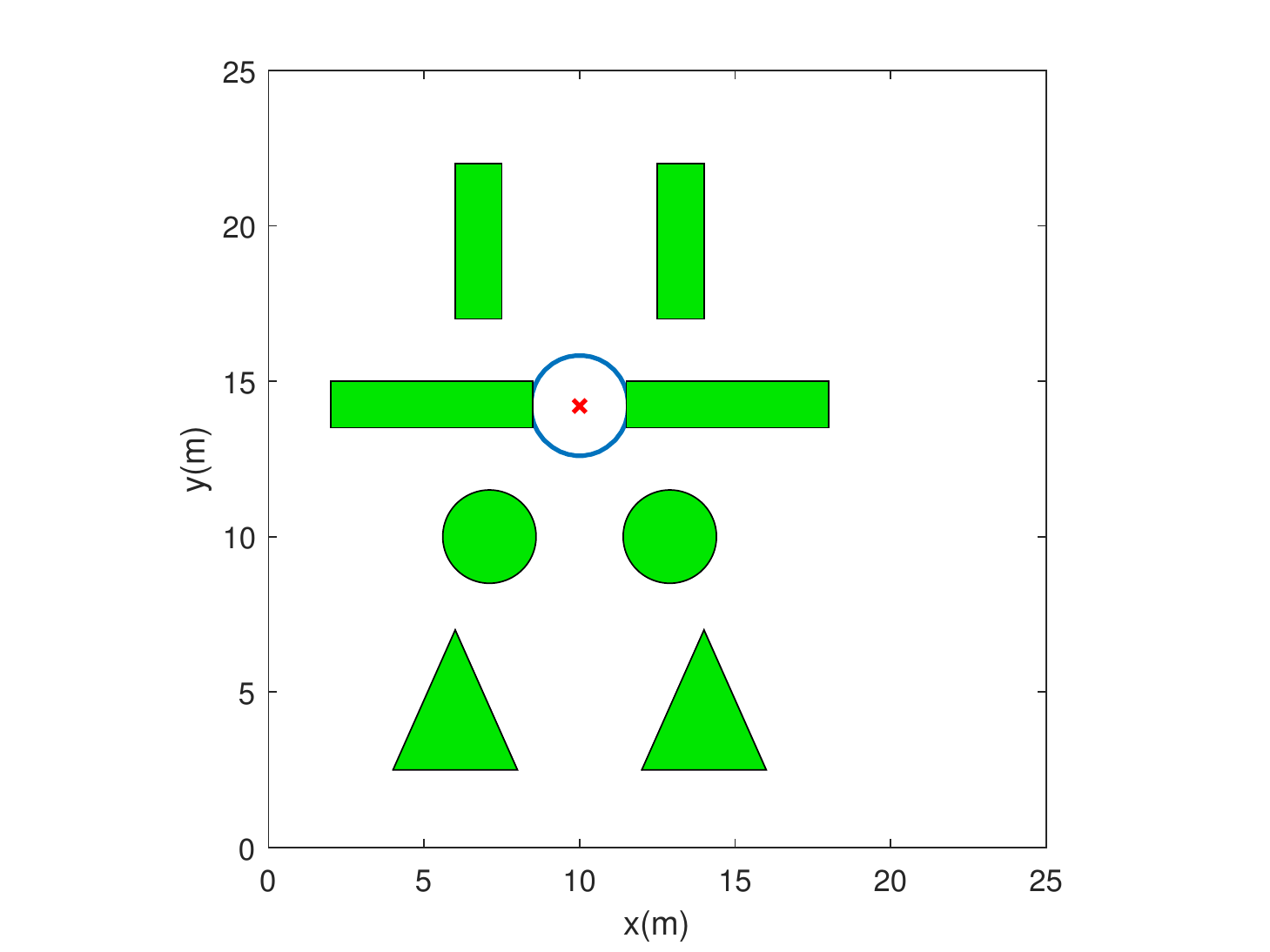}\\
			\centering{\scriptsize{(d)}}
		\end{minipage}
		\begin{minipage}[t]{0.32\linewidth}
			\includegraphics[width = 6.8cm, height = 5.8cm]{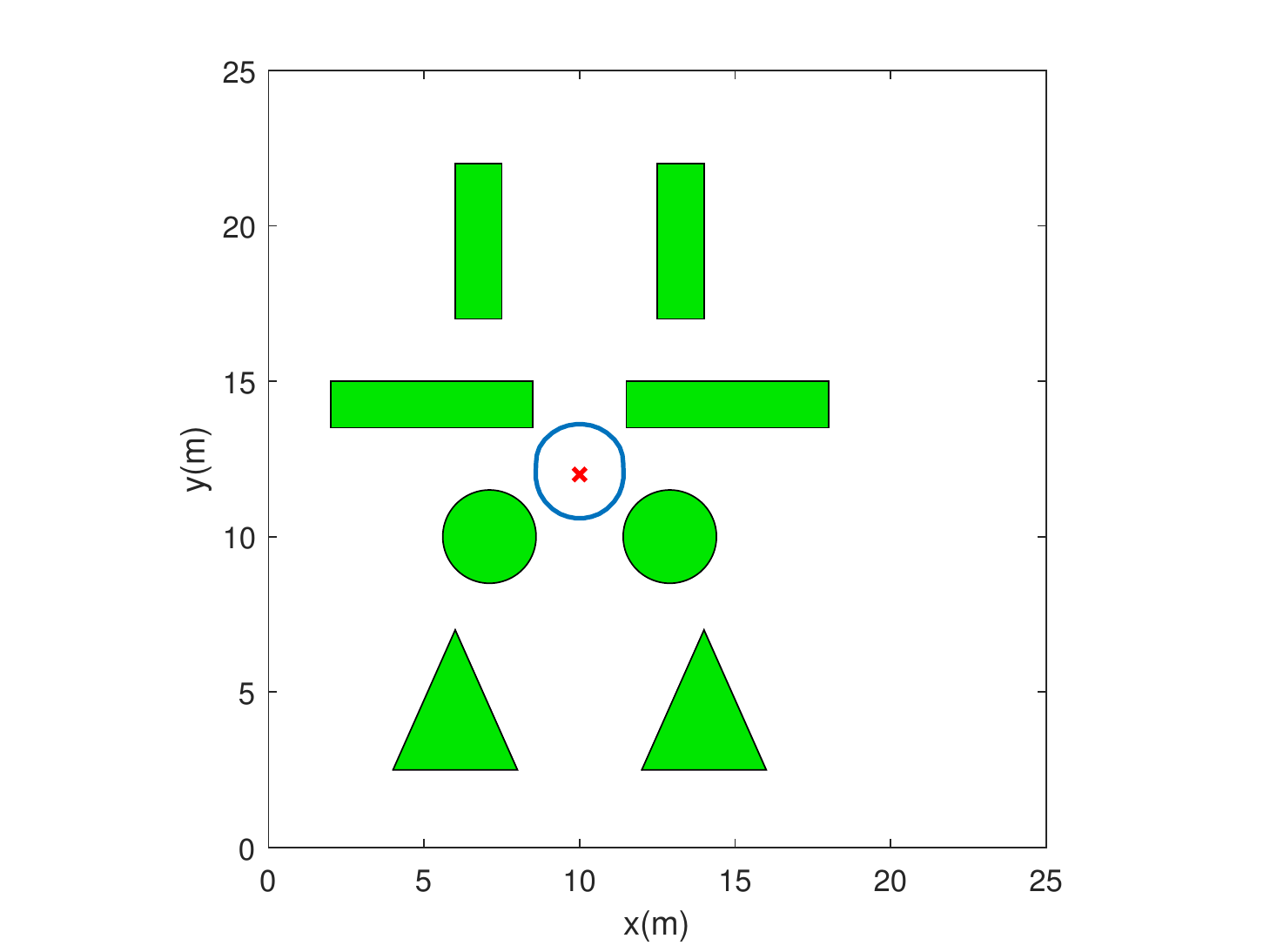}\\
			\centering{\scriptsize{(e)}}
		\end{minipage}
        \begin{minipage}[t]{0.32\linewidth}
			\includegraphics[width = 6.8cm, height = 5.8cm]{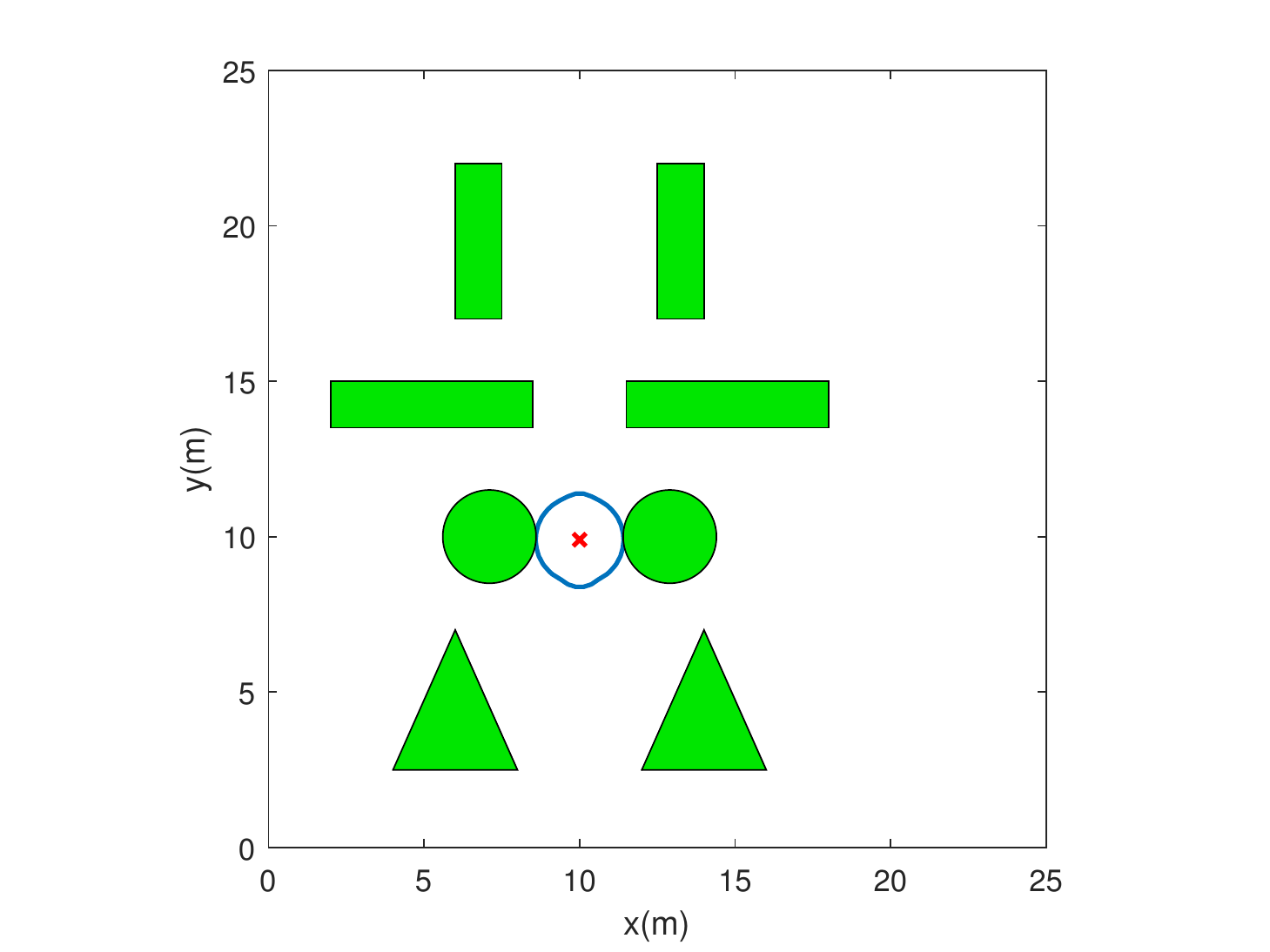}\\
			\centering{\scriptsize{(f)}}
		\end{minipage}
	\end{tabular}
    \hspace{0.5cm}
    \begin{tabular}{cc}	
        \begin{minipage}[t]{0.32\linewidth}
			\includegraphics[width = 6.8cm, height = 5.8cm]{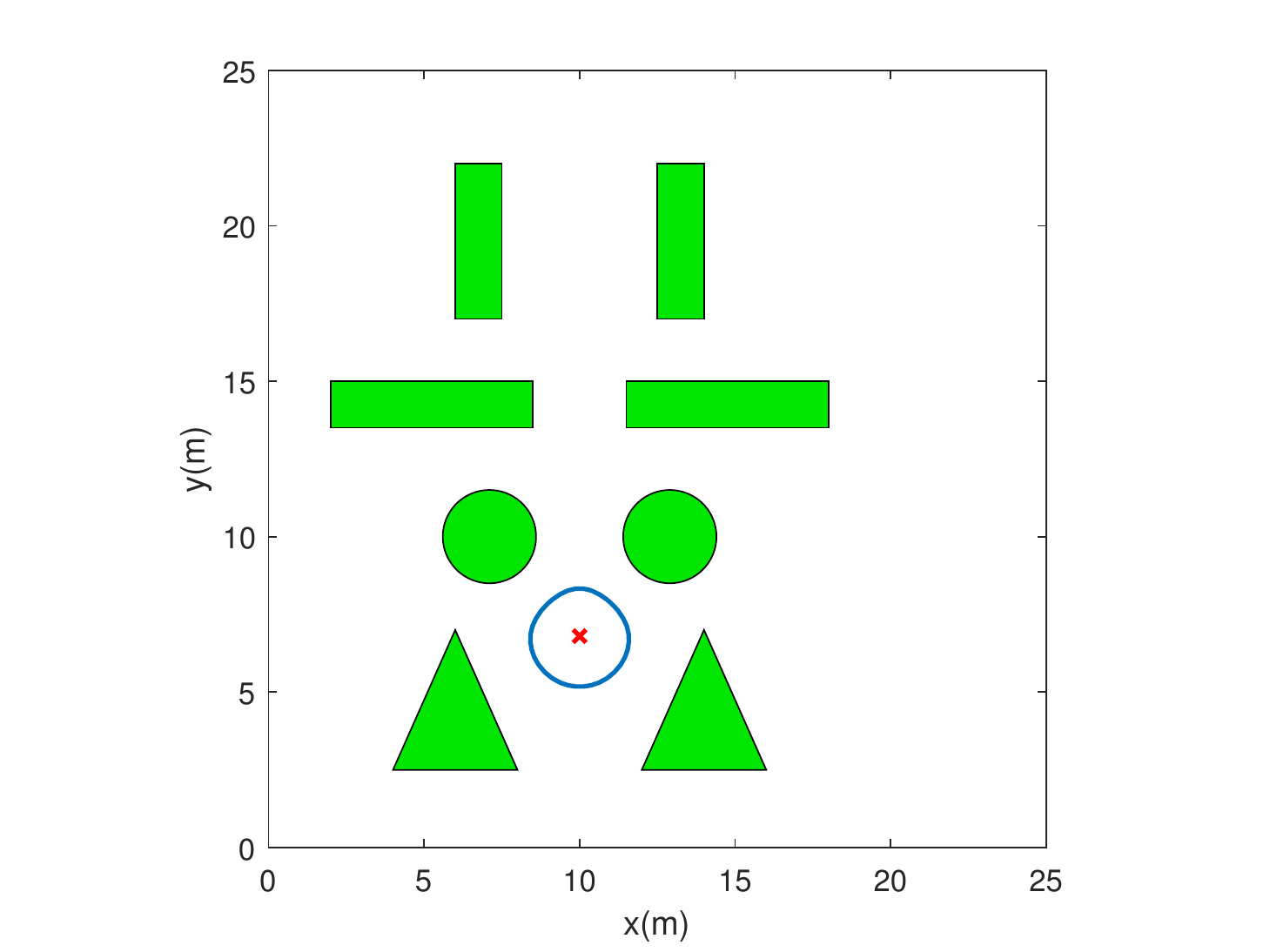}\\
			\centering{\scriptsize{(g)}}
		\end{minipage}
		\begin{minipage}[t]{0.32\linewidth}
			\includegraphics[width = 6.8cm, height = 5.8cm]{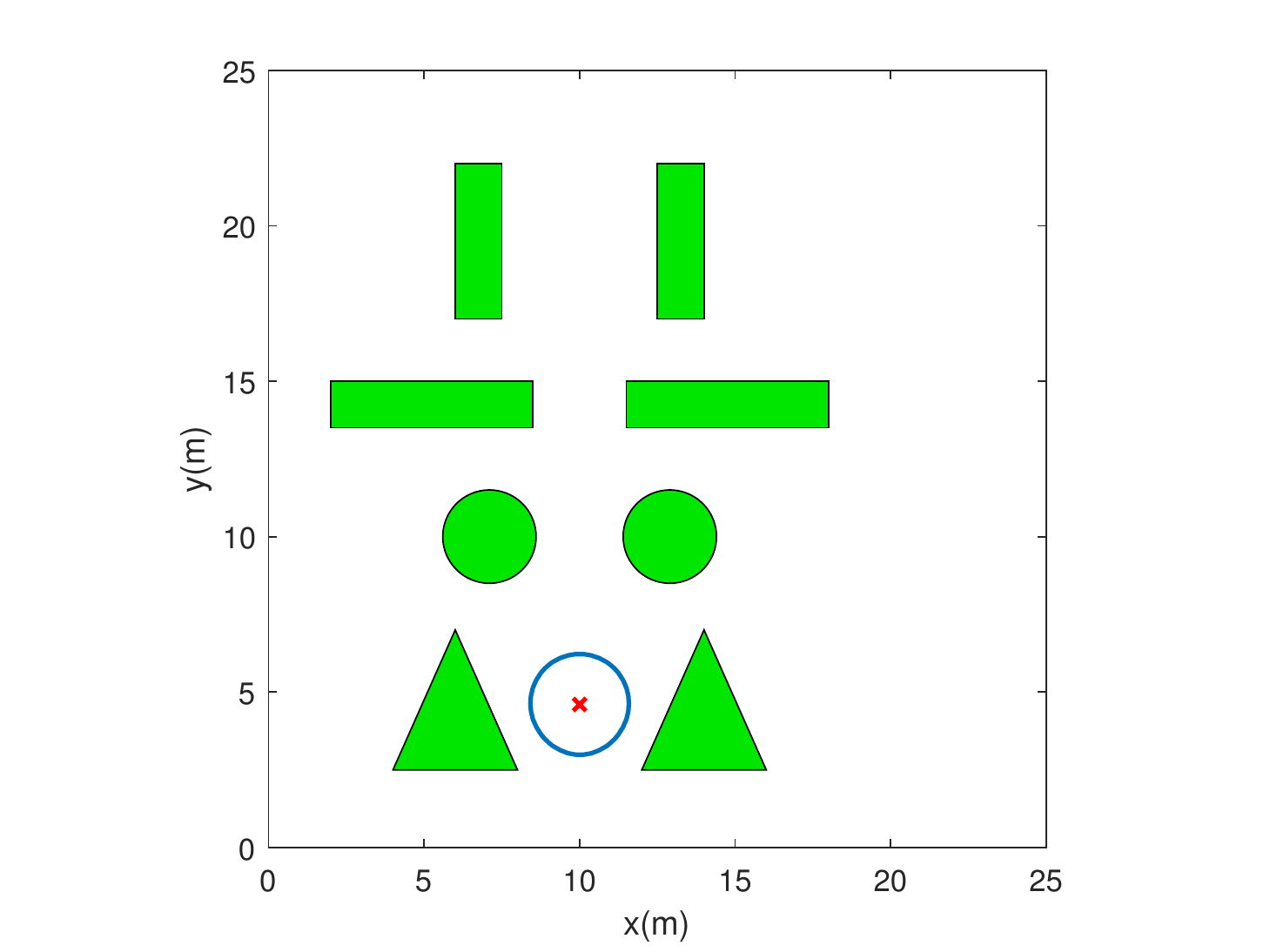}\\
			\centering{\scriptsize{(h)}}
		\end{minipage}
        \begin{minipage}[t]{0.32\linewidth}
			\includegraphics[width = 6.8cm, height = 5.8cm]{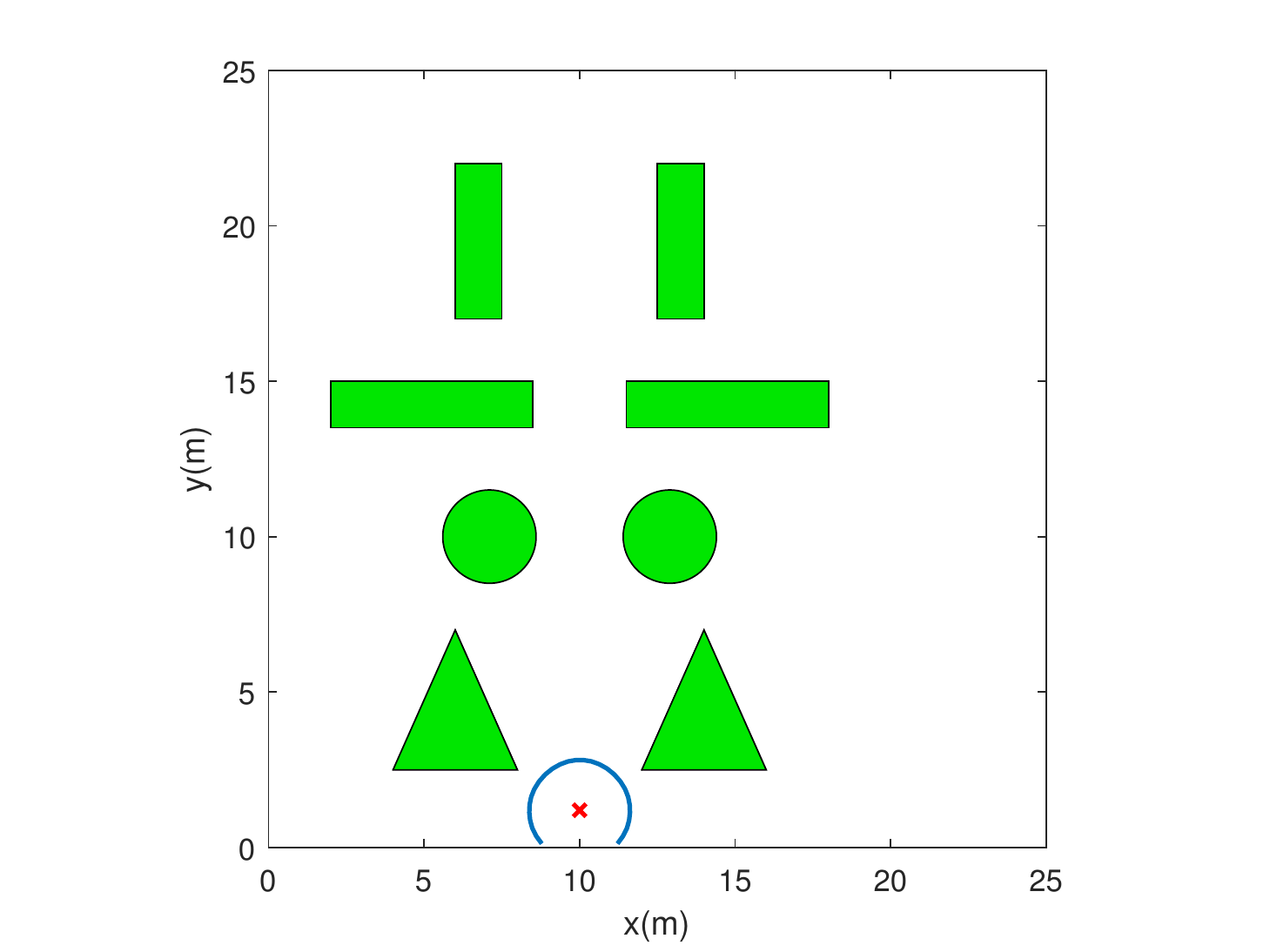}\\
			\centering{\scriptsize{(i)}}
		\end{minipage}
	\end{tabular}
    \hspace{0.5cm}
	\caption{\label{fig:case2_EHGRN} When a target pass through a compound channel, the swarm pattern is formed by EP-GRN, which encircles a target without colliding the channel. (a)-(e) show that the swarm pattern encircle the target in a  shapes way, when the target passes through the channel.}
\end{figure*}

Fig. \ref{fig:case2_EHGRN} shows when a target pass through a compound channel, the swarm pattern that formed by EP-GRN encircle the target in a circle shape. Fig. \ref{fig:case2_EHGRN} (d) and (f) are the narrowest part of the composite channel. That is, the width of the narrow channel and the width of the circular narrow channel are the narrowest part of the composite channel. In these the narrowest parts, swarm pattern collides with obstacles when encircling the target. It shows that EH-GRN optimized swarm pattern can not adapt to these two scenarios well. The limitations of predetermined GRN structure are revealed from the side.

In summary, the proposed automatic design framework can find two GRN-based models, which encircles a target without colliding the channel and verifies the adaptability of the proposed framework.

\section{Conclusion}
\label{sec:conclusion}
This paper proposes an automatic design framework of swarm pattern formation based on multi-objective genetic programming. The proposed framework improves the flexibility of the swarm pattern generation to be applied in various complex and changeable environments as it applies some basic network motifs to automatically structure the GRN-based model. In addition, by introducing obstacles as one of environmental input sources along with that of targets, we address the weakness of the previous H-GRN pattern not being adaptable to obstacles in the sense that a target entrapping pattern itself changes as an obstacle approaches. Numerical simulations considering static/moving targets and obstacles have demonstrated that the proposed approach is able to automatically generate complex patterns highly adaptable and robust to dynamic and unknown environments. As future work, a proof-of-concept experiment will be performed to evaluate the proposed pattern formation algorithm using e-puck education robots in real-world environments.

% or


%\appendices
%\section{Additional experiments}


\section*{Acknowledgment}
This work was supported by the Key Lab of Digital Signal and Image Processing of Guangdong Province, by the Key Laboratory of Intelligent Manufacturing Technology (Shantou University), Ministry of Education, by the Science and Technology Planning Project of Guangdong Province of China under grant 180917144960530, by the Project of Educational Commission of Guangdong Province of China under grant 2017KZDXM032, by the State Key Lab of Digital Manufacturing Equipment \& Technology under grant DMETKF2019020, and by the National Defense Technology Innovation Special Zone Projects.

\ifCLASSOPTIONcaptionsoff
  \newpage
\fi



%\bibliographystyle{IEEEtran}
%\begin{thebibliography}{1}

\bibliographystyle{elsarticle-num}
\bibliography{bare_jrnl_transmag.bib}

%\end{thebibliography}


%\end{IEEEbiography}

%\newpage



%\vfill


\end{document}